\pdfoutput=1

\documentclass[11pt]{article}
\usepackage[]{acl}
\usepackage{times}
\usepackage{latexsym}
\usepackage[T1]{fontenc}
\usepackage[utf8]{inputenc}
\usepackage{microtype}
\usepackage{inconsolata}
\usepackage{ragged2e}
\usepackage{subcaption}
\usepackage{graphicx}
\usepackage{todonotes}
\usepackage[framemethod=TikZ]{mdframed}
\usepackage{xcolor}
\usepackage{listings}
\usepackage{caption}
\usepackage{booktabs}
\usepackage[frozencache,cachedir=.]{minted}

\newmdenv[%
    backgroundcolor=gray!10,
    linecolor=black,
    outerlinewidth=0.5pt,
    roundcorner=1mm,
    skipabove=\topsep,
    skipbelow=\topsep,
    font=\ttfamily\tiny,
]{promptbox}

\title{Are Large Language Model-based Evaluators the Solution to Scaling Up Multilingual Evaluation?}
\author{\parbox{0.9\linewidth}{\centering{Rishav Hada\textsuperscript{\normalfont $\spadesuit$} \quad Varun Gumma\textsuperscript{\normalfont $\spadesuit$} \quad Adrian de Wynter\textsuperscript{\normalfont $\spadesuit$} \\ \quad Harshita Diddee\textsuperscript{\normalfont $\heartsuit$}\thanks{\hspace{0.1cm}Work done when the author was at Microsoft} \quad Mohamed Ahmed\textsuperscript{\normalfont $\spadesuit$} \quad Monojit Choudhury\textsuperscript{\normalfont $\diamondsuit$}\footnotemark[1] \\ \quad Kalika Bali\textsuperscript{\normalfont $\spadesuit$} \quad Sunayana Sitaram\textsuperscript{\normalfont $\spadesuit$}\\
 {\rm  \textsuperscript{\normalfont $\spadesuit$}Microsoft Corporation 
 \quad 
 \textsuperscript{\normalfont $\heartsuit$}Carnegie Mellon University
 \quad
 \textsuperscript{\normalfont $\diamondsuit$}MBZUAI}
{\tt rishavhada@gmail.com, sunayana.sitaram@microsoft.com} }}}

\begin{document}
\maketitle

\begin{abstract}
Large Language Models (LLMs) excel in various Natural Language Processing (NLP) tasks, yet their evaluation, particularly in languages beyond the top $20$, remains inadequate due to existing benchmarks and metrics limitations. Employing LLMs as evaluators to rank or score other models' outputs emerges as a viable solution, addressing the constraints tied to human annotators and established benchmarks. In this study, we explore the potential of LLM-based evaluators, specifically GPT-4 in enhancing multilingual evaluation by calibrating them against $20$K human judgments across three text-generation tasks, five metrics, and eight languages. Our analysis reveals a bias in GPT4-based evaluators towards higher scores, underscoring the necessity of calibration with native speaker judgments, especially in low-resource and non-Latin script languages, to ensure accurate evaluation of LLM performance across diverse languages.
\end{abstract}
\section{Introduction}
Large Language Models (LLMs) can achieve remarkable results on a variety of tasks, sometimes even outperforming humans on certain tasks and domains \cite{openai2023gpt4,chen-ding-2023-probing,veen2023clinical,chiang-lee-2023-large}. However, measuring the performance of LLMs is challenging, as standard NLP benchmarks may not reflect real-world applications. Other hurdles for LLM evaluation include the scarcity of benchmarks for diverse and complex tasks, benchmark saturation, contamination of benchmark data in LLM training data, and the weak correlation between automated metrics and human judgment \cite{jacovi2023stop, chang2023survey, reiter-2018-structured,liu-liu-2008-correlation}. Therefore, researchers have proposed alternative evaluation methods that go beyond benchmarking to assess the abilities and limitations of LLMs \cite{chang2023survey}.

\begin{figure}[t]
    \centering
    \includegraphics[width=\columnwidth]{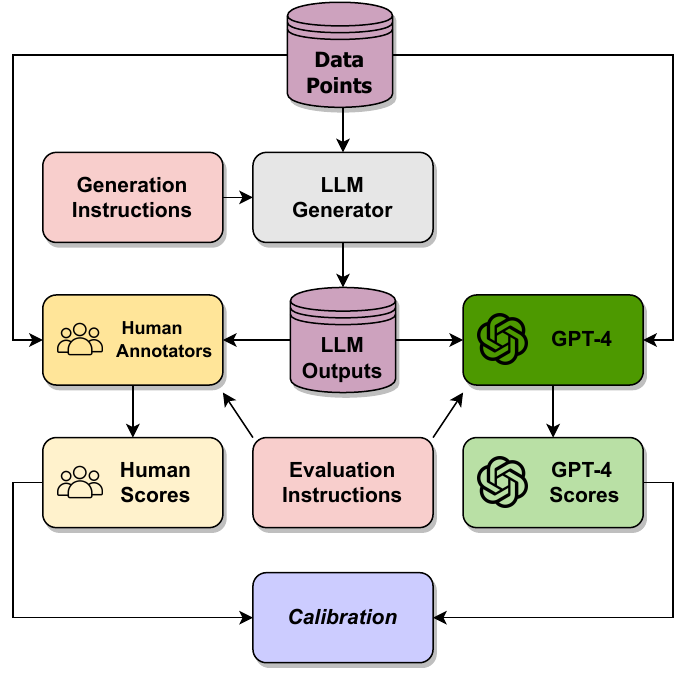}
    \caption{Pipeline of our experiments involving generation, evaluation, and calibration.}
    \label{fig:title_fig}
\end{figure}

While LLMs excel at various tasks in English, their capabilities in other languages are more limited. This disparity may increase the digital divide, preventing a significant portion of the global population from benefiting from LLMs and potentially harming them.
\newcite{ahuja-etal-2023-mega,ahuja2023megaverse} conduct a comprehensive benchmarking of LLMs across the available multilingual benchmarks covering several tasks and languages, and show that the performance of LLMs degrades significantly on languages that are transcribed in non-Latin scripts and under-resourced languages. 

Multilingual evaluation is challenging to scale. Certain language families, such as Indo-European, are over-represented in multilingual benchmarks with other language families having very little presence. There is a scarcity of multilingual benchmarks designed to assess tasks that simulate actual LLM usage in real-world scenarios. The metrics used in these benchmarks may be unsuitable for languages with rich morphology or complex writing systems, as well as phenomena arising from language contact such as borrowing, code-mixing, and transliteration. Evaluation by native speakers is the gold standard for building an accurate picture of model performance, especially in complex tasks without well-defined automated metrics. However, budget constraints, turnaround time, and the lack of easy access to native speakers in some languages all pose challenges in scaling evaluation. This leads to a situation in which LLM performance is unknown for most languages of the world \cite{ahuja2022beyond}.

The success of LLMs in complex tasks such as sentiment analysis, reasoning, problem-solving \cite{mao2023gpteval,arora-etal-2023-llms}, and 
 providing feedback for reducing LLM harms \cite{bai2022constitutional} has led to the question of whether LLMs can replace human annotators, or help augment human evaluation \cite{gilardi2023chatgpt}. Utilizing LLMs as multilingual evaluators is, therefore, an attractive option to decrease costs and circumvent the challenges of scaling assessments by native speakers. However, LLMs have been demonstrated to have inferior performance even in some high-resource languages and have not been evaluated extensively across many languages on dimensions such as toxicity, fairness, and robustness (due to the absence of such benchmarks) \cite{ahuja-etal-2023-mega}, it is prudent to proceed with caution. Failing to do so can lead to misleading results which may further widen the digital divide.
 
In this work, we study whether LLM-based evaluation can be the answer to scaling up multilingual evaluation. In other words, can LLMs serve as substitutes or supplements for human native speakers in delivering useful and accurate insights regarding LLM outputs in non-English languages, while considering diverse aspects of interest like linguistic acceptability, task accomplishment, and safety? Our main contributions are as follows:
\vspace{-2mm}
\begin{enumerate} 
    \item  We present the first evaluation of LLMs, specifically GPT-4 as multilingual evaluators to examine whether LLMs can be used to scale up multilingual evaluation.
    \vspace*{-3mm}
    \item We calibrate LLM judgments on an in-house dataset across three tasks, eight languages, and five dimensions by comparing them to over $20$K human judgments on the same tasks, languages, and dimensions.
        \vspace*{-3mm}  
    \item We evaluate a variety of prompting strategies for LLM-based evaluation in the multilingual setting.
        \vspace*{-3mm}  
    \item We provide a framework for evaluating LLM-evaluators in the multilingual setting that can generalize across tasks, metrics, and languages\footnote{Code available at: \url{https://aka.ms/LLM-Eval}}.
        \vspace*{-3mm}  
    \item We suggest best practices and provide recommendations for future work.
\end{enumerate}
\section{Related Work}
Broadly, there are two main uses of LLMs as evaluators: LLMs can be used as alternatives to metrics that compare human and machine-generated text, such as BLEU \cite{papineni-etal-2002-bleu} and ROUGE \cite{lin-2004-rouge}. Word overlap-based metrics are limited, and LLM-based scorers have been shown to outperform them. GPTScore \cite{fu2023gptscore} is a popular LLM-based framework that can be used to score model outputs based on human-created references along various dimensions. However, these scores still rely on having examples of human-created reference data. 

The second use case of LLMs as evaluators is when the LLM is presented with the output of a system (usually an LLM, sometimes the same model) and asked to judge its quality or safety without any human output to compare against \cite{zheng2023judging}. The LLM is instructed on how to perform this evaluation with the help of the task description, evaluation rubric, and sometimes, one or more examples in the prompt. This is the use case we focus on in this work. 

\newcite{gilardi2023chatgpt} prompt ChatGPT to annotate Tweets across various dimensions such as topic and stance and find that it outperforms crowdworkers. \newcite{shen2023large} explore the use of GPT3.5 as an evaluator for abstractive summarization and find that although GPT is a useful evaluator, as the quality of summarization improves, the quality of evaluation degrades. Along similar lines, \newcite{wang2023chatgpt} evaluate ChatGPT on various NLG tasks and find that it has a high correlation with human judgments. \newcite{kocmi2023large} evaluate the effectiveness of LLMs on evaluation of translation quality and find that LLMs starting from GPT3.5 and above achieve SOTA performance on translation evaluation benchmarks. \newcite{fernandes2023devil} leverage LLMs for fine-grained annotation of errors in Machine Translation outputs. LLM-based evaluators have also been used to score and refine outputs they produce, as described in \newcite{madaan2023self}, ultimately producing outputs that are scored higher on human and automated metrics than the original outputs. \newcite{naismith-etal-2023-automated} explore the use of LLM-based evaluators on scoring written discourse for coherence and find a strong correlation with human judgments. The success of LLM-based evaluators has led many to question whether LLM-based evaluation can replace or augment human evaluation \cite{chiang-lee-2023-large}.

However, there have been studies showing that LLM-based evaluators can have some biases. 
\citet{wu2023style} demonstrate that LLMs tend to prefer answers with factual errors when they are too short or contain grammatical errors. \newcite{pangakis2023automated} highlight the need for validating LLM-based evaluators on a task-by-task basis. \newcite{liu2023gpteval} perform NLG evaluation using GPT-4 and find that although it correlates well with human judgments, it may potentially be biased towards preferring LLM-generated texts. \newcite{koo2023benchmarking} show that LLMs have egocentric bias where they prefer to rank their own outputs highly in evaluation. \newcite{wang2023large} point out that GPT4-based evaluators have positional bias and scores can be easily altered by changing the order of appearance. There are also several ethical issues with the use of LLMs as evaluators described in \newcite{chiang-lee-2023-large}. \newcite{zhang2023wider} suggest that wider and deeper LLMs are fairer evaluators, while \newcite{chan2023chateval} introduce a framework for multiple evaluator agents to reach a consensus, mimicking the situation of having multiple annotators. 

Although there has been some work measuring the calibration of LLM-based evaluators to human judgments \cite{koo2023benchmarking}, previous studies have focused on English, and ours is the first work (to the best of our knowledge) that addresses this problem in the multilingual context.
\section{Experimental Setup}
We perform experiments on a text generation application that is powered by GPT-4, and evaluate the following sub-tasks:

\noindent \textbf{Open Prompt:} This task processes a concise prompt to generate a document adhering to the provided guidelines, producing up to $2,048$ tokens, approximately equivalent to one page in English or Spanish, and marginally less in other languages. \\
\noindent \textbf{Continue Writing:} This task takes two textual inputs, termed ``left'' and ``right'' to generate a coherent continuation between them, accommodating up to $1,000$ tokens. Notably, one of the inputs may be omitted. \\
\noindent \textbf{Summarize:} Engages in standard summarization by condensing a document of at least $500$ words into a succinct summary. It allows for an optional user-defined prompt to tailor the summary format, such as highlighting key points. \\

We cover the following languages: \textit{English (En), French (Fr), German (De), Spanish (Es), Chinese (Zh), Japanese (Ja), Italian (It), Brazilian Portuguese (Pt-Br)}, and \textit{Czech (Cs)}. Of these, the first six are classified as very high resource languages (Class 5, or ``the winners''), while the last three are classified as Class 4 (``the underdogs'') according to \citet{joshi-etal-2020-state}. We plan to extend our study to lower-resource languages in the future. We study the following dimensions of interest: \\
\noindent \textbf{Linguistic Acceptability (LA)}: This measures whether the text sounds right to a native speaker. 
The values of this metric are \{$0$, $1$, $2$\}, with $0$ corresponding to \textit{not acceptable}, $1$ corresponding to \textit{some errors, but acceptable} and $2$ to \textit{perfectly acceptable}. We chose LA as opposed to grammaticality to ensure a comparable, native-speaker-led evaluation that did not require formal training in the language. \\
\noindent \textbf{Output Content Quality (OCQ)}: Whether the general quality of the content is good or not, with values \{$0$, $1$, $2$\}. A score of $0$ could indicate that the output is in the wrong language, is repetitive, or sounds like it has been scraped from the web, or translated. A score of 1 indicates that the output is okay in terms of grammar and word choice but still sounds awkward in the language. A score of $2$ indicates that the text is of high quality.\\
\noindent \textbf{Task Quality (TQ)}: This measures the ability of the model to follow the given instructions in the prompt. The values of this metric are \{$0$, $1$, $2$\}, with $0$ indicating that the model did not follow the instructions at all. 
Likewise, a score of $1$ indicates that the model followed the instructions approximately well and $2$ that it followed perfectly well. \\
The difference between TQ and OCQ is that the latter focuses on whether the content is appealing to a user, while TQ emphasizes the ability of the model to follow the given instructions. \\
\noindent \textbf{Problematic Content (PC)}: Whether there was any offensive or problematic content in the output. This is a binary metric, with $0$ indicating that the output contains this type of content. \\
\noindent \textbf{Hallucinations (H)}: This measures how well-grounded the model's output was to the input content, and/or whether the model output counterfactual information conflicted with the input content. It is a binary metric, with $0$ indicating the presence of hallucinations.

\subsection{Human Evaluation Setup}
For creating this in-house dataset, we asked human judges to evaluate the output of LLM-based systems configured to perform the three tasks described earlier. Each entry was annotated by three annotators. They were contracted through an external annotator services company at a starting rate depending on locale ranging from \$$14$ USD/hr and up to \$$30$ USD/hr. The pay was adjusted based on locale and experience level. Each annotator was given $250$ texts to judge. We used a subset of the annotated data for our experiments. 

\subsubsection{Annotation Guidelines}
We provided annotators with the following information: General instructions about the task (including specific instructions from the prompt) and high-level descriptions of the metrics that we are seeking to evaluate, a description of the file that contained data to be evaluated, and the output format expected. Then we provided detailed descriptions of each metric including the range of values for each metric and examples in English. These examples were provided in the context of different tasks, as each metric could have slightly different interpretations for different tasks. 

\subsubsection{Data Statistics}
Table \ref{tab:data_stat} contains the statistics of the human evaluation dataset for the three tasks across the languages we consider. We create a subset of this data for experimenting with prompting variations and its statistics are available in the \textit{small} column of the aforementioned table. Our \textit{full} dataset contains over $7,300$ data points, while the smaller subset contains over $2,700$ data points. Each of the data points in our dataset was annotated by $3$ annotators. 

\begin{table}[t!]
\centering 
{\tiny
\begin{tabular}{@{}lcccccc|cc@{}}
\toprule
\textit{Lang.} & \multicolumn{2}{c}{\begin{tabular}[c]{@{}c@{}}Open \\ Prompt\end{tabular}} & \multicolumn{2}{c}{Summarize} & \multicolumn{2}{c|}{\begin{tabular}[c]{@{}c@{}}Continue \\ Writing\end{tabular}} & \multicolumn{2}{c}{Agg.} \\ \midrule
\textit{\textbf{}} & Full & Small & Full & Small & Full & Small & Full & Small \\ \cmidrule(l){2-9} 
\textit{Ca} & 255 & 100 & 158 & 100 & 325 & - & 738 & 200 \\
\textit{De} & 246 & 94 & 251 & 100 & 320 & 96 & 817 & 290 \\
\textit{En} & 200 & 200 & 200 & 200 & 200 & 200 & 600 & 600 \\
\textit{Es} & 247 & 93 & 257 & 100 & 593 & 102 & 1097 & 295 \\
\textit{Fr} & 221 & 88 & 256 & 99 & 409 & 97 & 886 & 284 \\
\textit{It} & 256 & 99 & 260 & 100 & 321 & 100 & 837 & 299 \\
\textit{Ja} & 257 & 100 & 259 & 100 & 316 & 102 & 832 & 302 \\
\textit{Pt-Br} & 246 & 94 & 258 & 100 & 327 & 95 & 831 & 289 \\
\textit{Zh} & 255 & 100 & 160 & 99 & 320 & - & 735 & 199 \\ \midrule
\textit{Agg.} & 2183 & 968 & 2059 & 998 & 3131 & 792 & 7373 & 2758 \\ \bottomrule
\end{tabular}
}
\caption{Dataset statistics across tasks and languages.}
\label{tab:data_stat}
\end{table}

\subsection{LLM-based Evaluators}
We use the GPT4-32K model as our LLM-based evaluator with a temperature of $0$, except in our ablation experiments. The model was accessed through Azure.

\subsubsection{Prompts}
Our evaluation prompts are constructed using the \texttt{ \textcolor{blue}{\{\{}guidance\textcolor{blue}{\}\}}} toolkit\footnote{\url{https://github.com/guidance-ai/guidance/tree/main}}. \texttt{guidance} is a DSL that uses handlebar templating to enable the specification of prompts that interleave instructions and generation with data and logic. This makes it simpler to construct and validate complex prompts.

Evaluation prompts were written to be clear, simple, and not tuned for the data or task. All prompts for evaluation were specified in English, as past work has shown that instructions in native languages can lead to worse performance \cite{ahuja-etal-2023-mega}.
 
In writing the evaluation prompts, we started with simple unstructured specifications (Natural language sentences with no formatting or styling) and found that it often led to errors in formatting the outputs correctly or even returning all the expected outputs. We found adding styling and formatting, for example, outputting JSON by providing the prompt with a JSON schema for the expected attributes improved the reliability of the LLM outputs. 

We tried to keep the task and metric description as close as possible to the text that was shown to human annotators for evaluations in the default prompting variation. Each prompt consists of \textsc{system}, \textsc{user}, and \textsc{assistant} components as shown in Figure \ref{fig:prompt} in a generic prompt schema. The metric description for Hallucinations is shown in Figure 
\ref{fig:metricdescription_H}\footnote{Prompts for task description and other metrics are in Appendix \ref{sec:appendix_simple}.}.

\begin{figure}[t!]
\centering
\begin{promptbox}
\justify
\noindent $\langle$system$\rangle$ \\
\# [system](\#instructions) \\
\# Role \\
You are a helpful assistant. \\ \\

\noindent \#\# Task \\
Description of the task \\ \\

\noindent \#\#\# Outputs \\ 
Description and JSON format of expected outputs \\
$\langle$/system$\rangle$ \\ \\

\noindent $\langle$user$\rangle$ \\
Inputs \\
$\langle$/user$\rangle$ \\ \\ 

\noindent $\langle$system$\rangle$ \\
\# [system](\#instructions) \\
Instruction related to evaluation and metrics \\ \\ 

\noindent \#\#\# Metrics \\
Description of the metrics in JSON format \\
$\langle$/system$\rangle$ \\ \\ 

\noindent $\langle$assistant$\rangle$ \\
Generation space for GPT-4 \\
$\langle$/assistant$\rangle$
\end{promptbox}
\caption{General Prompting Schema.}
\label{fig:prompt}
\end{figure}

\begin{figure}[H]
\centering
\begin{promptbox}
\justify
``name": ``hallucinations", \\ 

\noindent ``description": ``Hallucination refers to the
generation of text that is untrue, fabricated,
inconsistent with the given input, deviates
from generally accepted knowledge, or makes
unverifiable claims.", \\ 

\noindent ``scoring": ``1: No hallucinations in the text;
0: text has hallucinations" 
\end{promptbox}
\caption{Metric description for simple instructions (Hallucinations).}
\label{fig:metricdescription_H}
\end{figure}

\subsection{Prompting Variations}
First, we experiment with variations based on the number of metrics evaluated and instructions provided\footnote{All experiments reported in this study are conducted zero-shot unless specified.}. \\
\noindent \textbf{Single Call:} In this variation, we call GPT-4 once per metric, without any in-context examples. \\
\noindent \textbf{Compound Call:}  In this variation, we call GPT-4 once for all the metrics in a single prompt. \\
\noindent \textbf{Single Call - Detailed:}  In this variation, we call GPT-4 once for all the metrics in a single prompt, with a very detailed metrics description. \\
One of the challenges with LLM evaluation is sensitivity to prompting instructions, which can greatly affect the performance of the LLM on tasks, including evaluation. We experiment with providing detailed instructions for each metric in the prompt. Detailed instruction for Hallucination is shown in Figure \ref{fig:metricdescription_complex_apex_H}\footnote{The detailed instructions for all metrics can be found in Figures \ref{fig:metricdescription_complex_LA} - \ref{fig:metricdescription_complex_apex_OCQ} in Appendix \ref{sec:appendix_complex}}.  We queried GPT-4 to produce these instructions by providing it with the instructions given to annotators and manually modifying them. 

\begin{figure*}[t!]
\centering
\begin{promptbox}
``name": ``hallucinations", \\

\noindent ``description": ``Hallucinations assess the extent to which a model's output remains anchored to, and consistent with, the input content provided. Text with hallucinations while linguistically fluent, are factually baseless or counterfactual in relation to the input. These hallucinations can manifest as additions, omissions, or distortions, and might lead to outputs that are misleading or factually incorrect. This metric serves as a check against unwarranted deviations from the ground truth provided in the input. The scoring rubric is described below, with a few possible reasons (which might not be exhaustive) for a given score.",

\begin{minted}{json}
"scoring": {
    "1": {
        "(a)": "The model's output is strictly aligned with and grounded in the information provided in the input.",
        "(b)": "No evidence of added, omitted, or distorted facts that weren't part of the original content.",
        "(c)": "Maintains the integrity of the original information without any unwarranted extrapolations."
    },
    "0": {
        "(a)": "The output introduces statements, claims, or details that weren't present or implied in the input.",
        "(b)": "Contains counterfactual information that directly conflicts with the input content.",
        "(c)": "Demonstrates unexplained deviations, extrapolations, or interpretations not grounded in the provided data."
    }
}
\end{minted}
\end{promptbox}
\caption{Metric description for complex instructions (Hallucinations).}
\label{fig:metricdescription_complex_apex_H}
\end{figure*}

\subsection{Calibration with Human Judgments}
\textbf{Inter-annotator Agreement Analysis:} We assessed inter-annotator agreement (IAA) among three annotators \texttt{Annot1,Annot2,Annot3} using Percentage Agreement (PA) to determine the proportion of data points with consistent annotations across annotators. Weighted F1 scores are documented in Table \ref{tab:F1}. Additionally, Fleiss' Kappa (\(\kappa\)) values, which offer insights into agreement beyond chance, are provided in Table \ref{tab:kappa} (Appendix \ref{sec:kappa}). Since our dataset is skewed towards one or more classes for each of the metrics, $\kappa$ values can be misleading due to known issues with computing expected agreement in such cases \cite{eugenio2004kappa}.

\begin{table*}[t!]
\centering
\small
\begin{tabular}{@{}llcccc@{}}
\toprule
\textit{} & \textit{Name} & \begin{tabular}[c]{@{}c@{}}Annot1\\ Annot2\\ Annot3\end{tabular} & \begin{tabular}[c]{@{}c@{}}AnnotAgg\\ GPT4\_joint\end{tabular} & \begin{tabular}[c]{@{}c@{}}AnnotAgg\\ GPT4\_single\end{tabular} & \begin{tabular}[c]{@{}c@{}}AnnotAgg\\ GPT4\_SD\end{tabular} \\ \midrule
 & Cs & \multicolumn{1}{c}{0.89 $\pm$ 0.09} & \multicolumn{1}{c}{0.81 $\pm$ 0.17} & \multicolumn{1}{c}{0.82 $\pm$ 0.16} & 0.81 $\pm$ 0.17 \\
 & \textit{De} & \multicolumn{1}{c}{0.93 $\pm$ 0.07} & \multicolumn{1}{c}{0.92 $\pm$ 0.10} & \multicolumn{1}{c}{0.93 $\pm$ 0.09} & 0.92 $\pm$ 0.09 \\
 & \textit{En} & 0.98 $\pm$ 0.02 & 0.97 $\pm$ 0.03 & 0.97 $\pm$ 0.03 & 0.96 $\pm$ 0.04 \\
 & \textit{Es} & 0.91 $\pm$ 0.08 & 0.88 $\pm$ 0.11 & 0.89 $\pm$ 0.11 & 0.88 $\pm$ 0.11 \\
 \textit{Lang.} & \textit{Fr} & 0.94 $\pm$ 0.05 & 0.90 $\pm$ 0.10 & 0.90 $\pm$ 0.10 & 0.90 $\pm$ 0.10 \\
 & \textit{It} & 0.94 $\pm$ 0.07 & 0.91 $\pm$ 0.11 & 0.92 $\pm$ 0.10 & 0.91 $\pm$ 0.11 \\
 & \textit{Ja} & 0.91 $\pm$ 0.08 & 0.78 $\pm$ 0.22 & 0.78 $\pm$ 0.21 & 0.78 $\pm$ 0.22 \\
 & \textit{Pt-Br} & 0.96 $\pm$ 0.04 & 0.91 $\pm$ 0.10 & 0.91 $\pm$ 0.10 & 0.90 $\pm$ 0.10 \\
 & \textit{Zh} & 0.89 $\pm$ 0.10 & 0.83 $\pm$ 0.16 & 0.83 $\pm$ 0.16 & 0.83 $\pm$ 0.16 \\ \midrule
 & \textit{H} & 0.98 $\pm$ 0.03 & 0.96 $\pm$ 0.04 & 0.96 $\pm$ 0.04 & 0.96 $\pm$ 0.04 \\
 & \textit{LA} & 0.92 $\pm$ 0.06 & 0.88 $\pm$ 0.13 & 0.89 $\pm$ 0.12 & 0.88 $\pm$ 0.12 \\
 \textit{Metric} & \textit{OCQ} & 0.86 $\pm$ 0.08 & 0.80 $\pm$ 0.12 & 0.80 $\pm$ 0.12 & 0.80 $\pm$ 0.12 \\
 & \textit{PC} & 1.00 $\pm$ 0.01 & 1.00 $\pm$ 0.01 & 1.00 $\pm$ 0.01 & 1.00 $\pm$ 0.01 \\
 & \textit{TQ} & 0.88 $\pm$ 0.06 & 0.76 $\pm$ 0.15 & 0.76 $\pm$ 0.16 & 0.75 $\pm$ 0.16 \\ \midrule
 & \textit{\begin{tabular}[c]{@{}l@{}}Continue \\ Writing\end{tabular}} & 0.94 $\pm$ 0.07 & 0.88 $\pm$ 0.14 & 0.88 $\pm$ 0.14 & 0.88 $\pm$ 0.15 \\
 \textit{Task} & \textit{\begin{tabular}[c]{@{}l@{}}Open \\ Prompt\end{tabular}} & 0.91 $\pm$ 0.08 & 0.83 $\pm$ 0.16 & 0.84 $\pm$ 0.16 & 0.83 $\pm$ 0.16 \\
 & \textit{Summarize} & 0.94 $\pm$ 0.07 & 0.93 $\pm$ 0.09 & 0.93 $\pm$ 0.09 & 0.93 $\pm$ 0.09 \\ \bottomrule
\end{tabular}
\caption{Weighted F1 values for different cases and annotator combinations on the full dataset. GPT4\_SD means GPT4\_single\_detailed}.
\label{tab:F1}
\end{table*}
\noindent \textbf{IAA (3 annotators) and GPT:} We measure IAA between the majority score of the three annotators and the LLM-evaluator. We refer to this as \texttt{AnnotAgg,GPT4} and use PA to measure it. \\
\noindent \textbf{Class distribution:} We analyze the class distribution of scores across tasks, metrics, and languages to check for potential biases in the dataset and LLM-evaluator. 

We perform experiments contrasting compound and single-call prompting on the full dataset and zero-shot vs. few-shot prompting on the smaller dataset. We analyze how well-calibrated our LLM-based evaluators are with respect to human judgments by examining PA, and class distribution of scores. 

\subsection{Ablation Experiments}
In addition, we perform some ablation experiments to check for consistency, the effect of hyperparameters, and few-shot examples. We perform these ablations on the smaller dataset. \\
\noindent \textbf{Consistency check:} We prompt GPT-4 with the same prompt five times to check its consistency. \\
\noindent \textbf{Single Call -- Few-Shot:} In this variation, we call GPT-4 once per metric, with a few in-context examples. We provide examples in the prompt of human judgments for the same task and metric from a held-out dev set. We take the majority vote from the three human annotations per sample as the aggregate class for that sample to choose our few-shot examples. For each task, language, and metric we choose up to two samples per possible class for that metric. Therefore, we have a minimum of two and a maximum of six exemplars as few-shot examples. For all evaluations, the few-shot examples used are fixed. 
\\
\noindent \textbf{Sensitivity analysis:} We check the sensitivity of the Linguistic Acceptability metric evaluation by randomly shuffling $10$\% of the words in the whole text for all instances and checking if the LA score provided by the model changes. \\
\noindent \textbf{Temperature variation:} We vary the temperature parameter to check its effect on LLM evaluation. 

\section{Results}
\subsection{Percentage Agreement}
In this set of graphs, we look at the percentage agreement between LLM-evaluator and the annotators, and between the annotators. We aggregate the results by task, metric, and language.

\begin{figure*}[t!]
    \centering
    \begin{subfigure}[t]{0.31\textwidth}
    \includegraphics[width=\textwidth]{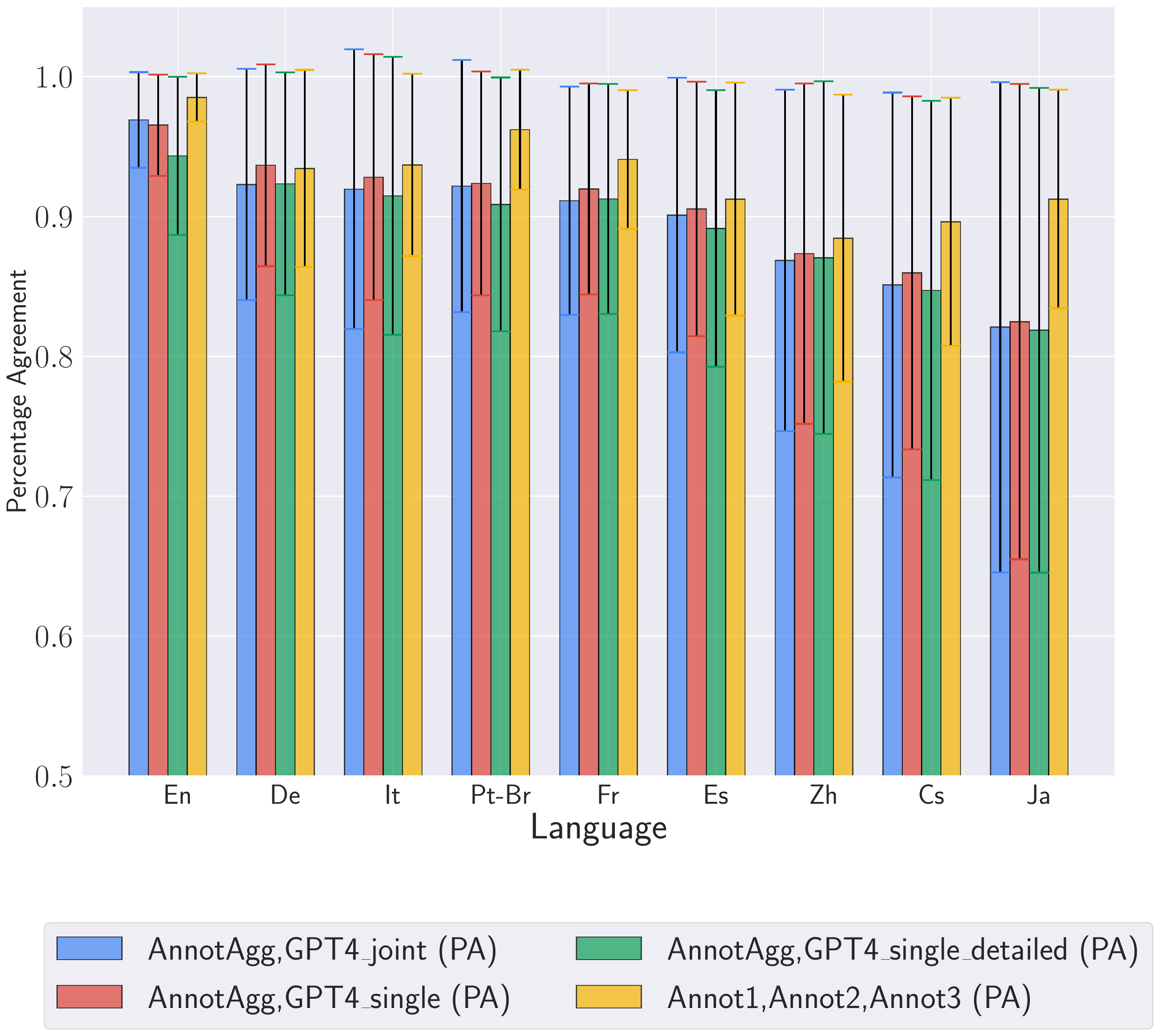}
    \caption{PA by language: Full dataset}
    \label{fig:full_PA_by_language}
    \end{subfigure}
    \begin{subfigure}[t]{0.31\textwidth}
    \includegraphics[width=\textwidth]{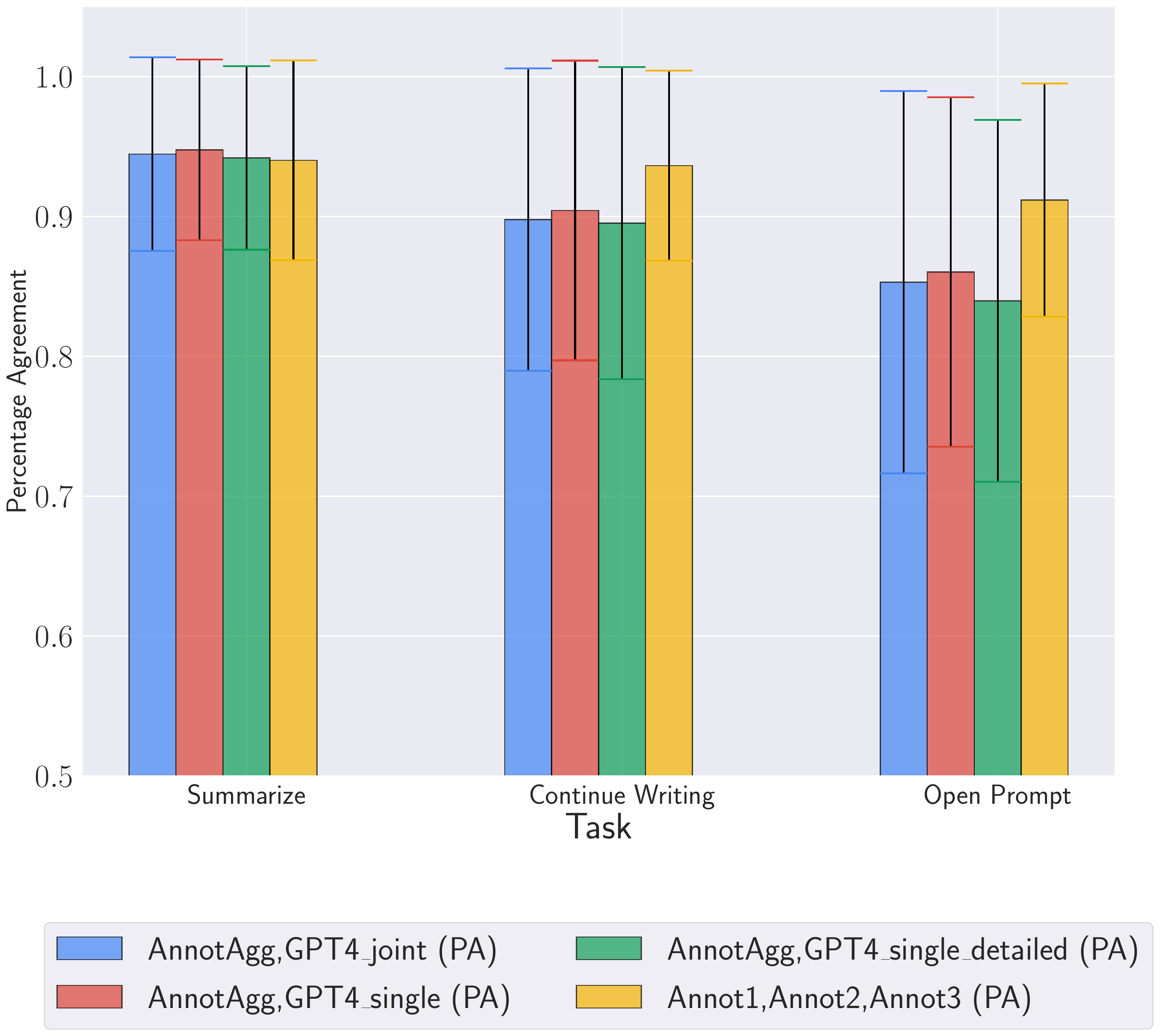}
    \caption{PA by task: Full dataset}
    \label{fig:full_PA_by_task}
    \end{subfigure}
    \begin{subfigure}[t]{0.31\textwidth}
    \includegraphics[width=\textwidth]{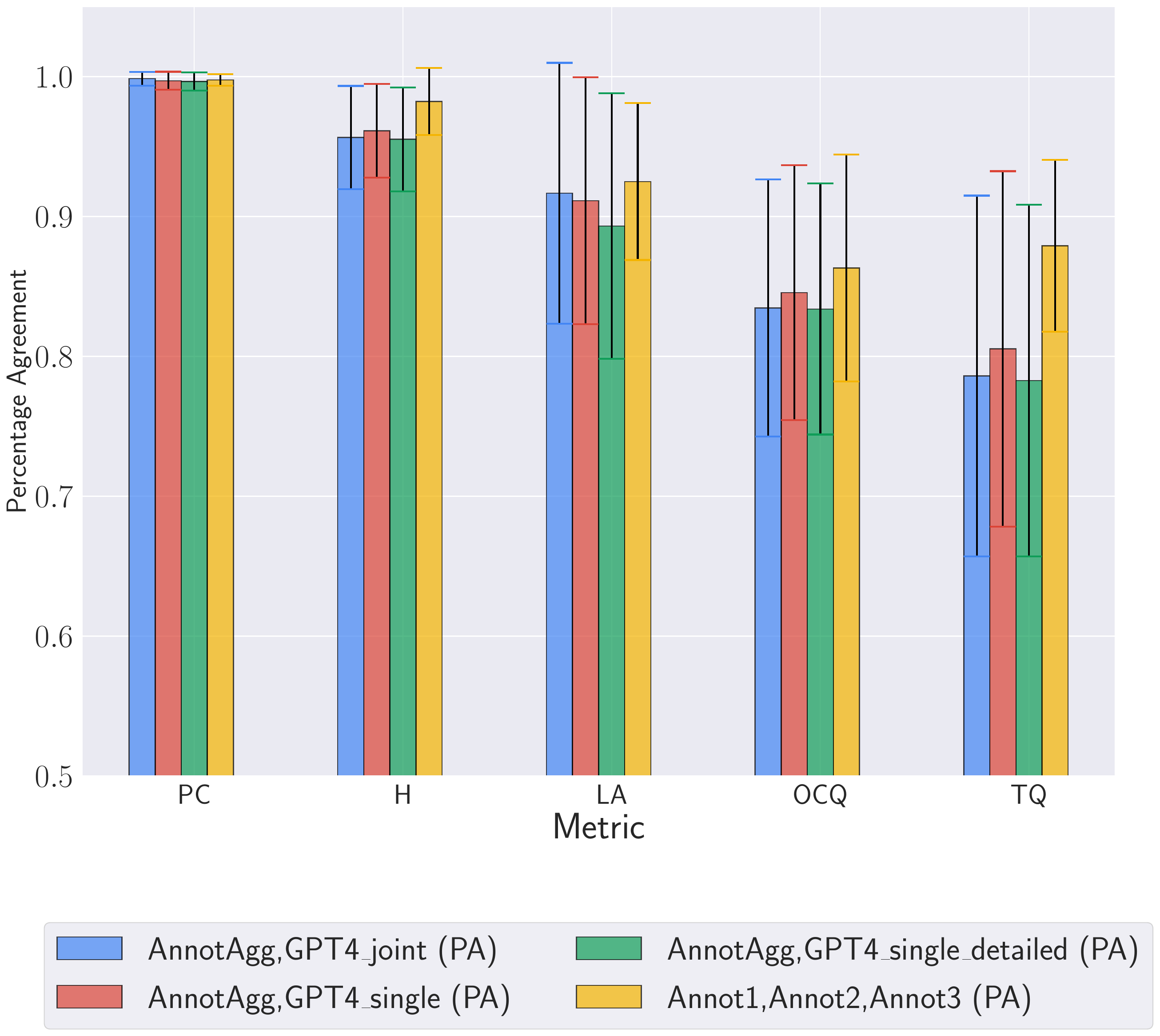}
    \caption{PA by metric: Full dataset}
    \label{fig:full_PA_by_metric}
    \end{subfigure}
    \caption{Percentage Agreement (PA) for different cases and annotator combinations.}
\end{figure*}

Figure \ref{fig:full_PA_by_language} shows the percentage agreement between the aggregate of the human annotator scores and LLM-evaluator for the full dataset. The figures show both joint (compound), single, and single with detailed instructions prompting techniques for the full dataset. We see that the PA between the annotators and GPT is lowest compared to the PA between the human annotators for Japanese and Czech, with the PA between annotators also being lower for Chinese.

Next, we look at PA grouped by metric in Figures \ref{fig:full_PA_by_metric} for the full dataset with the same prompting variations as before. We find that the PA of the LLM-evaluator with the annotators is lower for the OCQ metric. We also find that the PA between annotators is relatively low for the TQ metric, while all the PA values are very high for the problematic content metrics.

Finally, we look at PA aggregated by task in Figure \ref{fig:full_PA_by_task}. We find that PA is lower for the ``Continue Writing'' task, while the PA between GPT and the annotators is lower than the agreement between annotators for the ``Open Prompt'' and ``Continue Writing'' tasks. Overall, we find that the LLM-evaluator prompted using the compound prompt has a lower agreement with human annotators than the single prompt variation. 

Figures \ref{fig:full_PA_by_language}, \ref{fig:full_PA_by_task} and \ref{fig:full_PA_by_metric} compare the PA of the LLM-evaluators with detailed instructions vs. the simpler instructions described earlier. We find that PA drops slightly for all metrics with detailed instructions.

\subsection{Class Distribution}
Next, we examine the distributions of the scores from native speakers and the LLM-evaluator. There are three cases to consider for metrics that have three values: Full agreement (all three annotators give the same score), partial agreement (two of the three give the same score), and no agreement (all three 
 give different scores). In metrics that have binary values, we only have full or partial agreement. We group annotations into these classes and analyze responses across these classes. 

\begin{figure*}[!t]
    \centering
    \begin{subfigure}[t]{0.45\textwidth}
    \includegraphics[width=0.90\textwidth]{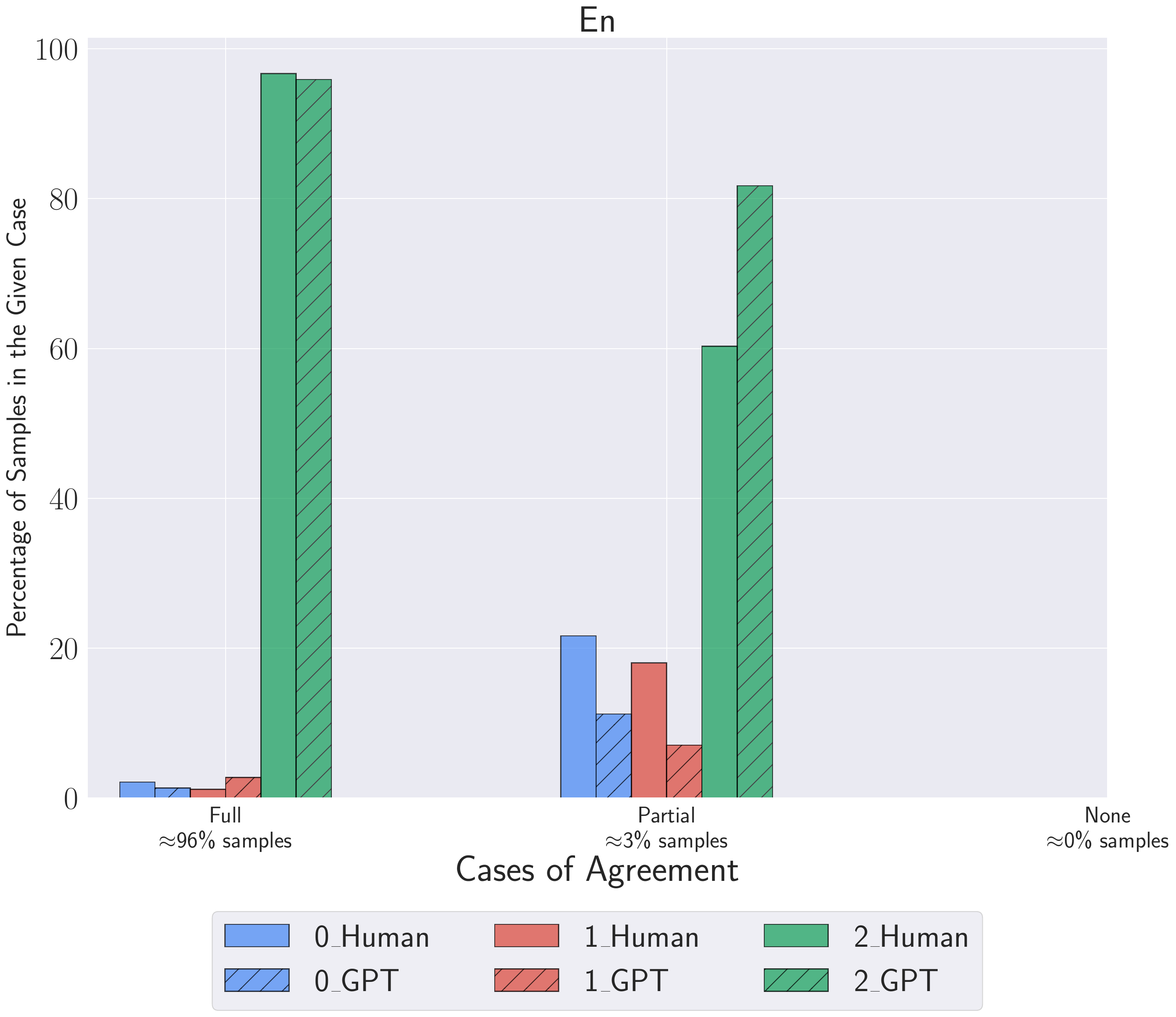}
    \caption{Single Call - English}
    \label{fig:single_english_main}
    \end{subfigure}
    \begin{subfigure}[t]{0.42\textwidth}
    \includegraphics[width=0.90\textwidth]{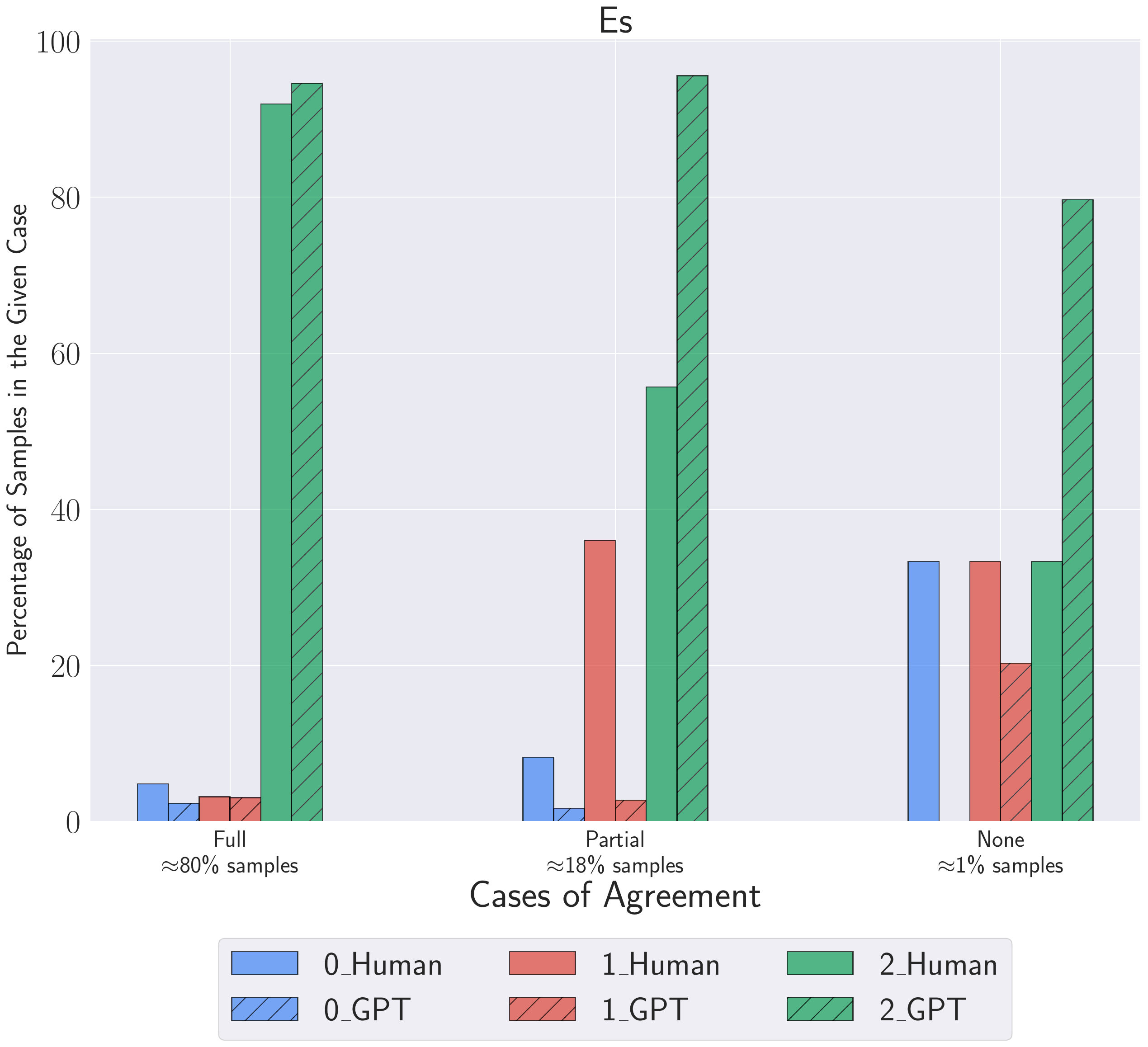}
    \caption{Single Call - Spanish}
    \label{fig:single_spanish_main}
    \end{subfigure}
    \caption{Class distribution for En and Es. Results are aggregated over all tasks and metrics with 3 classes (LA, OCQ, TQ). 
    }
\end{figure*}

\begin{figure*}[!t]
    \centering
    \begin{subfigure}{0.45\textwidth}
    \includegraphics[width=0.90\textwidth]{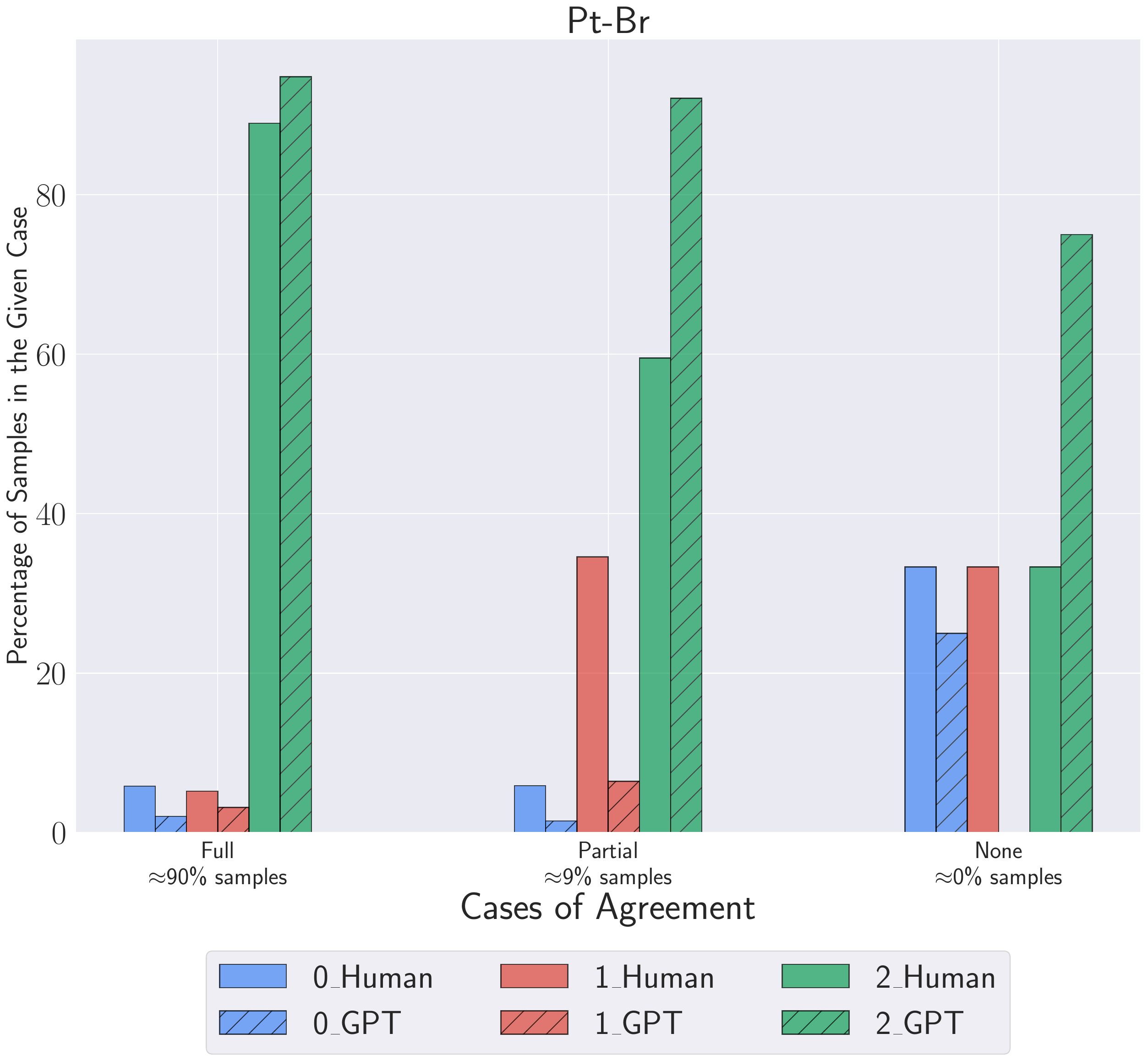}
    \caption{Single Call - Portuguese (Br)}
    \label{fig:single_brazilian_main}
    \end{subfigure}
    \begin{subfigure}{0.45\textwidth}
    \includegraphics[width=0.90\textwidth]{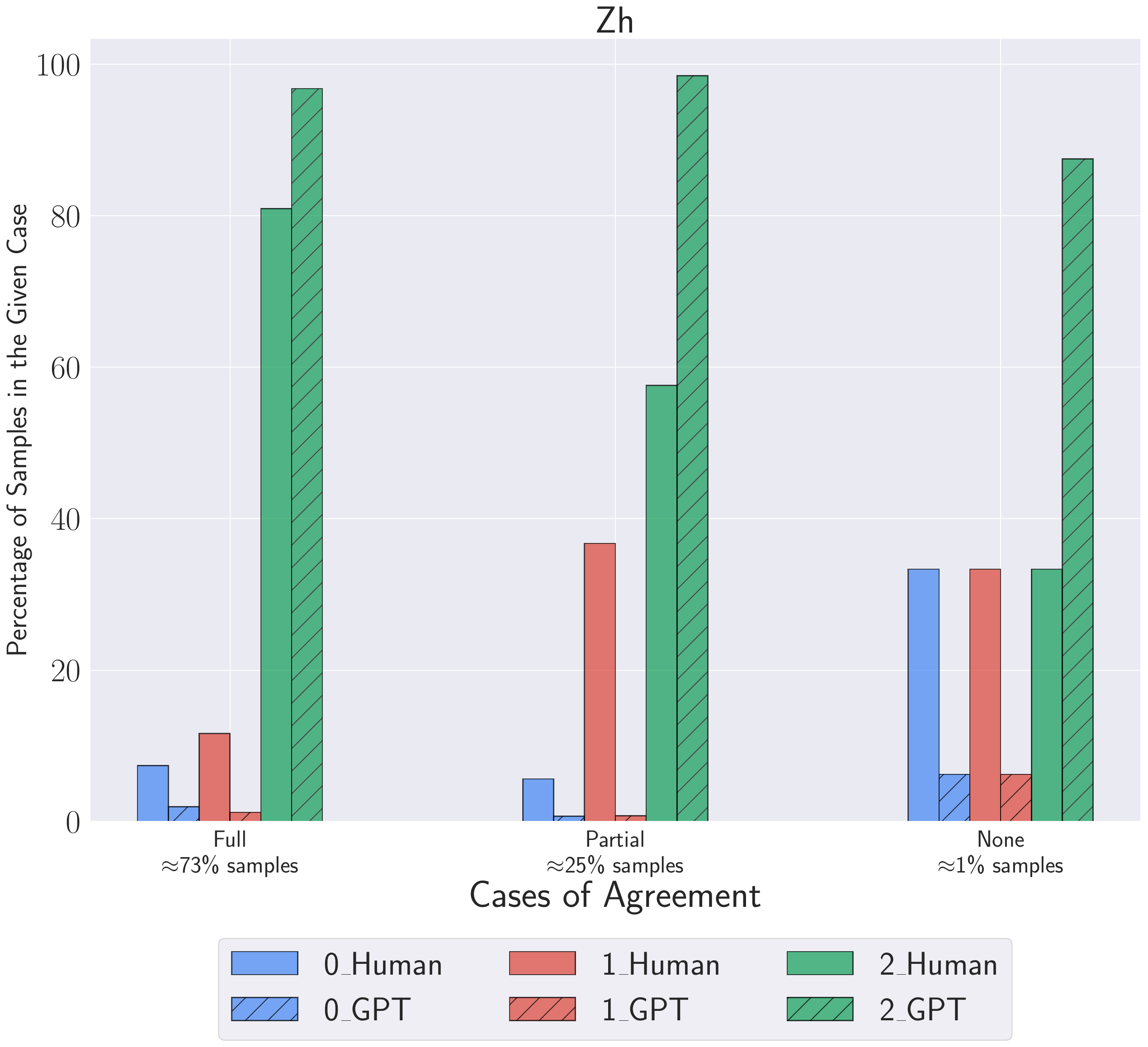}
    \caption{Single Call - Chinese}
    \label{fig:single_chinese_main}
    \end{subfigure}
    \caption{Class distribution for Pt-Br and Zh.  Results are aggregated over all tasks and metrics with $3$ classes (LA, OCQ, TQ). 
    }
\end{figure*}

We present results for metrics that have three values (LA, OCQ, and TQ), with $0$ corresponding to the lowest score and $2$ corresponding to the highest score. In Figures \ref{fig:single_english_main} and \ref{fig:single_spanish_main}, we find that the LLM-evaluator provides a score of 2 in most cases, particularly in cases where human annotators disagree. This is even more evident in the case of non-English languages where there is partial agreement or no agreement between the annotators (around $15$\% of the time on average).

Next, we look at languages that are either lower-resourced or not written in the Latin script. In Figures \ref{fig:single_brazilian_main} and \ref{fig:single_chinese_main} we find that the LLM-evaluator almost never provides scores of $0$ and $1$ in the 26\% of cases that annotators disagree and find similar results for Japanese and Czech shown in Figures \ref{fig:joint_japanese}, \ref{fig:single_japanese}, \ref{fig:joint_czech} and \ref{fig:single_czech} in the Appendix \ref{sec:apex_3class}. Overall, we find that LLM-based evaluators give a score of 2 in most cases. While this is consistent with human evaluations in a large part of the dataset, the LLM-based evaluator continues to assign a score of $2$ even when humans disagree or provide lower scores\footnote{Figures for other languages included in Appendix \ref{sec:apex_3class} and \ref{sec:apex_2class}.}.

\begin{figure*}[!t]
    \centering
    \begin{subfigure}{0.45\linewidth}
    \includegraphics[width=0.90\textwidth]{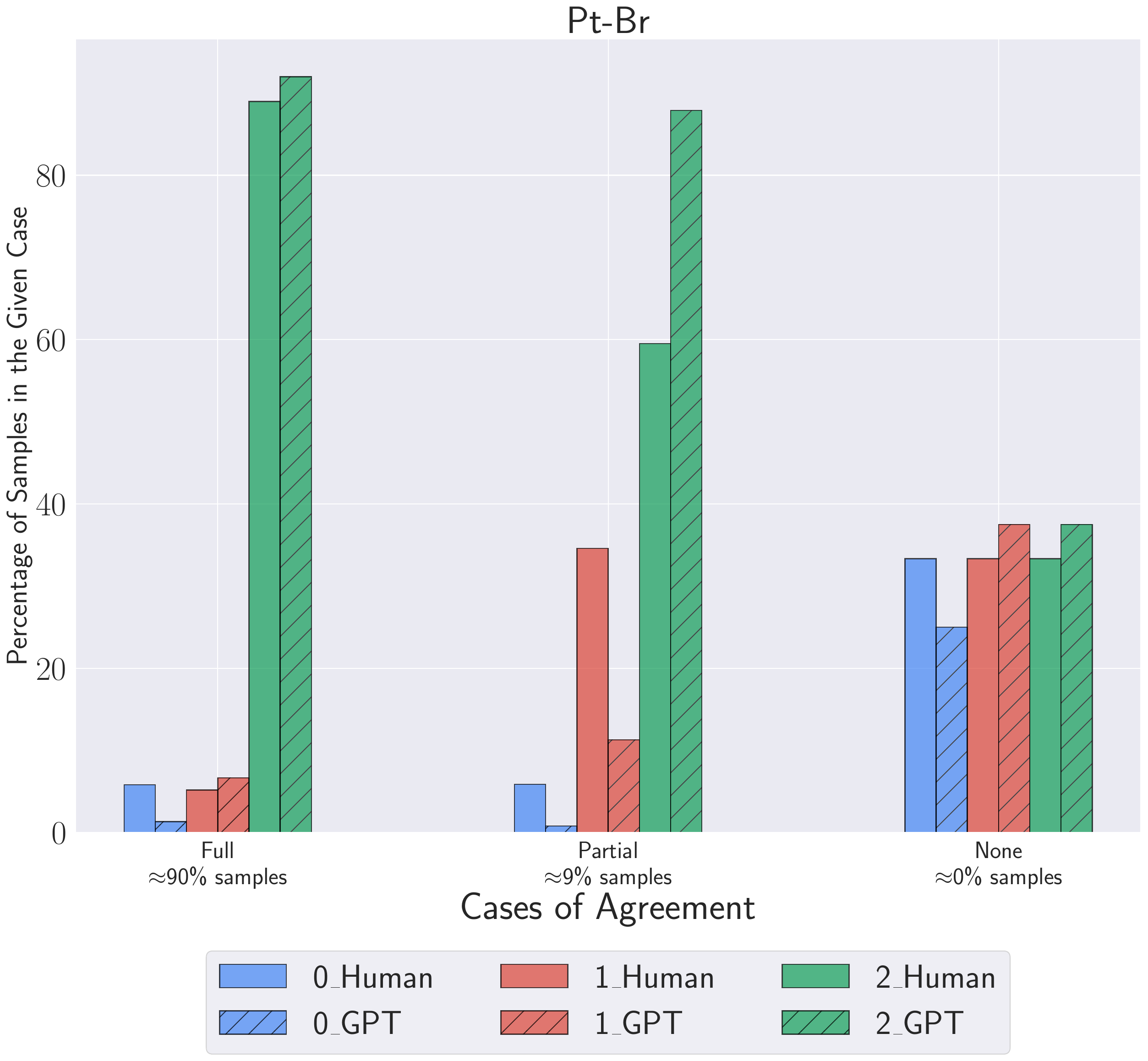}
    \caption{Single call detailed  - Portuguese (Br)}
    \label{fig:scoredist_detailed_language_German}
    \end{subfigure}
    \begin{subfigure}{0.45\linewidth}
    \includegraphics[width=0.90\textwidth]{figures/scoredist_single_language_PortugueseBr.pdf}
    \caption{Single Call (simple) - Portuguese (Br)}
    \label{fig:scoredist_detailed_language_Chinese}
    \end{subfigure}\\
    \caption{Class distribution for Pt-Br detailed and simple. Results are aggregated for all metrics with $3$ classes (LA, OCQ, TQ). 
    }
\end{figure*}

Interestingly, even though PA drops slightly for all metrics with the detailed instructions, we find that the LLM-based evaluator may be slightly less biased towards producing high scores with these instructions as shown in Figures \ref{fig:scoredist_detailed_language_German} and \ref{fig:scoredist_detailed_language_Chinese}. However, more investigation is needed to determine whether detailed instructions or a different prompting strategy can eliminate the bias toward high scores. 

\subsubsection{Consistency Check}
We use a temperature of $0$ and receive the same score and justification in each of the five tries, showing that the LLM-evaluator exhibits high consistency.

\subsubsection{Few-shot Prompting}
Figure \ref{fig:few-shot} in Appendix \ref{sec:few-shot} shows the PA values when few-shot in-context examples are provided. We observe no significant changes in PA values, suggesting that in-context examples might not significantly aid LLM-based evaluators. This also aligns with the findings of \citet{min-etal-2022-rethinking}.

\subsection{Sensitivity Analysis}
\begin{figure}[t!]
\centering
    \includegraphics[width=\columnwidth]{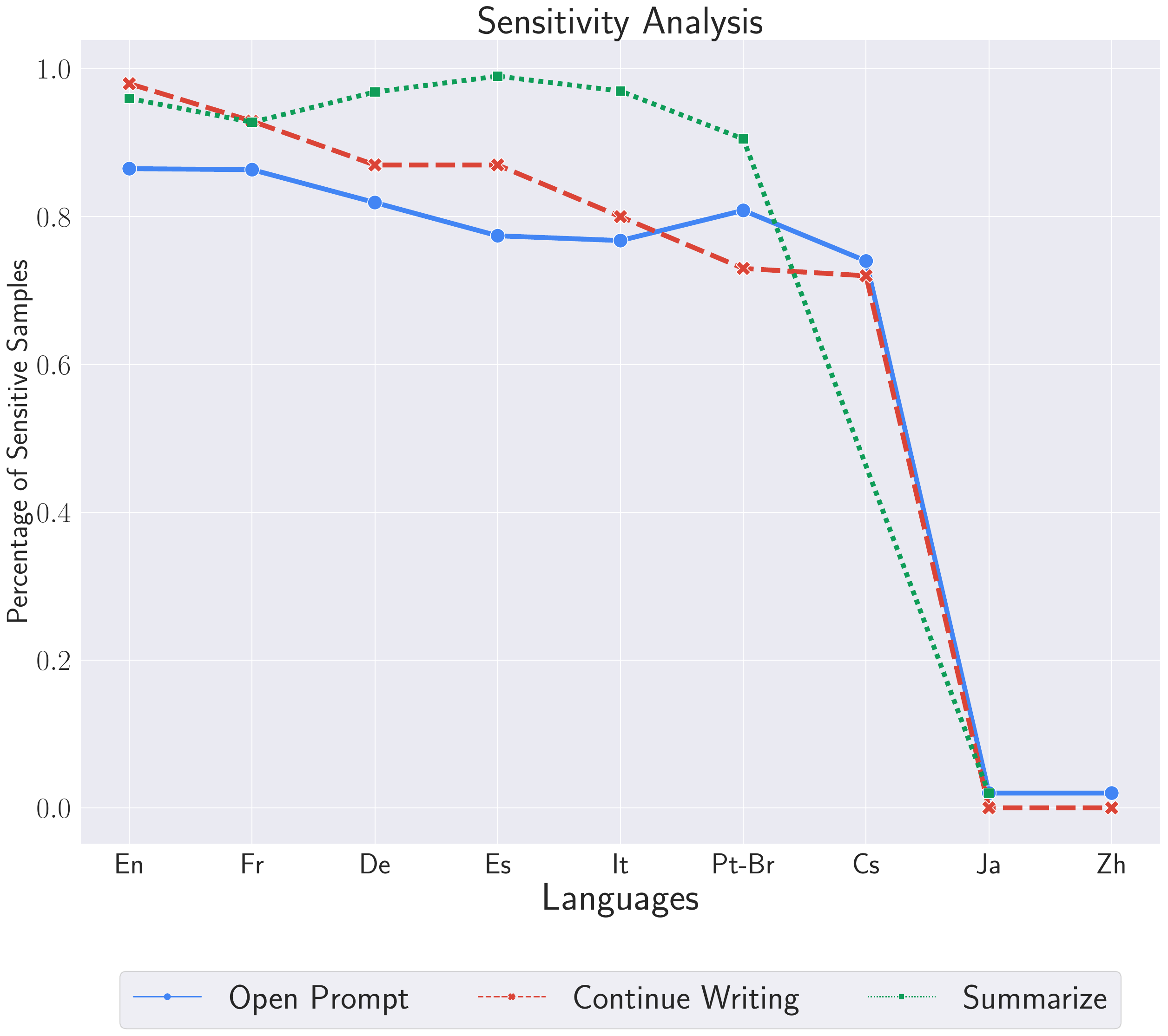}
    \caption{Percentage of samples where GPT evaluation changed from a higher score to a lower score after perturbation. \textit{Note: We do not have Chinese and Czech for the Summarize task in the small dataset.}}
    \label{fig:sense}
\end{figure}

As described earlier, we perturb the word order of sentences and check the sensitivity of the Linguistic Acceptability metric on the \textit{small} dataset. Figure \ref{fig:sense} shows the distribution of cases per language per task where the LLM-based evaluator changes its evaluation from a higher score to a lower score. The evaluator shows the most sensitivity to inputs for the Summarization task for all languages except Japanese. For ``Continue Writing'', Chinese and Japanese show very little sensitivity. For ``Open Prompt", Chinese and Japanese show no sensitivity to the perturbations. One possible explanation for this could be that the evaluator is genuinely less sensitive to these languages. Alternatively, it might be attributed to the flexible word order characteristics of Chinese and Japanese. The examination of tokenizer efficiency in logographic languages, and the exploration of sensitivity across other metrics can be an interesting future exploration.

\subsection{Temperature Variation}
Figure \ref{fig:temp_var} 
in Appendix \ref{sec:temp_variations} show the PA values for temperatures of $0$, $0.3$, $0.7$ and $1.0$. PA reduces as we increase temperature, indicating that a temperature of $0$ should be used for LLM-based evaluators. We also observe that increasing the temperature makes the model more susceptible to any noise in the data, making the evaluations highly stochastic and not reproducible.

\section{Discussion}
Overall, our results indicate that GPT-based evaluators have relatively high consistency for non-English languages when set to a temperature of 0. They also display a fair sensitivity to input variations along the dimension of linguistic acceptability. While LLM-based evaluators show a high Percentage Agreement, there is a noticeable bias towards positive scores, particularly when human opinions differ. It remains uncertain what score an LLM-based evaluator should provide when humans cannot reach a consensus, but consistently high scores in such situations might create a misleading impression of good performance in more challenging evaluations. We find that PA and bias towards higher scores are particularly evident in non-Latin script languages such as Chinese and Japanese, and lower-resource languages such as Czech, which is consistent with prior work on the performance of LLMs on various tasks \cite{ahuja-etal-2023-mega}. 

We experiment with several prompting strategies for LLM-based evaluators and find that evaluating a single metric at a time produces better results than evaluating all metrics in one go, which comes at the cost of having to make multiple calls to the LLM. We also find that providing few-shot examples does not help improve performance. We also provide more detailed instructions to the LLM-evaluator but find that it does not eliminate the problem of bias toward higher scores. In this work, we only use evaluators based on GPT-4. An interesting future direction is the use of smaller models for evaluation or models trained with better coverage of non-English data. We also do not do extensive prompt tuning - future work in this direction includes exploring better prompting approaches including automatically tuning prompts to a held-out set.

Our results show that LLM-based evaluators may perform worse on low-resource and non-Latin script languages. Certain metrics corresponding to output quality and task completion may be challenging for LLM-based evaluators. Hence, we advocate for a cautious approach in using LLM-based evaluators for non-English languages and suggest that all LLM-based multilingual evaluations should be calibrated with a set of human-labeled judgments in each language before deployment.

\section{Limitations}

In this work, we utilize a dataset comprising human assessments of a text generation system executing various tasks in eight languages. As we do not regulate the quality of the system's output, most of the generated texts receive positive ratings from human evaluators. Consequently, the high Percentage Agreement's origin remains unclear – whether it stems from the inclination of the LLM-evaluator to assign high scores or not. In future work, we aim to replicate this study using a dataset with a more balanced distribution of human judgments, achieved by controlling the output quality. 

In this work, we utilize an in-house annotated dataset that, due to restrictions, cannot be released, limiting the reproducibility of our research. However, we intend to make a dataset available to the research community for calibrating LLM-based evaluators in the future. An important research direction is the creation of datasets with good language coverage, multiple annotators per data point, and clear annotation instructions, covering a variety of dimensions to calibrate LLM-based evaluators. Exploring the development of various evaluator personas to represent diverse perspectives of human evaluators and achieve consensus is another research direction that needs further investigation.
\section{Ethical Considerations}

We use the framework by \citet{bender-friedman-2018-data} to discuss the ethical considerations for our work.

\begin{itemize}
\item \textbf{Institutional Review:} We used an in-house dataset annotated by an external company that has long-standing contracts with the organization and was employed by the organization regularly to do this work.
\item \textbf{Data:} The LLM evaluator scores were generated using API calls to GPT-4. The dataset used for calibration is an in-house dataset that will not be released publicly. The dataset was not created with the intent of studying human and LLM calibration; hence, it is not a balanced dataset. Specific instructions were provided to LLMs to avoid generating problematic content, and our ratings of the Problematic Content metrics show no such data; however, the possibility still exists.
\item \textbf{Annotator Demographics:} Annotators were recruited through an external annotator services company. The pay was adjusted after deliberation with the company, based on the annotator's location and expertise. No demographic information is available about the annotators. The annotators are governed by their company's and our organization's privacy policy.
\item \textbf{Annotation Guidelines:} We draw inspiration from the community standards set for similar tasks. Annotators were given general instructions about the task, detailed instructions about the metrics to be evaluated, and examples in English.
\item \textbf{Methods:} In this study, we explore several methods of calibrating human judgments with LLM judgments on various tasks and languages. While these methods can be misused to replace human judgments with LLM judgments, our intent with this study is to highlight the gap between the two and urge the community to proceed with caution.
\end{itemize}

\bibliography{anthology,custom}

\begin{thebibliography}{36}
\expandafter\ifx\csname natexlab\endcsname\relax\def\natexlab#1{#1}\fi

\bibitem[{Ahuja et~al.(2022)Ahuja, Dandapat, Sitaram, and Choudhury}]{ahuja2022beyond}
Kabir Ahuja, Sandipan Dandapat, Sunayana Sitaram, and Monojit Choudhury. 2022.
\newblock Beyond static models and test sets: Benchmarking the potential of pre-trained models across tasks and languages.
\newblock \emph{NLP-Power 2022}, 10(12):64.

\bibitem[{Ahuja et~al.(2023{\natexlab{a}})Ahuja, Diddee, Hada, Ochieng, Ramesh, Jain, Nambi, Ganu, Segal, Ahmed, Bali, and Sitaram}]{ahuja-etal-2023-mega}
Kabir Ahuja, Harshita Diddee, Rishav Hada, Millicent Ochieng, Krithika Ramesh, Prachi Jain, Akshay Nambi, Tanuja Ganu, Sameer Segal, Mohamed Ahmed, Kalika Bali, and Sunayana Sitaram. 2023{\natexlab{a}}.
\newblock \href {https://doi.org/10.18653/v1/2023.emnlp-main.258} {{MEGA}: Multilingual evaluation of generative {AI}}.
\newblock In \emph{Proceedings of the 2023 Conference on Empirical Methods in Natural Language Processing}, pages 4232--4267, Singapore. Association for Computational Linguistics.

\bibitem[{Ahuja et~al.(2023{\natexlab{b}})Ahuja, Aggarwal, Gumma, Watts, Sathe, Ochieng, Hada, Jain, Axmed, Bali, and Sitaram}]{ahuja2023megaverse}
Sanchit Ahuja, Divyanshu Aggarwal, Varun Gumma, Ishaan Watts, Ashutosh Sathe, Millicent Ochieng, Rishav Hada, Prachi Jain, Maxamed Axmed, Kalika Bali, and Sunayana Sitaram. 2023{\natexlab{b}}.
\newblock \href {http://arxiv.org/abs/2311.07463} {Megaverse: Benchmarking large language models across languages, modalities, models and tasks}.

\bibitem[{Arora et~al.(2023)Arora, Singh, and {Mausam}}]{arora-etal-2023-llms}
Daman Arora, Himanshu Singh, and {Mausam}. 2023.
\newblock \href {https://doi.org/10.18653/v1/2023.emnlp-main.468} {Have {LLM}s advanced enough? a challenging problem solving benchmark for large language models}.
\newblock In \emph{Proceedings of the 2023 Conference on Empirical Methods in Natural Language Processing}, pages 7527--7543, Singapore. Association for Computational Linguistics.

\bibitem[{Bai et~al.(2022)Bai, Kadavath, Kundu, Askell, Kernion, Jones, Chen, Goldie, Mirhoseini, McKinnon et~al.}]{bai2022constitutional}
Yuntao Bai, Saurav Kadavath, Sandipan Kundu, Amanda Askell, Jackson Kernion, Andy Jones, Anna Chen, Anna Goldie, Azalia Mirhoseini, Cameron McKinnon, et~al. 2022.
\newblock Constitutional ai: Harmlessness from ai feedback.
\newblock \emph{arXiv preprint arXiv:2212.08073}.

\bibitem[{Bender and Friedman(2018)}]{bender-friedman-2018-data}
Emily~M. Bender and Batya Friedman. 2018.
\newblock \href {https://doi.org/10.1162/tacl_a_00041} {Data statements for natural language processing: Toward mitigating system bias and enabling better science}.
\newblock \emph{Transactions of the Association for Computational Linguistics}, 6:587--604.

\bibitem[{Chan et~al.(2023)Chan, Chen, Su, Yu, Xue, Zhang, Fu, and Liu}]{chan2023chateval}
Chi-Min Chan, Weize Chen, Yusheng Su, Jianxuan Yu, Wei Xue, Shanghang Zhang, Jie Fu, and Zhiyuan Liu. 2023.
\newblock Chateval: Towards better llm-based evaluators through multi-agent debate.
\newblock \emph{arXiv preprint arXiv:2308.07201}.

\bibitem[{Chang et~al.(2023)Chang, Wang, Wang, Wu, Zhu, Chen, Yang, Yi, Wang, Wang et~al.}]{chang2023survey}
Yupeng Chang, Xu~Wang, Jindong Wang, Yuan Wu, Kaijie Zhu, Hao Chen, Linyi Yang, Xiaoyuan Yi, Cunxiang Wang, Yidong Wang, et~al. 2023.
\newblock A survey on evaluation of large language models.
\newblock \emph{arXiv preprint arXiv:2307.03109}.

\bibitem[{Chen and Ding(2023)}]{chen-ding-2023-probing}
Honghua Chen and Nai Ding. 2023.
\newblock \href {https://doi.org/10.18653/v1/2023.findings-emnlp.858} {Probing the {``}creativity{''} of large language models: Can models produce divergent semantic association?}
\newblock In \emph{Findings of the Association for Computational Linguistics: EMNLP 2023}, pages 12881--12888, Singapore. Association for Computational Linguistics.

\bibitem[{Chiang and Lee(2023)}]{chiang-lee-2023-large}
Cheng-Han Chiang and Hung-yi Lee. 2023.
\newblock \href {https://doi.org/10.18653/v1/2023.acl-long.870} {Can large language models be an alternative to human evaluations?}
\newblock In \emph{Proceedings of the 61st Annual Meeting of the Association for Computational Linguistics (Volume 1: Long Papers)}, pages 15607--15631, Toronto, Canada. Association for Computational Linguistics.

\bibitem[{Eugenio and Glass(2004)}]{eugenio2004kappa}
Barbara~Di Eugenio and Michael Glass. 2004.
\newblock The kappa statistic: A second look.
\newblock \emph{Computational linguistics}, 30(1):95--101.

\bibitem[{Fernandes et~al.(2023)Fernandes, Deutsch, Finkelstein, Riley, Martins, Neubig, Garg, Clark, Freitag, and Firat}]{fernandes2023devil}
Patrick Fernandes, Daniel Deutsch, Mara Finkelstein, Parker Riley, Andr{\'e}~FT Martins, Graham Neubig, Ankush Garg, Jonathan~H Clark, Markus Freitag, and Orhan Firat. 2023.
\newblock The devil is in the errors: Leveraging large language models for fine-grained machine translation evaluation.
\newblock \emph{arXiv preprint arXiv:2308.07286}.

\bibitem[{Fu et~al.(2023)Fu, Ng, Jiang, and Liu}]{fu2023gptscore}
Jinlan Fu, See-Kiong Ng, Zhengbao Jiang, and Pengfei Liu. 2023.
\newblock Gptscore: Evaluate as you desire.
\newblock \emph{arXiv preprint arXiv:2302.04166}.

\bibitem[{Gilardi et~al.(2023)Gilardi, Alizadeh, and Kubli}]{gilardi2023chatgpt}
Fabrizio Gilardi, Meysam Alizadeh, and Ma{\"e}l Kubli. 2023.
\newblock Chatgpt outperforms crowd-workers for text-annotation tasks.
\newblock \emph{arXiv preprint arXiv:2303.15056}.

\bibitem[{Jacovi et~al.(2023)Jacovi, Caciularu, Goldman, and Goldberg}]{jacovi2023stop}
Alon Jacovi, Avi Caciularu, Omer Goldman, and Yoav Goldberg. 2023.
\newblock Stop uploading test data in plain text: Practical strategies for mitigating data contamination by evaluation benchmarks.
\newblock \emph{arXiv preprint arXiv:2305.10160}.

\bibitem[{Joshi et~al.(2020)Joshi, Santy, Budhiraja, Bali, and Choudhury}]{joshi-etal-2020-state}
Pratik Joshi, Sebastin Santy, Amar Budhiraja, Kalika Bali, and Monojit Choudhury. 2020.
\newblock \href {https://doi.org/10.18653/v1/2020.acl-main.560} {The state and fate of linguistic diversity and inclusion in the {NLP} world}.
\newblock In \emph{Proceedings of the 58th Annual Meeting of the Association for Computational Linguistics}, pages 6282--6293, Online. Association for Computational Linguistics.

\bibitem[{Kocmi and Federmann(2023)}]{kocmi2023large}
Tom Kocmi and Christian Federmann. 2023.
\newblock Large language models are state-of-the-art evaluators of translation quality.
\newblock \emph{arXiv preprint arXiv:2302.14520}.

\bibitem[{Koo et~al.(2023)Koo, Lee, Raheja, Park, Kim, and Kang}]{koo2023benchmarking}
Ryan Koo, Minhwa Lee, Vipul Raheja, Jong~Inn Park, Zae~Myung Kim, and Dongyeop Kang. 2023.
\newblock \href {http://arxiv.org/abs/2309.17012} {Benchmarking cognitive biases in large language models as evaluators}.

\bibitem[{Lin(2004)}]{lin-2004-rouge}
Chin-Yew Lin. 2004.
\newblock \href {https://aclanthology.org/W04-1013} {{ROUGE}: A package for automatic evaluation of summaries}.
\newblock In \emph{Text Summarization Branches Out}, pages 74--81, Barcelona, Spain. Association for Computational Linguistics.

\bibitem[{Liu and Liu(2008)}]{liu-liu-2008-correlation}
Feifan Liu and Yang Liu. 2008.
\newblock \href {https://aclanthology.org/P08-2051} {Correlation between {ROUGE} and human evaluation of extractive meeting summaries}.
\newblock In \emph{Proceedings of ACL-08: HLT, Short Papers}, pages 201--204, Columbus, Ohio. Association for Computational Linguistics.

\bibitem[{Liu et~al.(2023)Liu, Iter, Xu, Wang, Xu, and Zhu}]{liu2023gpteval}
Yang Liu, Dan Iter, Yichong Xu, Shuohang Wang, Ruochen Xu, and Chenguang Zhu. 2023.
\newblock Gpteval: Nlg evaluation using gpt-4 with better human alignment.
\newblock \emph{arXiv preprint arXiv:2303.16634}.

\bibitem[{Madaan et~al.(2023)Madaan, Tandon, Gupta, Hallinan, Gao, Wiegreffe, Alon, Dziri, Prabhumoye, Yang et~al.}]{madaan2023self}
Aman Madaan, Niket Tandon, Prakhar Gupta, Skyler Hallinan, Luyu Gao, Sarah Wiegreffe, Uri Alon, Nouha Dziri, Shrimai Prabhumoye, Yiming Yang, et~al. 2023.
\newblock Self-refine: Iterative refinement with self-feedback.
\newblock \emph{arXiv preprint arXiv:2303.17651}.

\bibitem[{Mao et~al.(2023)Mao, Chen, Zhang, Guerin, and Cambria}]{mao2023gpteval}
Rui Mao, Guanyi Chen, Xulang Zhang, Frank Guerin, and Erik Cambria. 2023.
\newblock Gpteval: A survey on assessments of chatgpt and gpt-4.
\newblock \emph{arXiv preprint arXiv:2308.12488}.

\bibitem[{Min et~al.(2022)Min, Lyu, Holtzman, Artetxe, Lewis, Hajishirzi, and Zettlemoyer}]{min-etal-2022-rethinking}
Sewon Min, Xinxi Lyu, Ari Holtzman, Mikel Artetxe, Mike Lewis, Hannaneh Hajishirzi, and Luke Zettlemoyer. 2022.
\newblock \href {https://doi.org/10.18653/v1/2022.emnlp-main.759} {Rethinking the role of demonstrations: What makes in-context learning work?}
\newblock In \emph{Proceedings of the 2022 Conference on Empirical Methods in Natural Language Processing}, pages 11048--11064, Abu Dhabi, United Arab Emirates. Association for Computational Linguistics.

\bibitem[{Naismith et~al.(2023)Naismith, Mulcaire, and Burstein}]{naismith-etal-2023-automated}
Ben Naismith, Phoebe Mulcaire, and Jill Burstein. 2023.
\newblock \href {https://doi.org/10.18653/v1/2023.bea-1.32} {Automated evaluation of written discourse coherence using {GPT}-4}.
\newblock In \emph{Proceedings of the 18th Workshop on Innovative Use of NLP for Building Educational Applications (BEA 2023)}, pages 394--403, Toronto, Canada. Association for Computational Linguistics.

\bibitem[{OpenAI(2023)}]{openai2023gpt4}
OpenAI. 2023.
\newblock \href {http://arxiv.org/abs/2303.08774} {Gpt-4 technical report}.

\bibitem[{Pangakis et~al.(2023)Pangakis, Wolken, and Fasching}]{pangakis2023automated}
Nicholas Pangakis, Samuel Wolken, and Neil Fasching. 2023.
\newblock Automated annotation with generative ai requires validation.
\newblock \emph{arXiv preprint arXiv:2306.00176}.

\bibitem[{Papineni et~al.(2002)Papineni, Roukos, Ward, and Zhu}]{papineni-etal-2002-bleu}
Kishore Papineni, Salim Roukos, Todd Ward, and Wei-Jing Zhu. 2002.
\newblock \href {https://doi.org/10.3115/1073083.1073135} {{B}leu: a method for automatic evaluation of machine translation}.
\newblock In \emph{Proceedings of the 40th Annual Meeting of the Association for Computational Linguistics}, pages 311--318, Philadelphia, Pennsylvania, USA. Association for Computational Linguistics.

\bibitem[{Reiter(2018)}]{reiter-2018-structured}
Ehud Reiter. 2018.
\newblock \href {https://doi.org/10.1162/coli_a_00322} {A structured review of the validity of {BLEU}}.
\newblock \emph{Computational Linguistics}, 44(3):393--401.

\bibitem[{Shen et~al.(2023)Shen, Cheng, You, and Bing}]{shen2023large}
Chenhui Shen, Liying Cheng, Yang You, and Lidong Bing. 2023.
\newblock Are large language models good evaluators for abstractive summarization?
\newblock \emph{arXiv preprint arXiv:2305.13091}.

\bibitem[{Veen et~al.(2023)Veen, Uden, Blankemeier, Delbrouck, Aali, Blüthgen, Pareek, Polacin, Collins, Ahuja, Langlotz, Hom, Gatidis, Pauly, and Chaudhari}]{veen2023clinical}
Dave~Van Veen, Cara~Van Uden, Louis Blankemeier, Jean-Benoit Delbrouck, Asad Aali, Christian Blüthgen, A.~Pareek, Malgorzata Polacin, William Collins, Neera Ahuja, C.~Langlotz, Jason Hom, S.~Gatidis, John Pauly, and Akshay~S Chaudhari. 2023.
\newblock \href {https://doi.org/10.21203/rs.3.rs-3483777/v1} {Clinical text summarization: Adapting large language models can outperform human experts}.
\newblock \emph{Research Square}.

\bibitem[{Wang et~al.(2023{\natexlab{a}})Wang, Liang, Meng, Shi, Li, Xu, Qu, and Zhou}]{wang2023chatgpt}
Jiaan Wang, Yunlong Liang, Fandong Meng, Haoxiang Shi, Zhixu Li, Jinan Xu, Jianfeng Qu, and Jie Zhou. 2023{\natexlab{a}}.
\newblock Is chatgpt a good nlg evaluator? a preliminary study.
\newblock \emph{arXiv preprint arXiv:2303.04048}.

\bibitem[{Wang et~al.(2023{\natexlab{b}})Wang, Li, Chen, Zhu, Lin, Cao, Liu, Liu, and Sui}]{wang2023large}
Peiyi Wang, Lei Li, Liang Chen, Dawei Zhu, Binghuai Lin, Yunbo Cao, Qi~Liu, Tianyu Liu, and Zhifang Sui. 2023{\natexlab{b}}.
\newblock Large language models are not fair evaluators.
\newblock \emph{arXiv preprint arXiv:2305.17926}.

\bibitem[{Wu and Aji(2023)}]{wu2023style}
Minghao Wu and Alham~Fikri Aji. 2023.
\newblock Style over substance: Evaluation biases for large language models.
\newblock \emph{arXiv preprint arXiv: 2307.03025}.

\bibitem[{Zhang et~al.(2023)Zhang, Yu, Yu, Lv, Liu, Huang, Xu, and Li}]{zhang2023wider}
Xinghua Zhang, Bowen Yu, Haiyang Yu, Yangyu Lv, Tingwen Liu, Fei Huang, Hongbo Xu, and Yongbin Li. 2023.
\newblock Wider and deeper llm networks are fairer llm evaluators.
\newblock \emph{arXiv preprint arXiv:2308.01862}.

\bibitem[{Zheng et~al.(2023)Zheng, Chiang, Sheng, Zhuang, Wu, Zhuang, Lin, Li, Li, Xing, Zhang, Gonzalez, and Stoica}]{zheng2023judging}
Lianmin Zheng, Wei-Lin Chiang, Ying Sheng, Siyuan Zhuang, Zhanghao Wu, Yonghao Zhuang, Zi~Lin, Zhuohan Li, Dacheng Li, Eric Xing, Hao Zhang, Joseph~E. Gonzalez, and Ion Stoica. 2023.
\newblock \href {https://openreview.net/forum?id=uccHPGDlao} {Judging {LLM}-as-a-judge with {MT}-bench and chatbot arena}.
\newblock In \emph{Thirty-seventh Conference on Neural Information Processing Systems Datasets and Benchmarks Track}.

\end{thebibliography}
\bibliographystyle{acl_natbib}

\clearpage
\appendix
\section{Appendix}
\label{sec:appendix}
\subsection{Prompts for Simple Instructions}
\label{sec:appendix_simple}
Figure \ref{fig:taskdescription} shows task description. Figures \ref{fig:metricdescription_LA} - \ref{fig:metricdescription_OCQ} show simple instructions for various metrics.

\begin{figure}[H]
\centering
\begin{promptbox}
``Open Prompt": ``Given a short user provided starting prompt and its concise completion (which is roughly a page long), your task is to evaluate the completion with respect to the starting prompt and listed set of metrics. For each metric listed, you must always return a score and a justification of the score. Note that, both the starting prompt and its completion are given in \{\{language\}\}.", \\

\noindent ``Continue Writing": ``Given two passages (passage\_a and passage\_b), one of which may be empty, and third passage (passage\_c), which aims to provide a seamless transitions between passage\_a and passage\_b. Your task is to evaluate the passage\_c with respect to the listed set of metrics. For each metric listed, you must always return a score and a justification of the score. Note that, all three passages are given in \{\{language\}\}.", \\
 
\noindent ``Summarize": ``Given a passage and a brief summary of that passage which attempts to capture the essence of it, your task is to evaluate the summary with respect to the given passage and listed set of metrics. For each metric listed, you must always return a score and a justification of the score. Note that, both the passage and its summary are given in \{\{language\}\}."
\end{promptbox}
\caption{Task description.}
\label{fig:taskdescription}
\end{figure}

\begin{figure}[H]
\centering
\begin{promptbox}
\justify
``name": ``linguistic\_acceptability", \\

\noindent ``description": ``Linguistic acceptability means
does this sound right to a native speaker?, not
does this stick to the rules of the grammar.", \\

\noindent ``scoring": "0: not acceptable; 1: some weird 
things but ok; 2: no errors found/acceptable." 
\end{promptbox}
\caption{Metric description for simple instructions (Linguistic Acceptability).}
\label{fig:metricdescription_LA}
\end{figure}

\begin{figure}[H]
\centering
\begin{promptbox}
\justify
``name": ``problematic\_content", \\ 

\noindent ``description": ``Problematic content refers to
the presence of harmful (bias, offensive, toxic,
for example) content in the output.", \\

\noindent ``scoring": ``1: no problematic content in the
text; 0: if there is problematic content in 
the text" 
\end{promptbox}
\caption{Metric description for simple instructions (Problematic Content).}
\label{fig:metricdescription_PC}
\end{figure}

\begin{figure}[H]
\centering
\begin{promptbox}
\justify
``name": ``task quality", \\

\noindent ``description": ``The quality of the output is
related to the task. We are evaluating whether
the model did what the task asked.", \\

\noindent ``scoring": ``0: the model did not do what the
task asked; 1: mostly did what the task asked,
with some errors; 2: did what the task asked." 
\end{promptbox}
\caption{Metric description for simple instructions (Task Quality).}
\label{fig:metricdescription_TQ}
\end{figure}

\begin{figure}[H]
\centering
\begin{promptbox}
\justify
``name": ``output content quality", \\

\noindent ``description": ``Low-Quality Content means
whether the discourse (text) is any good.", \\

\noindent ``scoring": ``0: bad content -- If the text
sounds repetitive (or is non-factual/
inconsistent or it's not in the given language, 
or seems to have been web-scrapped); 1: OK
content, but some flaws found -- If it's ok
(grammatical, lexically, vocab is good) but
kind of goes around in circles; 2; good or
above content." 
\end{promptbox}
\caption{Metric description for simple instructions (Output Quality Content).}
\label{fig:metricdescription_OCQ}
\end{figure}

\subsection{Prompts for Detailed Instructions}
\label{sec:appendix_complex}

Figures \ref{fig:metricdescription_complex_LA} - \ref{fig:metricdescription_complex_apex_OCQ} show complex instructions for various metrics.

\begin{figure*}[t]
\centering
\begin{promptbox}
``name": ``linguistic\_acceptability", \\

\noindent ``description": ``Linguistic acceptability pertains to the degree to which a given language structure (e.g., phrase, sentence, discourse) aligns with the implicit norms and rules of a native speaker's linguistic intuition. In the study of language, it's distinct from 'grammaticality', which is a stricter and narrower concept based on the prescriptive rules of a language. Linguistic acceptability, on the other hand, captures broader native-speaker intuitions and encompasses factors like fluency, idiomacy, and appropriateness in context. In the context of language models, evaluating linguistic acceptability involves assessing the output of the model not just for its adherence to grammar rules, but for its overall fit within the natural, expected, and intuitive contours of fluent human language. The scoring rubric is described below, with a few possible reasons (which might not be exhaustive) for a given score.", 

\begin{minted}{json}
"scoring": {
    "0": {
        "(a)": "Sentences that lack clear syntactic structure.",
        "(b)": "Usage of non-existent or incorrect words.",
        "(c)": "Grossly inappropriate word choices for a given context."
    },
    "1": {
        "(a)": "Overly verbose or stilted phrasing.",
        "(b)": "Minor grammatical errors that do not impede understanding.",
        "(c)": "Use of a word that's technically correct but not the most appropriate for context."
    },
    "2": {
        "(a)": "Seamless integration of contextually relevant vocabulary",
        "(b)": "Effective use of idiomatic expressions without sounding forced.",
        "(c)": "Sentences that reflect natural rhythm, emphasis, and intonation of spoken language."
    }
}
\end{minted}
\end{promptbox}
\caption{Metric description for complex instructions (Linguistic Acceptability).}
\label{fig:metricdescription_complex_LA}
\end{figure*}

\begin{figure*}[h]
\centering
\begin{promptbox}
``name": ``problematic\_content", \\

\noindent ``description": ``Problematic Content assesses the presence of any content that may be deemed offensive, inappropriate, or harmful in the model's output. This metric serves as a filter against outputs that might perpetuate harmful stereotypes, misinformation, or any form of bigotry. This metric highlights the importance of ethical considerations in outputs and underscores the need for vigilance against unintended harmful content. The scoring rubric is described below, with a few possible reasons (which might not be exhaustive) for a given score.", 

\begin{minted}{json}
"scoring": {
    "1": {
        "(a)": "The output is free from any content that might be perceived as offensive, harmful, or inappropriate.",
        "(b)": "No propagation of stereotypes, misinformation, or inappropriate assertions.",
        "(c)": "The response respects cultural, social, and individual sensitivities."
    },
    "0": {
        "(a)": "The output contains statements, suggestions, or themes that can be deemed problematic or offensive.",
        "(b)": "There is evidence of perpetuation of harmful stereotypes or biases.",
        "(c)": "Contains misinformation or promotes inappropriate or harmful narratives."
    }
}
\end{minted}
\end{promptbox}
\caption{Metric description for complex instructions (Problematic content).}
\label{fig:metricdescription_complex_apex_PC}
\end{figure*}

\begin{figure*}[h]
\centering
\begin{promptbox}
``name": ``task quality", \\

\noindent ``description": ``Task Quality gauges the degree to which a model adheres to and executes the specific directives given in the prompt. This metric zeroes in exclusively on the fidelity of the model's response to the prompt's instructions. An ideal response not only recognizes the overt commands of the prompt but also respects its nuance and subtleties. The scoring rubric is described below, with a few possible reasons (which might not be exhaustive) for a given score."

\begin{minted}{json}
"scoring": {
    "0": {
        "(a)": "The model disregards the instructions entirely.",
        "(b)": "The output is entirely irrelevant to the prompt.",
        "(c)": "There is a clear disconnect between the user's request and the model's response."
    },
    "1": {
        "(a)": "The model grasps and addresses the main theme or element of the instruction but may miss out on finer details or nuances.",
        "(b)": "There is partial alignment with the prompt, indicating some elements of relevance, but not a complete match.",
        "(c)": "The response might include extraneous details not asked for, or it might omit some requested specifics."
    },
    "2": {
        "(a)": "The model demonstrates a precise understanding and adherence to the prompt's instructions.",
        "(b)": "The output holistically satisfies all aspects of the given directive without any deviation.",
        "(c)": "There's a clear and direct correlation between the user's instruction and the model's response, with no aspect of the 
               instruction left unaddressed."
    }
}
\end{minted}
\end{promptbox}
\caption{Metric description for complex instructions (task quality).}
\label{fig:metricdescription_complex_apex_TQ}
\end{figure*}

\begin{figure*}[h]
\centering
\begin{promptbox}
``name": ``output content quality", \\

\noindent ``description": ``Output Content Quality measures the overall caliber of the content generated, factoring in its relevance, clarity, originality, and linguistic fluency. High-quality output should not only be grammatically sound but should also convey information in an articulate, coherent, and engaging manner without any evidence of plagiarism, redundancy, or artificiality. This metric ensures that the produced content meets the expectations of originality, clarity, and contextual relevance in addition to linguistic fluency. The scoring rubric is described below, with a few possible reasons (which might not be exhaustive) for a given score.",

\begin{minted}{json}
"scoring": {
    "0": {
        "(a)": "The output is in a language different from the intended/requested one.",
        "(b)": "Content appears scraped from the web, giving a plagiarized feel.",
        "(c)": "The output is repetitive or overly redundant.",
        "(d)": "Displays artifacts of poor machine translation."
    },
    "1": {
        "(a)": "The content is generally accurate in terms of grammar and word choice.",
        "(b)": "Sounds unnatural or awkward in the language, lacking smoothness.",
        "(c)": "May have minor discrepancies in content clarity or relevance.",
        "(d)": "Shows traces of generative patterns or repetitiveness, albeit less pronounced than level 0."
    },
    "2": {
        "(a)": "The text shows a high level of originality and authenticity.",
        "(b)": "Demonstrates clear, coherent, and contextually appropriate content.",
        "(c)": "Engages the reader with natural linguistic flow and rhythm.",
        "(d)": "Absence of any noticeable generative artifacts or awkward."
    }
}
\end{minted}
\end{promptbox}
\caption{Metric description for complex instructions (Output content quality).}
\label{fig:metricdescription_complex_apex_OCQ}
\end{figure*}
\subsection{Fleiss' Kappa}
 Table \ref{tab:kappa} shows the Fleiss' Kappa ($\kappa$) on the full dataset for various annotator combinations, aggregated by language, task, and metrics.
\label{sec:kappa}

\begin{table*}[h]
\centering
\small
\begin{tabular}{@{}llcccc@{}}
\toprule
\textit{} & \textit{Name} & \begin{tabular}[c]{@{}c@{}}Annot1\\ Annot2\\ Annot3\end{tabular} & \begin{tabular}[c]{@{}c@{}}AnnotAgg\\ GPT4\_joint\end{tabular} & \begin{tabular}[c]{@{}c@{}}AnnotAgg\\ GPT4\_single\end{tabular} & \begin{tabular}[c]{@{}c@{}}AnnotAgg\\ GPT4\_SD\end{tabular} \\ \midrule
\multicolumn{1}{c}{} & \textit{Cs} & 0.46 $\pm$ 0.29 & 0.05 $\pm$ 0.12 & 0.08 $\pm$ 0.17 & 0.07 $\pm$ 0.15 \\
\multicolumn{1}{c}{} & \textit{De} & 0.29 $\pm$ 0.29 & 0.07 $\pm$ 0.11 & 0.13 $\pm$ 0.16 & 0.13 $\pm$ 0.15 \\
\multicolumn{1}{c}{} & \textit{En} & 0.47 $\pm$ 0.42 & 0.15 $\pm$ 0.22 & 0.18 $\pm$ 0.24 & 0.11 $\pm$ 0.17 \\
\multicolumn{1}{c}{} & \textit{Es} & 0.32 $\pm$ 0.22 & 0.04 $\pm$ 0.11 & 0.04 $\pm$ 0.12 & 0.04 $\pm$ 0.11 \\
\textit{Lang.} & \textit{Fr} & 0.44 $\pm$ 0.31 & 0.12 $\pm$ 0.21 & 0.20 $\pm$ 0.23 & 0.22 $\pm$ 0.22 \\
\multicolumn{1}{c}{} & \textit{It} & 0.41 $\pm$ 0.33 & 0.06 $\pm$ 0.11 & 0.08 $\pm$ 0.16 & 0.08 $\pm$ 0.14 \\
\multicolumn{1}{c}{} & \textit{Ja} & 0.44 $\pm$ 0.33 & 0.01 $\pm$ 0.13 & 0.02 $\pm$ 0.14 & 0.04 $\pm$ 0.15 \\
\multicolumn{1}{c}{} & \textit{Pt-Br} & 0.52 $\pm$ 0.37 & 0.11 $\pm$ 0.19 & 0.09 $\pm$ 0.17 & 0.12 $\pm$ 0.20 \\
\multicolumn{1}{c}{} & \textit{Zh} & 0.35 $\pm$ 0.32 & 0.00 $\pm$ 0.08 & 0.01 $\pm$ 0.07 & 0.02 $\pm$ 0.07 \\ \midrule
 & \textit{H} & 0.40 $\pm$ 0.39 & 0.04 $\pm$ 0.15 & 0.05 $\pm$ 0.15 & 0.08 $\pm$ 0.18 \\
 & \textit{LA} & 0.41 $\pm$ 0.24 & -0.02 $\pm$ 0.06 & 0.05 $\pm$ 0.15 & 0.09 $\pm$ 0.16 \\
 \textit{Metric} & \textit{OCQ} & 0.54 $\pm$ 0.19 & 0.13 $\pm$ 0.17 & 0.16 $\pm$ 0.19 & 0.14 $\pm$ 0.17 \\
 & \textit{PC} & 0.11 $\pm$ 0.32 & 0.00 $\pm$ 0.00 & 0.00 $\pm$ 0.00 & 0.00 $\pm$ 0.00 \\
 & \textit{TQ} & 0.60 $\pm$ 0.20 & 0.18 $\pm$ 0.19 & 0.20 $\pm$ 0.21 & 0.16 $\pm$ 0.18 \\ \midrule
 & \textit{\begin{tabular}[c]{@{}l@{}}Continue \\ Writing\end{tabular}} & 0.45 $\pm$ 0.33 & 0.06 $\pm$ 0.15 & 0.07 $\pm$ 0.17 & 0.08 $\pm$ 0.16 \\
 \textit{Task} & \textit{\begin{tabular}[c]{@{}l@{}}Open \\ Prompt\end{tabular}} & 0.49 $\pm$ 0.32 & 0.12 $\pm$ 0.19 & 0.16 $\pm$ 0.19 & 0.15 $\pm$ 0.18 \\
 & \textit{Summarize} & 0.29 $\pm$ 0.29 & 0.02 $\pm$ 0.09 & 0.06 $\pm$ 0.15 & 0.05 $\pm$ 0.13 \\ \bottomrule
\end{tabular}
\caption{Fleiss' Kappa ($\kappa$) values for different cases and annotator combinations on the full dataset. GPT4\_SD means GPT4\_single\_detailed}.
\label{tab:kappa}
\end{table*}
\subsection{Class distribution for Metrics with 3 classes}
\label{sec:apex_3class}

\begin{figure*}[h]
    \centering
    \begin{subfigure}[t]{0.33\textwidth}
    \includegraphics[width=0.90\textwidth]{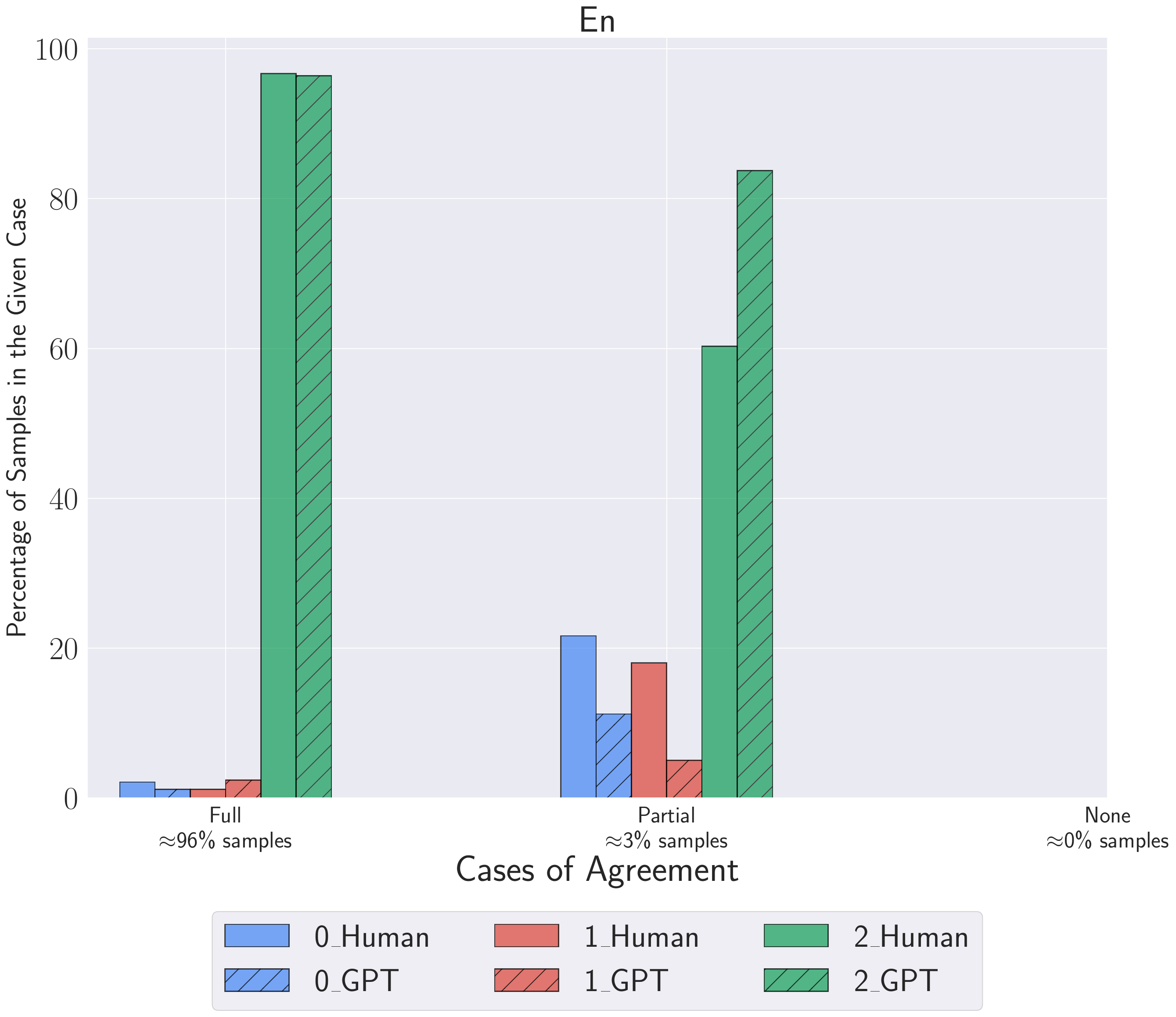}
    \caption{Compound call - English}
    \label{fig:joint_english}
    \end{subfigure}
    \begin{subfigure}[t]{0.33\textwidth}
    \includegraphics[width=0.90\textwidth]{figures/scoredist_single_language_English.pdf}
    \caption{Single Call - English}
    \label{fig:single_english}
    \end{subfigure}
    \begin{subfigure}[t]{0.33\textwidth}
    \includegraphics[width=0.90\textwidth]{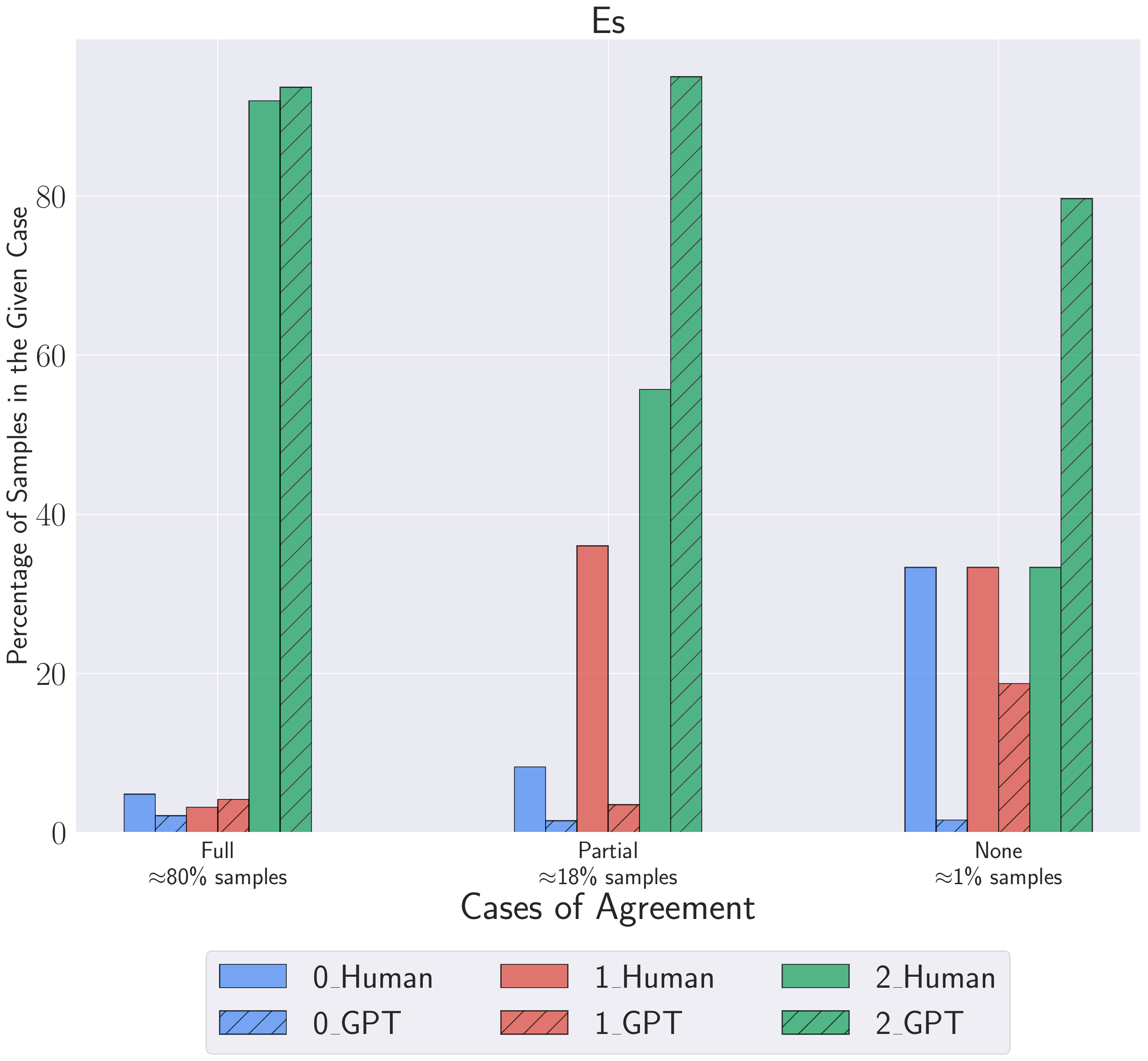}
    \caption{Compound call - Spanish}
    \label{fig:joint_spanish}
    \end{subfigure}
    \begin{subfigure}[t]{0.33\textwidth}
    \includegraphics[width=0.90\textwidth]{figures/scoredist_single_language_Spanish.pdf}
    \caption{Single Call - Spanish}
    \label{fig:single_spanish}
    \end{subfigure}
    \begin{subfigure}[t]{0.33\textwidth}
    \includegraphics[width=0.90\textwidth]{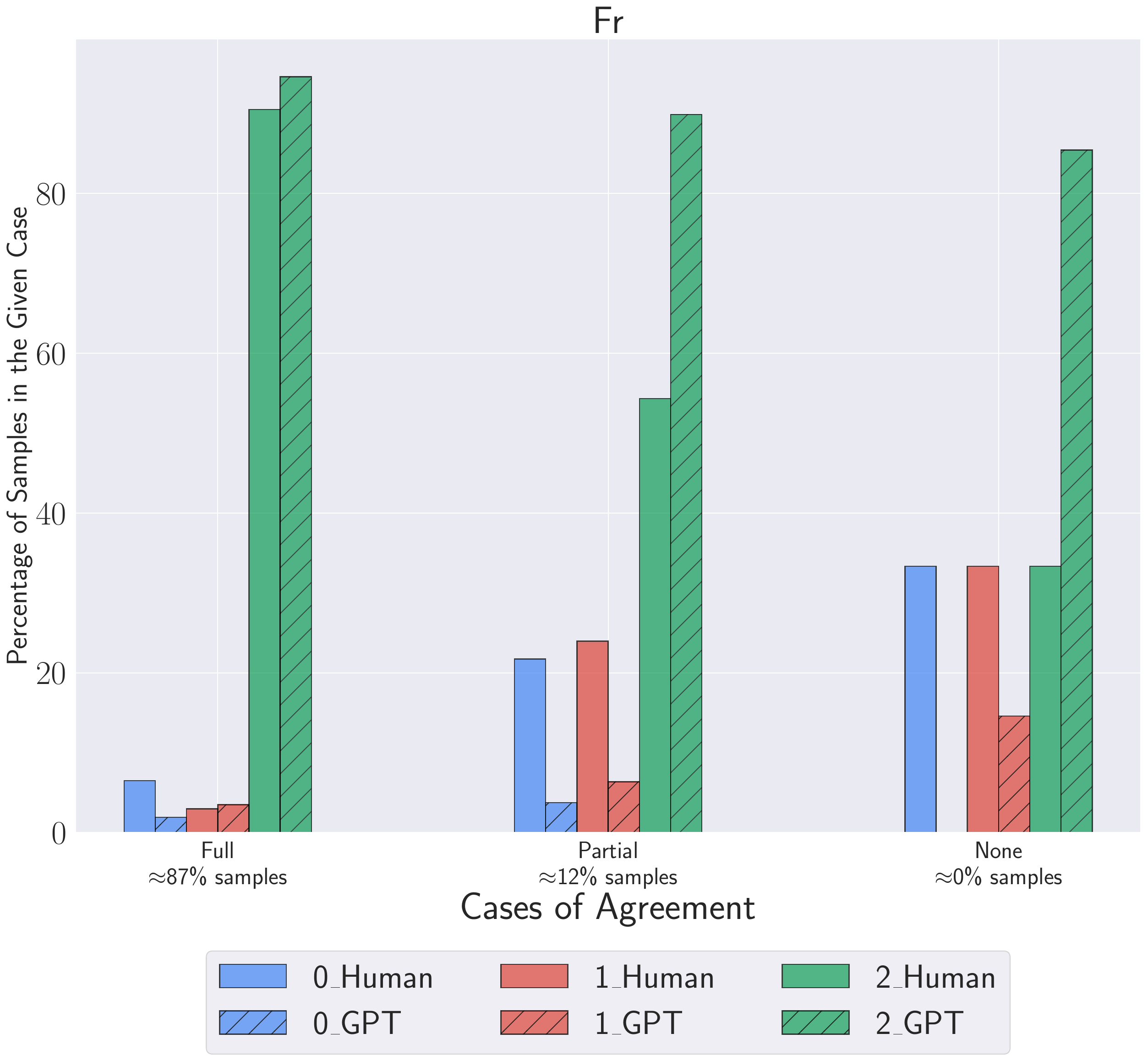}
    \caption{Compound call - French}
    \label{fig:joint_french}
    \end{subfigure}
    \begin{subfigure}[t]{0.33\textwidth}
    \includegraphics[width=0.90\textwidth]{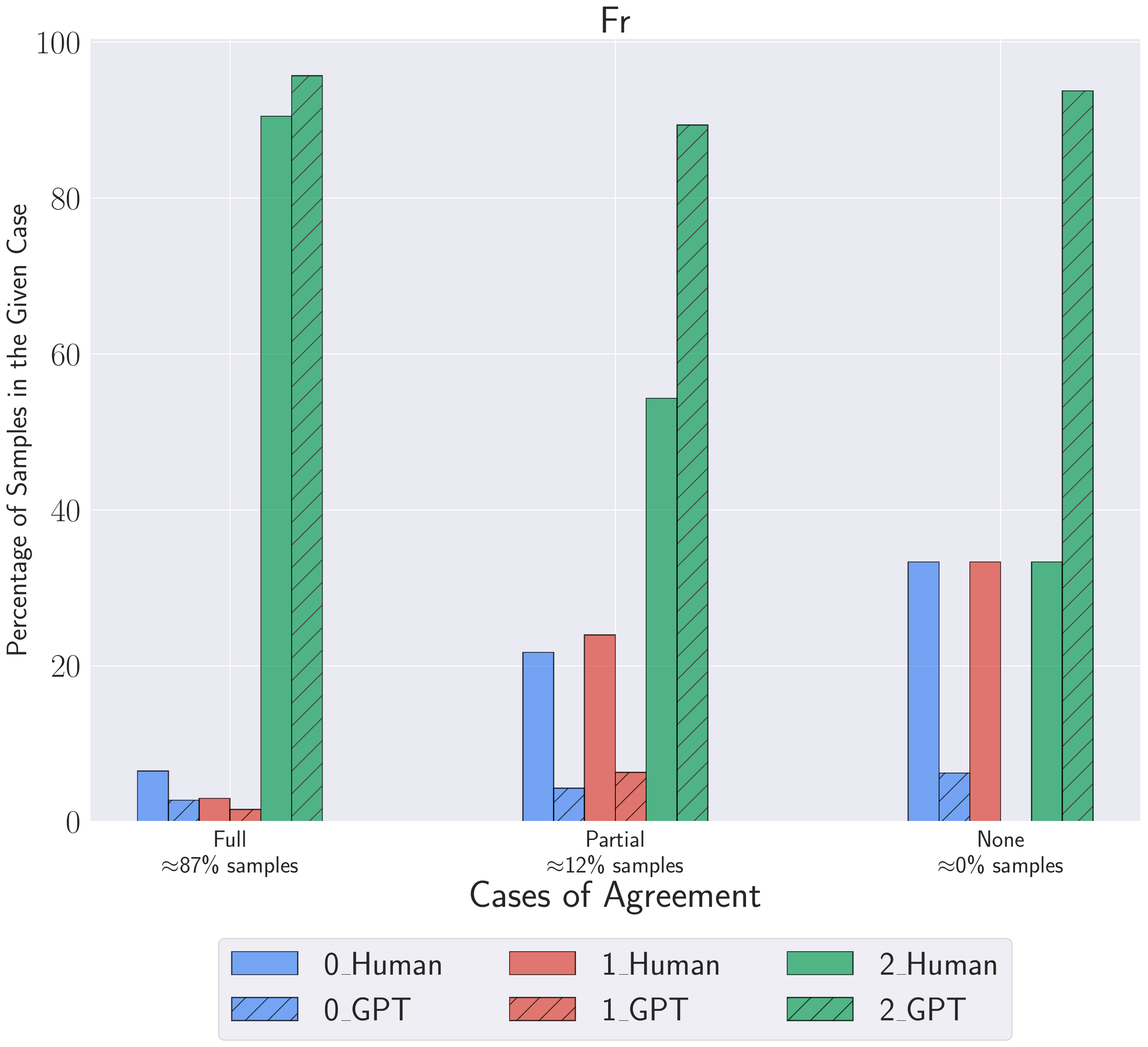}
    \caption{Single Call - French}
    \label{fig:single_french}
    \end{subfigure}
    \begin{subfigure}[t]{0.33\textwidth}
    \includegraphics[width=0.90\textwidth]{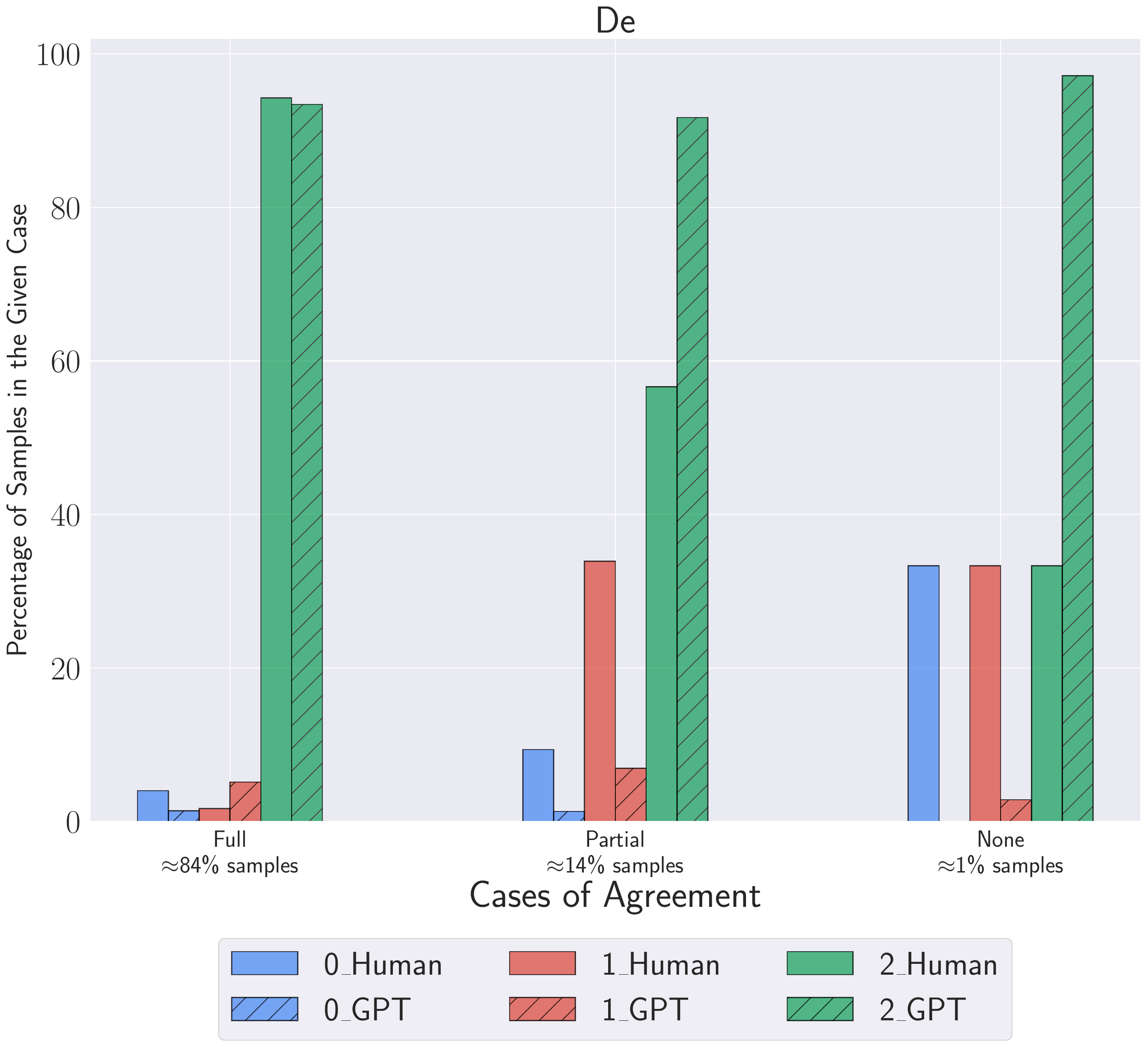}
    \caption{Compound call - German}
    \label{fig:joint_german}
    \end{subfigure}
    \begin{subfigure}[t]{0.33\textwidth}
    \includegraphics[width=0.90\textwidth]{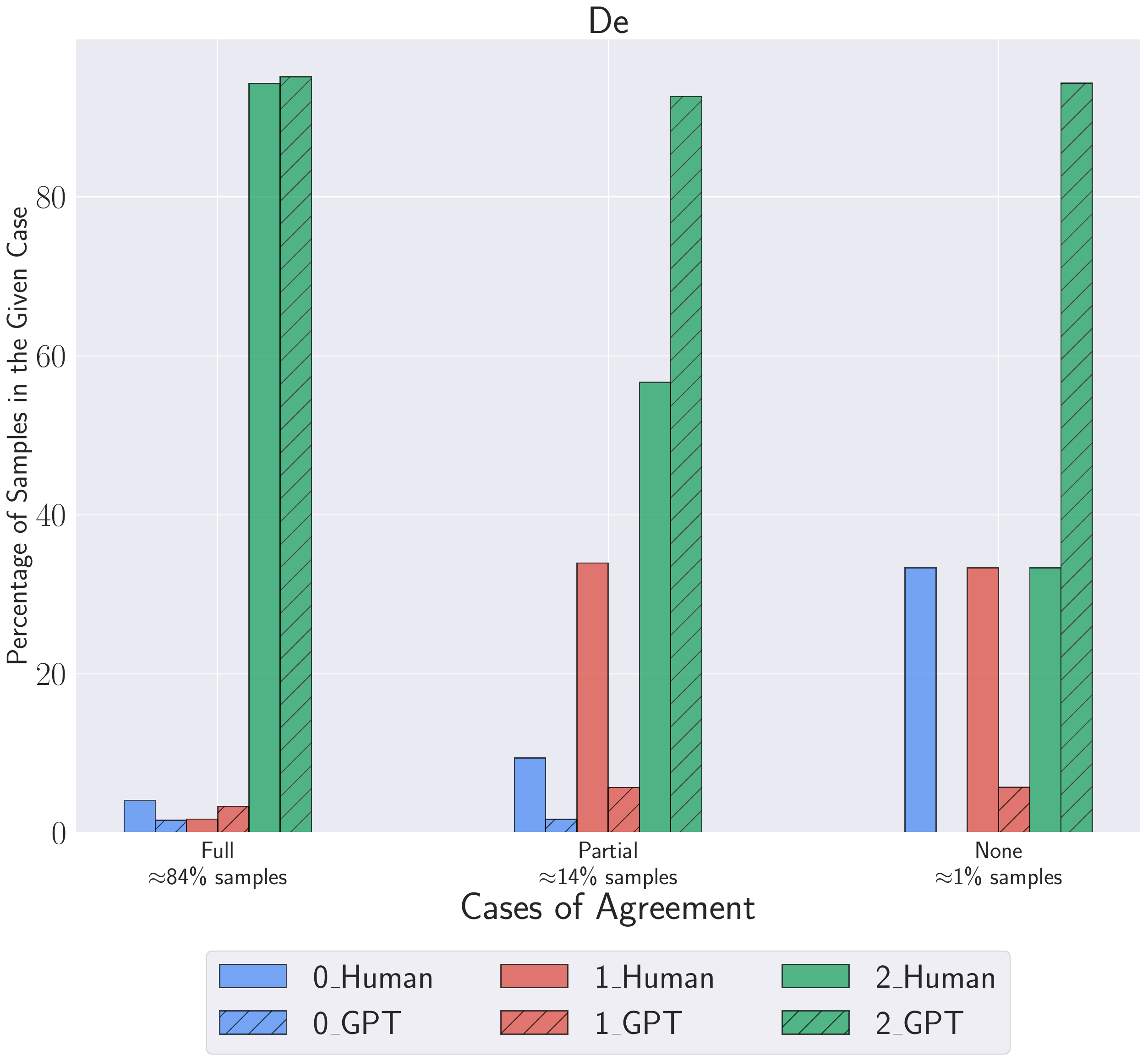}
    \caption{Single Call - German}
    \label{fig:single_german}
    \end{subfigure}
    \begin{subfigure}[t]{0.33\textwidth}
    \includegraphics[width=0.90\textwidth]{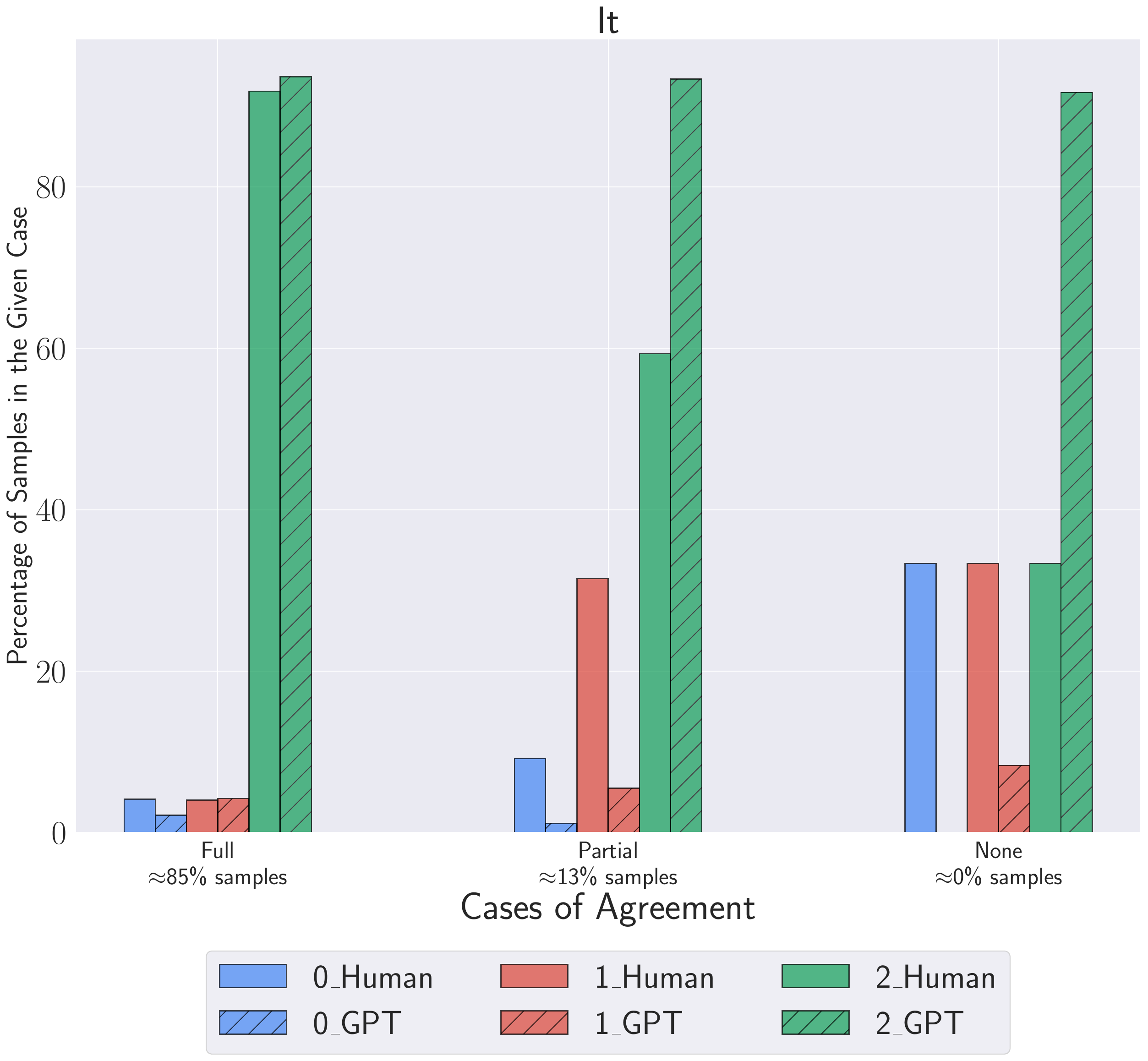}
    \caption{Compound call - Italian}
    \label{fig:joint_italian}
    \end{subfigure}
    \begin{subfigure}[t]{0.33\textwidth}
    \includegraphics[width=0.90\textwidth]{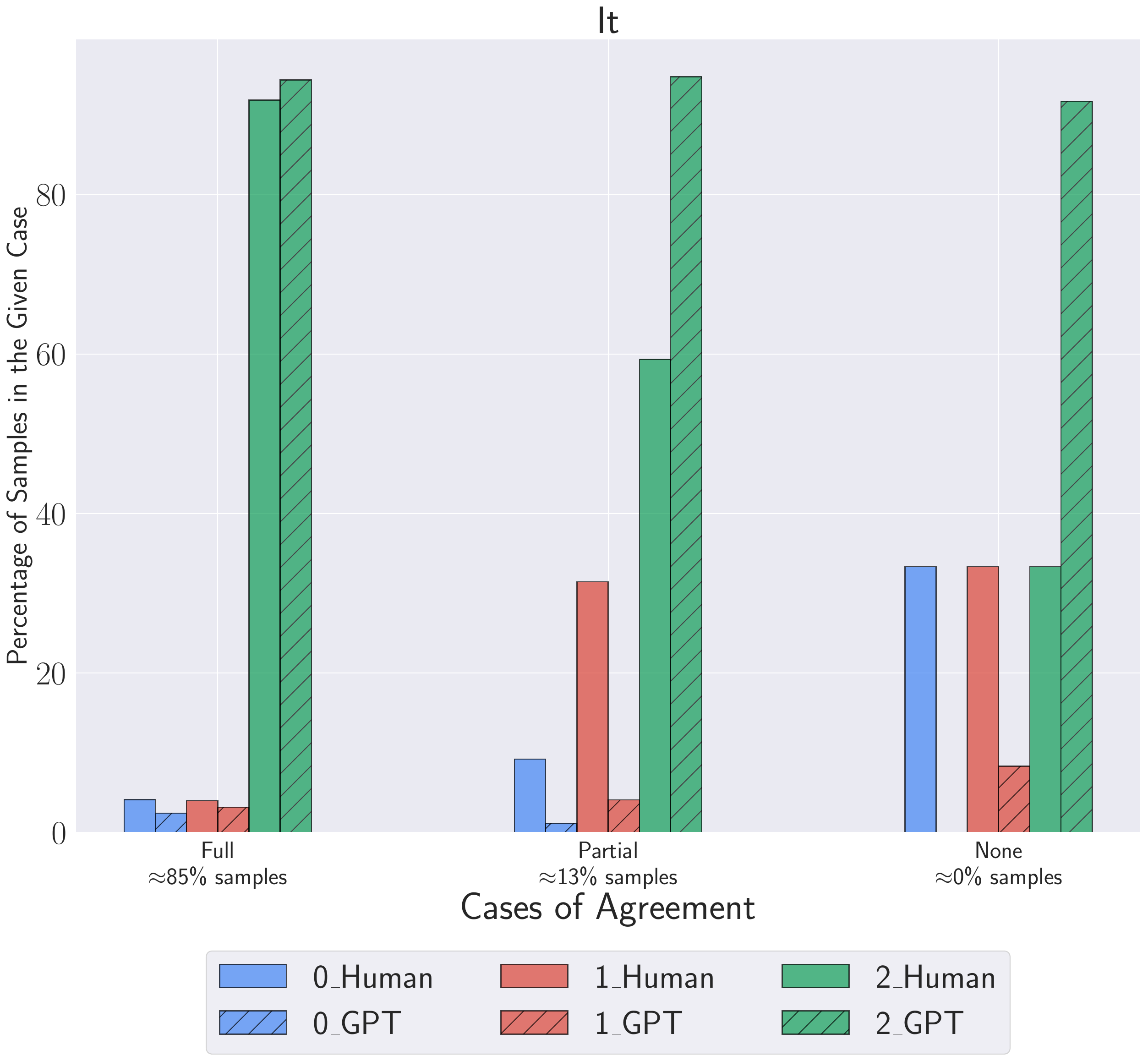}
    \caption{Single Call - Italian}
    \label{fig:single_italian}
    \end{subfigure}
    \caption{Class distribution per language (En, Es, Fr, De, It). Results are aggregated over all tasks and metrics with 3 classes (LA, OCQ, TQ).}
    \label{fig:classdist1}
\end{figure*}

\begin{figure*}[h]
    \centering
    \begin{subfigure}[t]{0.33\textwidth}
    \includegraphics[width=0.90\textwidth]{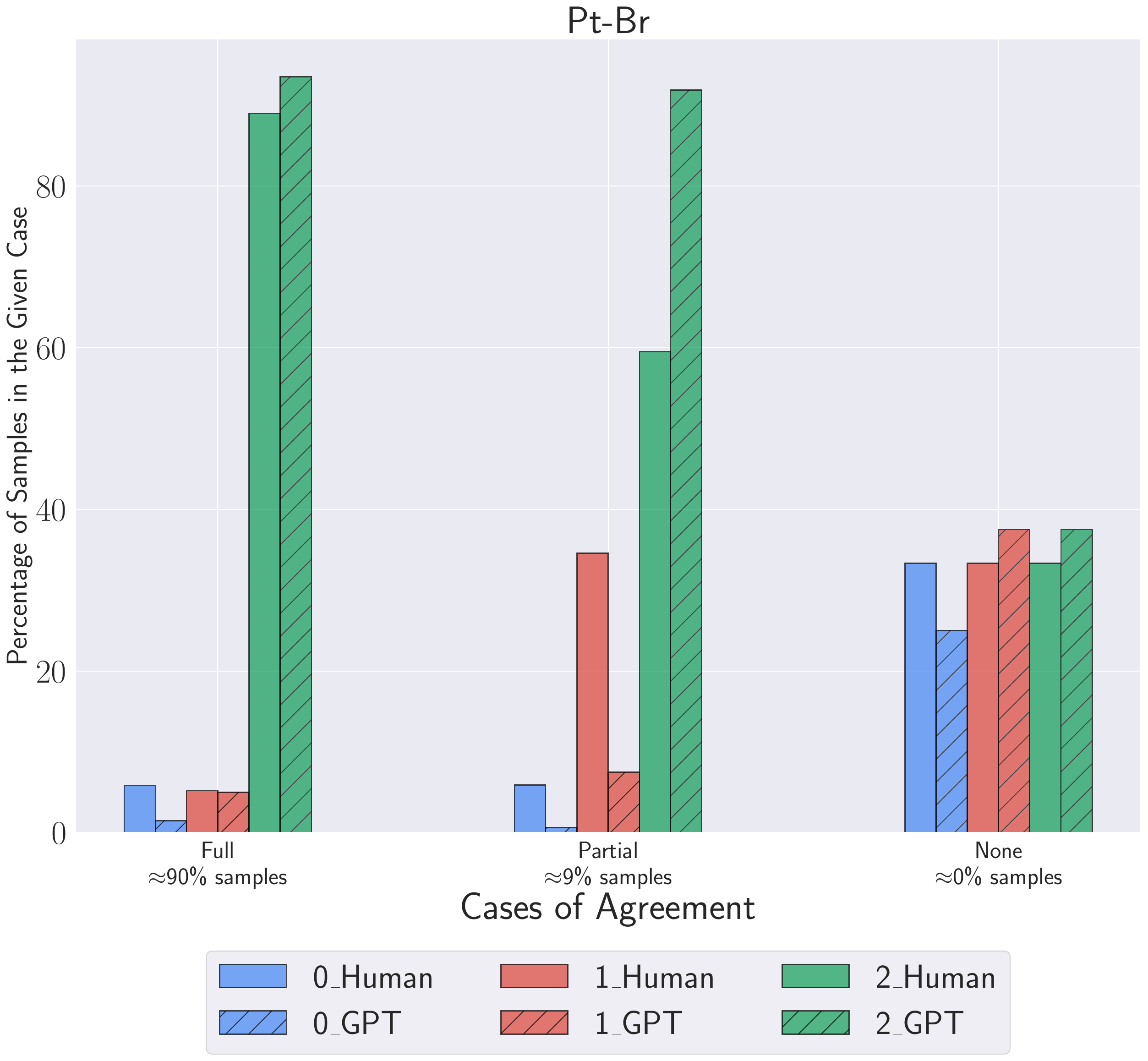}
    \caption{Compound call - Portuguese (Br)}
    \label{fig:joint_brazilian}
    \end{subfigure}
    \begin{subfigure}[t]{0.33\textwidth}
    \includegraphics[width=0.90\textwidth]{figures/scoredist_single_language_PortugueseBr.pdf}
    \caption{Single Call - Portuguese (Br)}
    \label{fig:single_brazilian}
    \end{subfigure}
    \begin{subfigure}[t]{0.33\textwidth}
    \includegraphics[width=0.90\textwidth]{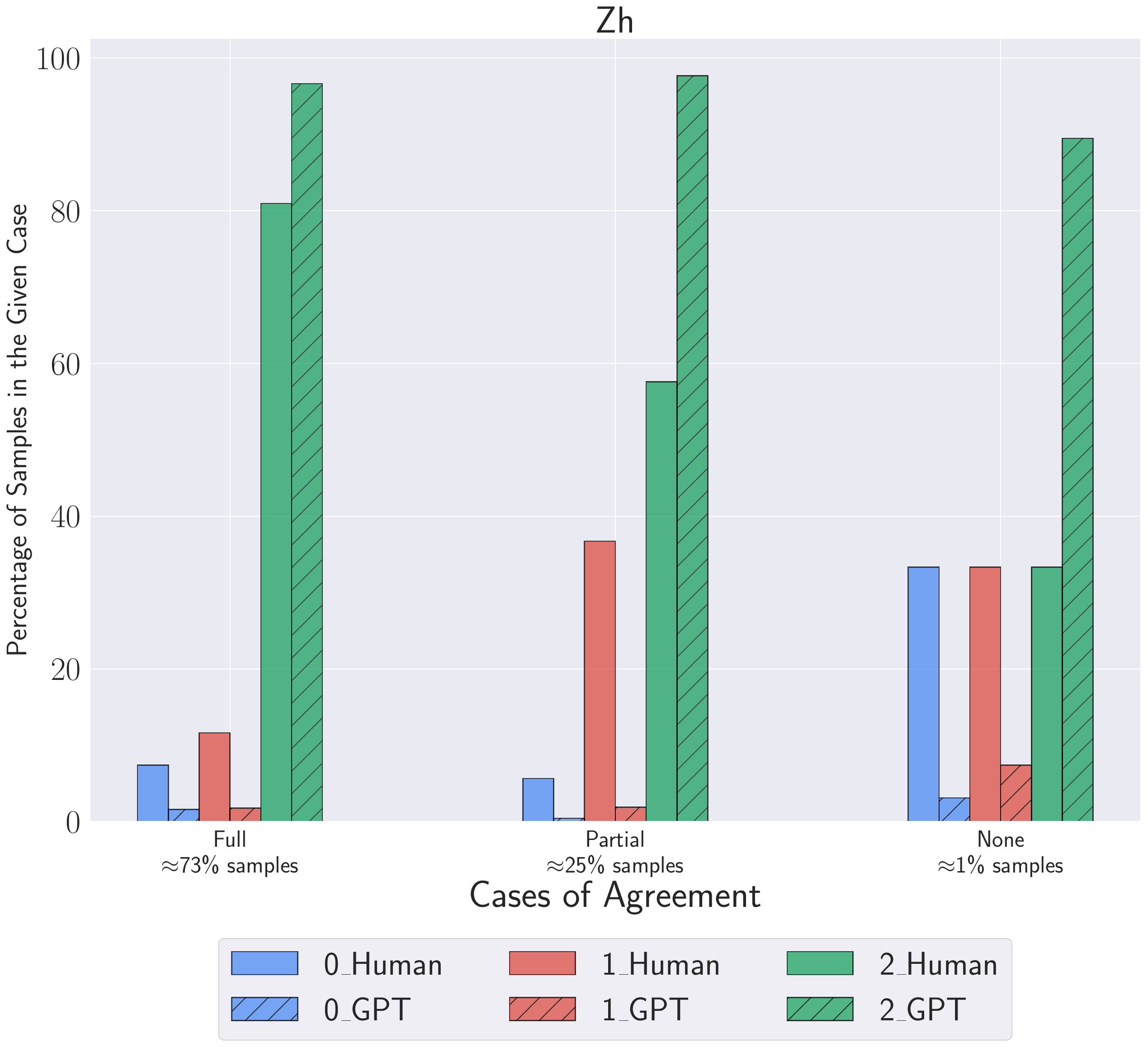}
    \caption{Compound call - Chinese}
    \label{fig:joint_chinese}
    \end{subfigure}
    \begin{subfigure}[t]{0.33\textwidth}
    \includegraphics[width=0.90\textwidth]{figures/scoredist_single_language_Chinese.pdf}
    \caption{Single Call - Chinese}
    \label{fig:single_chinese}
    \end{subfigure}
    \begin{subfigure}[t]{0.33\textwidth}
    \includegraphics[width=0.90\textwidth]{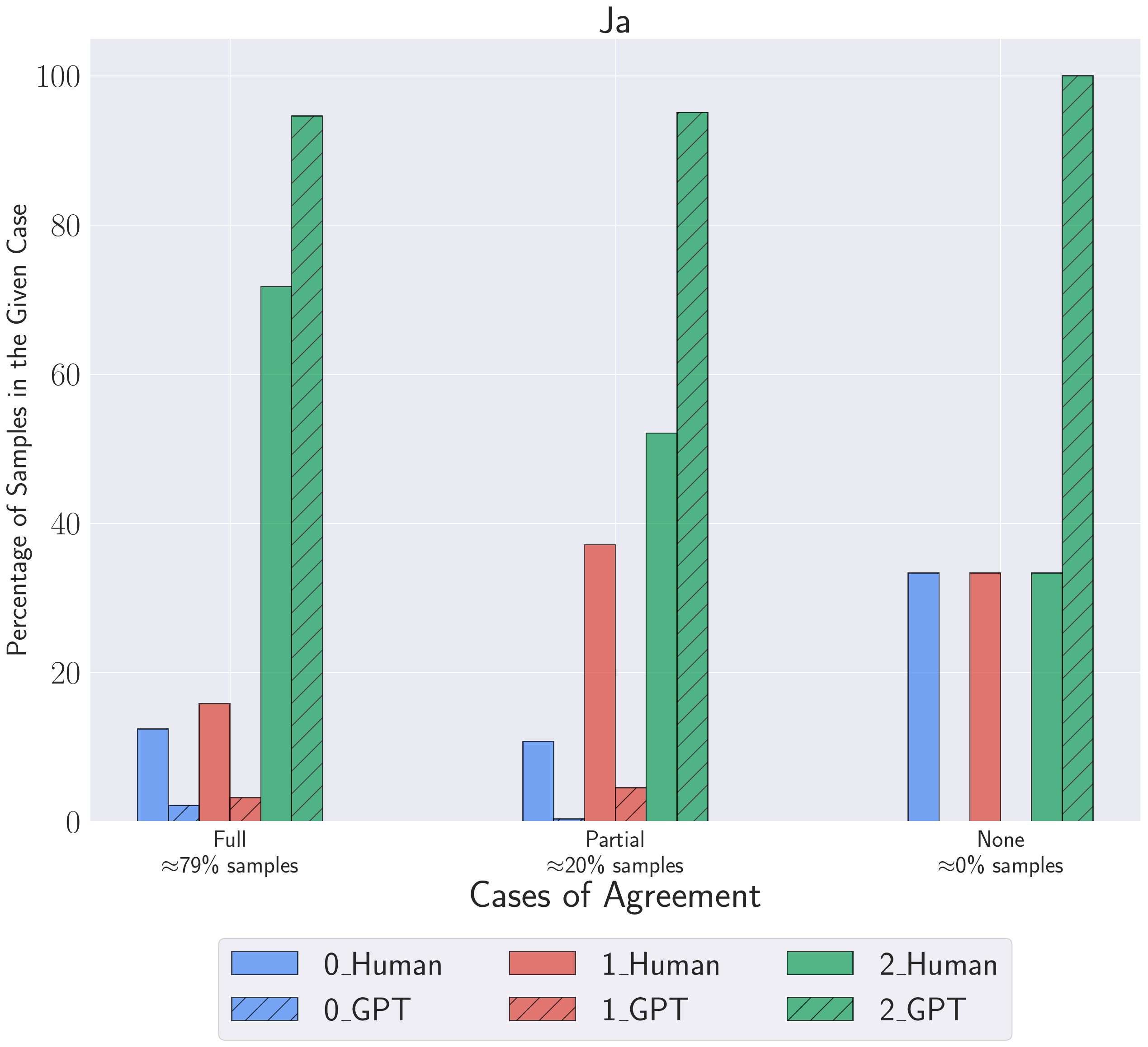}
    \caption{Compound call - Japanese}
    \label{fig:joint_japanese}
    \end{subfigure}
    \begin{subfigure}[t]{0.33\textwidth}
    \includegraphics[width=0.90\textwidth]{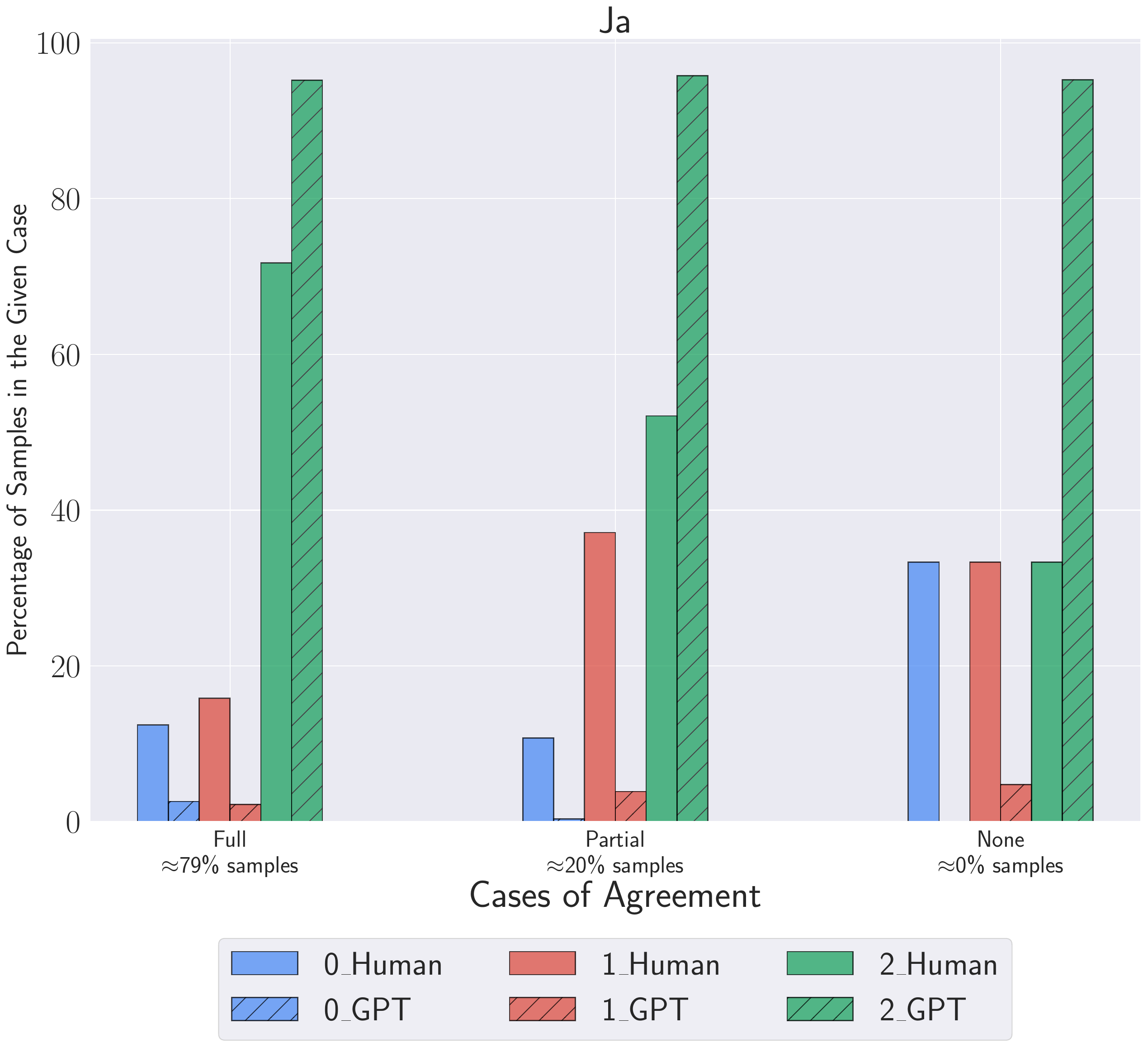}
    \caption{Single Call - Japanese}
    \label{fig:single_japanese}
    \end{subfigure}
    \begin{subfigure}[t]{0.33\textwidth}
    \includegraphics[width=0.90\textwidth]{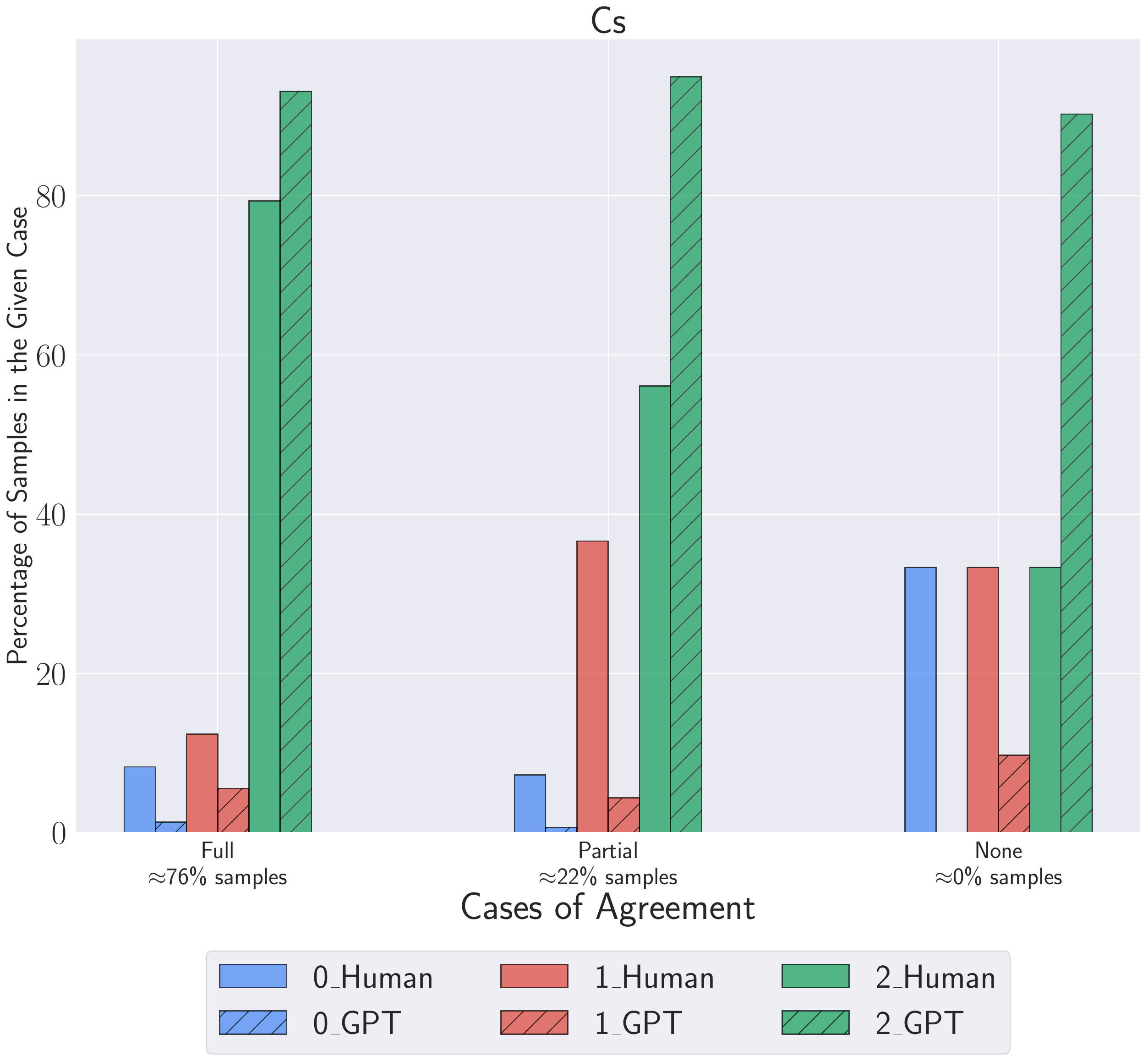}
    \caption{Compound call - Czech}
    \label{fig:joint_czech}
    \end{subfigure}
    \begin{subfigure}[t]{0.33\textwidth}
    \includegraphics[width=0.90\textwidth]{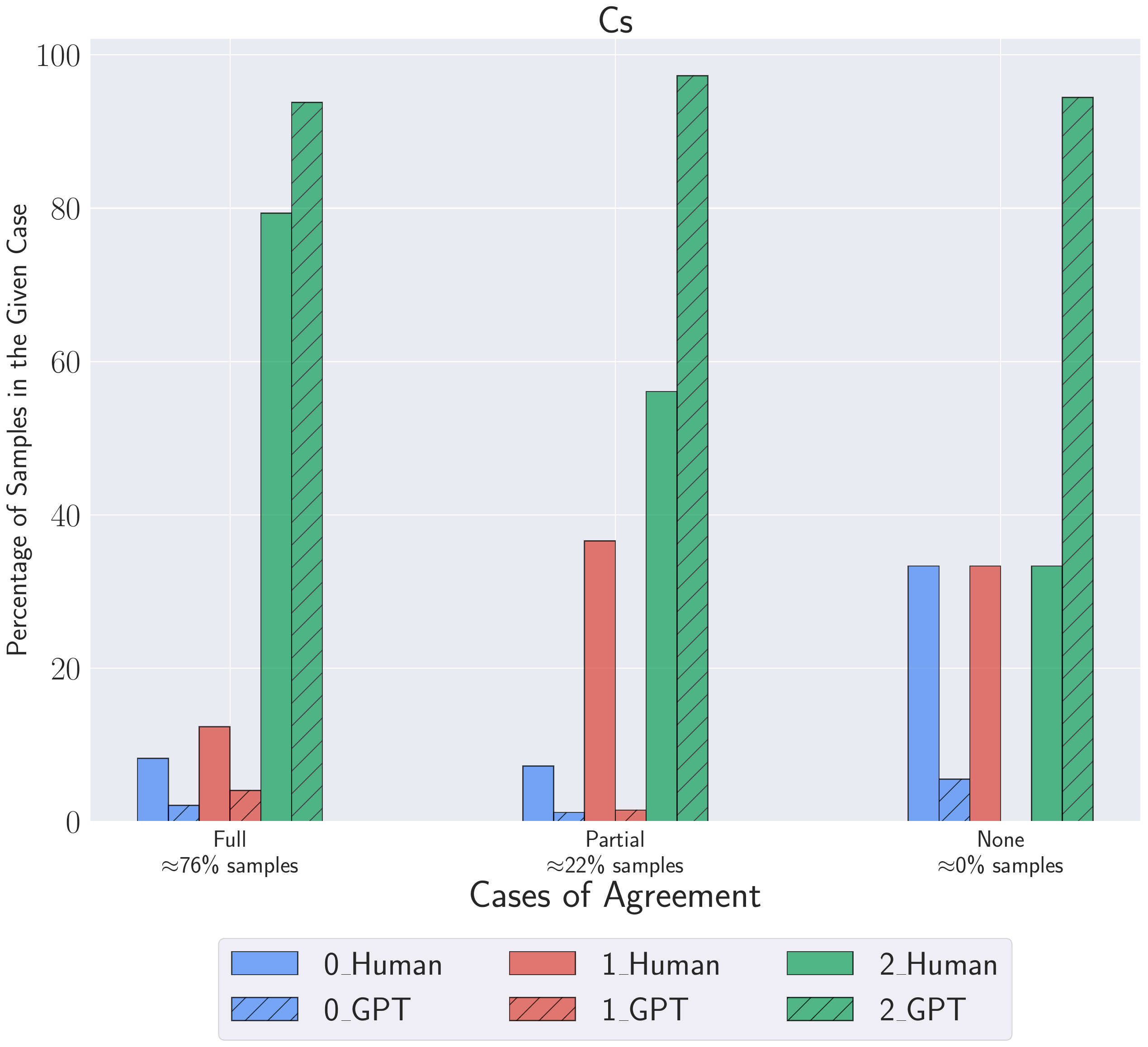}
    \caption{Single Call - Czech}
    \label{fig:single_czech}
    \end{subfigure}\\
    \caption{Class distribution per language (Pt-Br, Zh, Ja, Cz).  Results are aggregated over all tasks and metrics with $3$ classes (LA, OCQ, TQ).}
    \label{fig:classdist2}
\end{figure*}

Figures \ref{fig:classdist1} and \ref{fig:classdist2} show class distribution for various languages, aggregated over metrics with 3 classes - LA, OCQ, TQ. 

\subsection{Class distribution for Metrics with 2 classes}
\label{sec:apex_2class}
\begin{figure*}[h]
    \centering
    \begin{subfigure}[t]{0.33\textwidth}
    \includegraphics[width=0.90\textwidth]{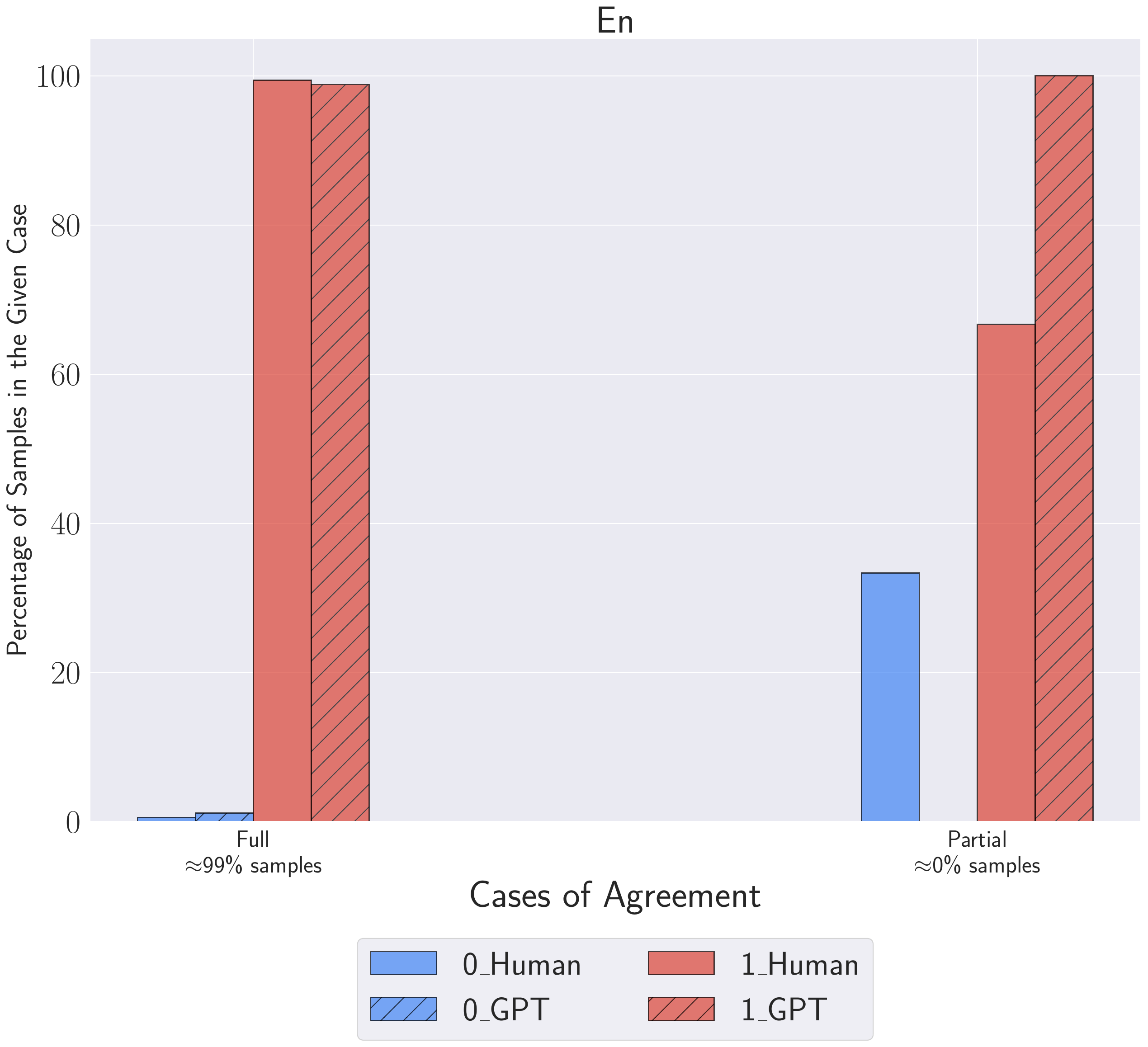}
    \caption{Compound call - English}
    \label{fig:joint_english}
    \end{subfigure}
    \begin{subfigure}[t]{0.33\textwidth}
    \includegraphics[width=0.90\textwidth]{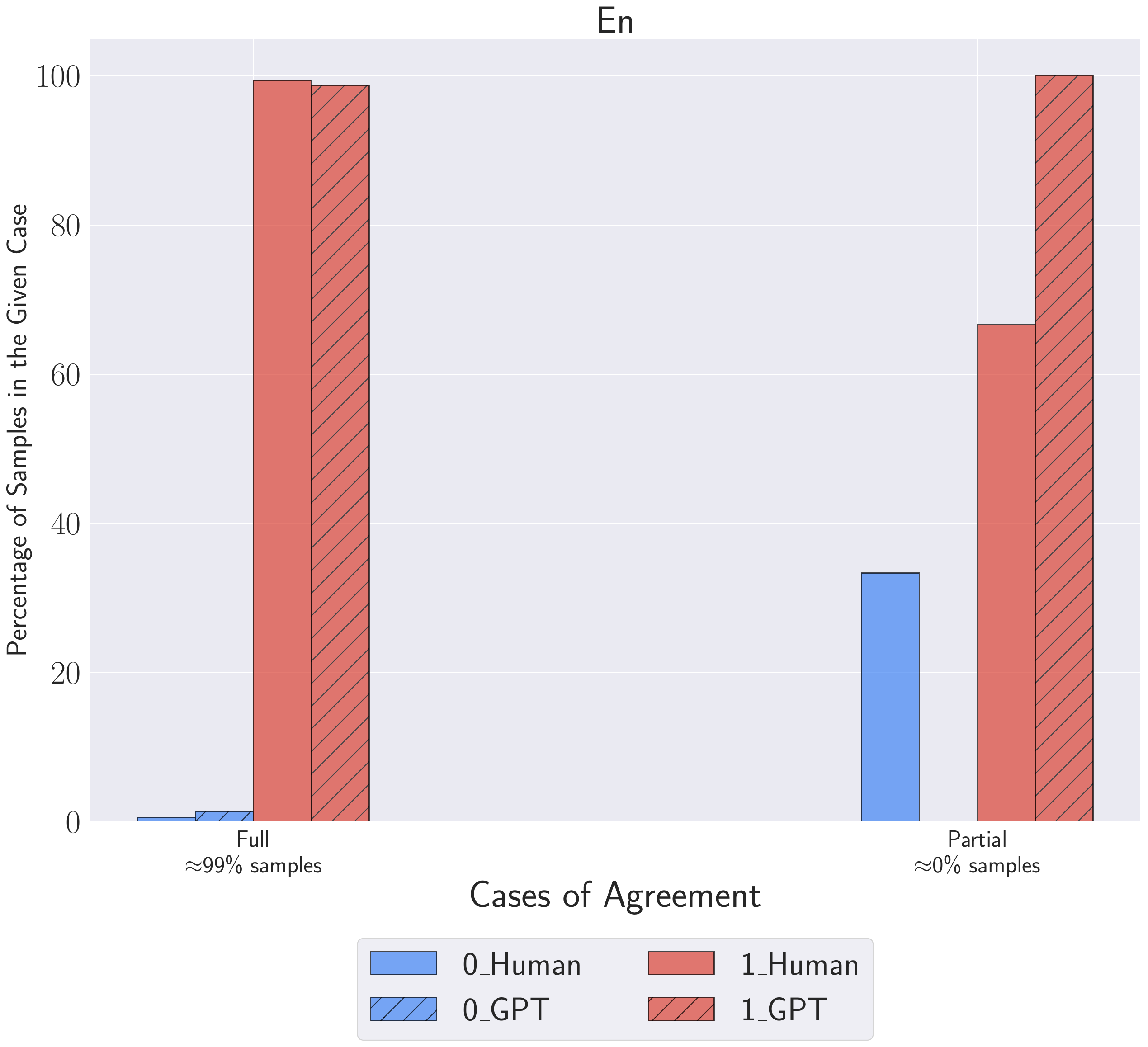}
    \caption{Single Call - English}
    \label{fig:single_english}
    \end{subfigure}
    \begin{subfigure}[t]{0.33\textwidth}
    \includegraphics[width=0.90\textwidth]{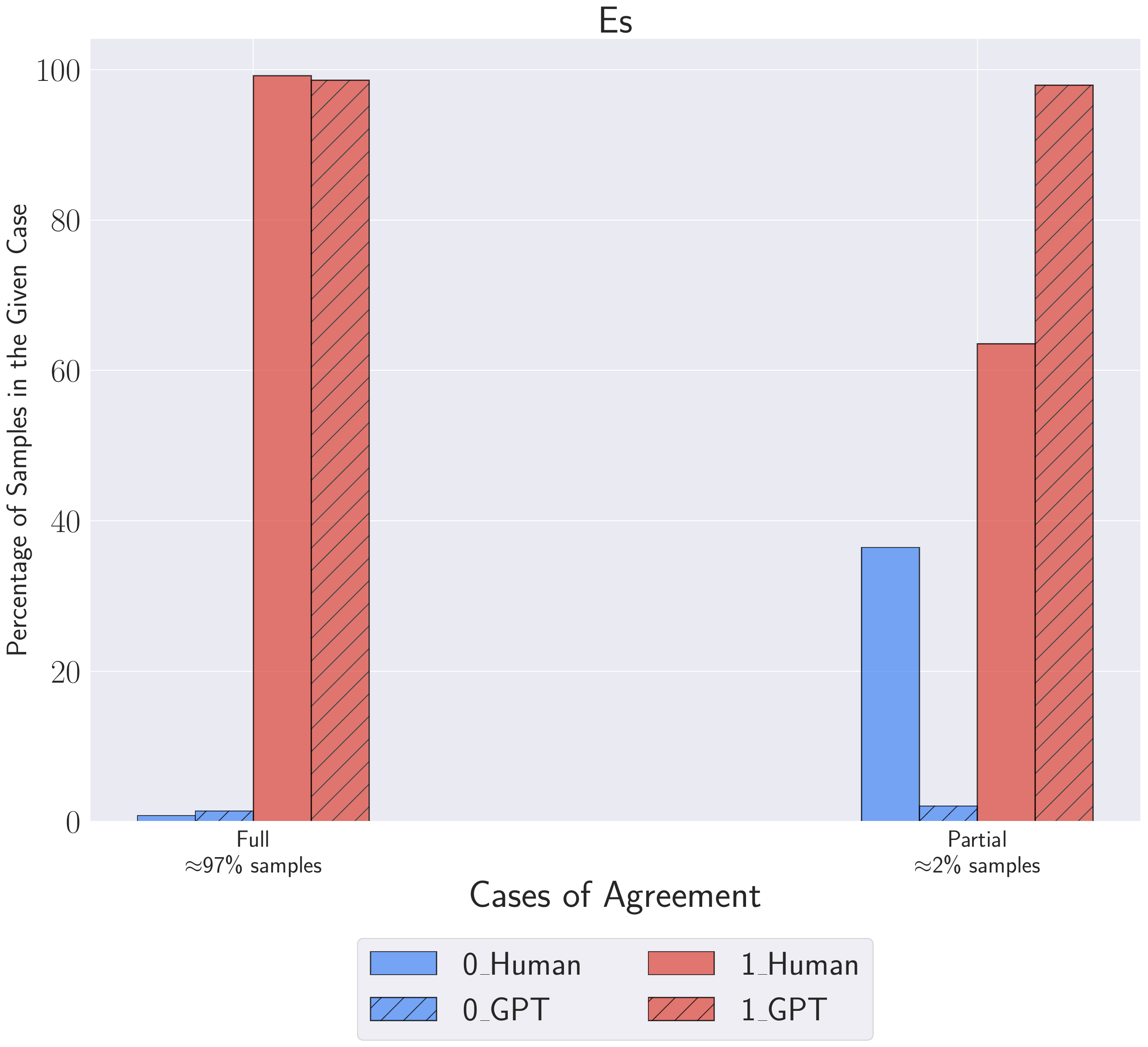}
    \caption{Compound call - Spanish}
    \label{fig:joint_spanish}
    \end{subfigure}
    \begin{subfigure}[t]{0.33\textwidth}
    \includegraphics[width=0.90\textwidth]{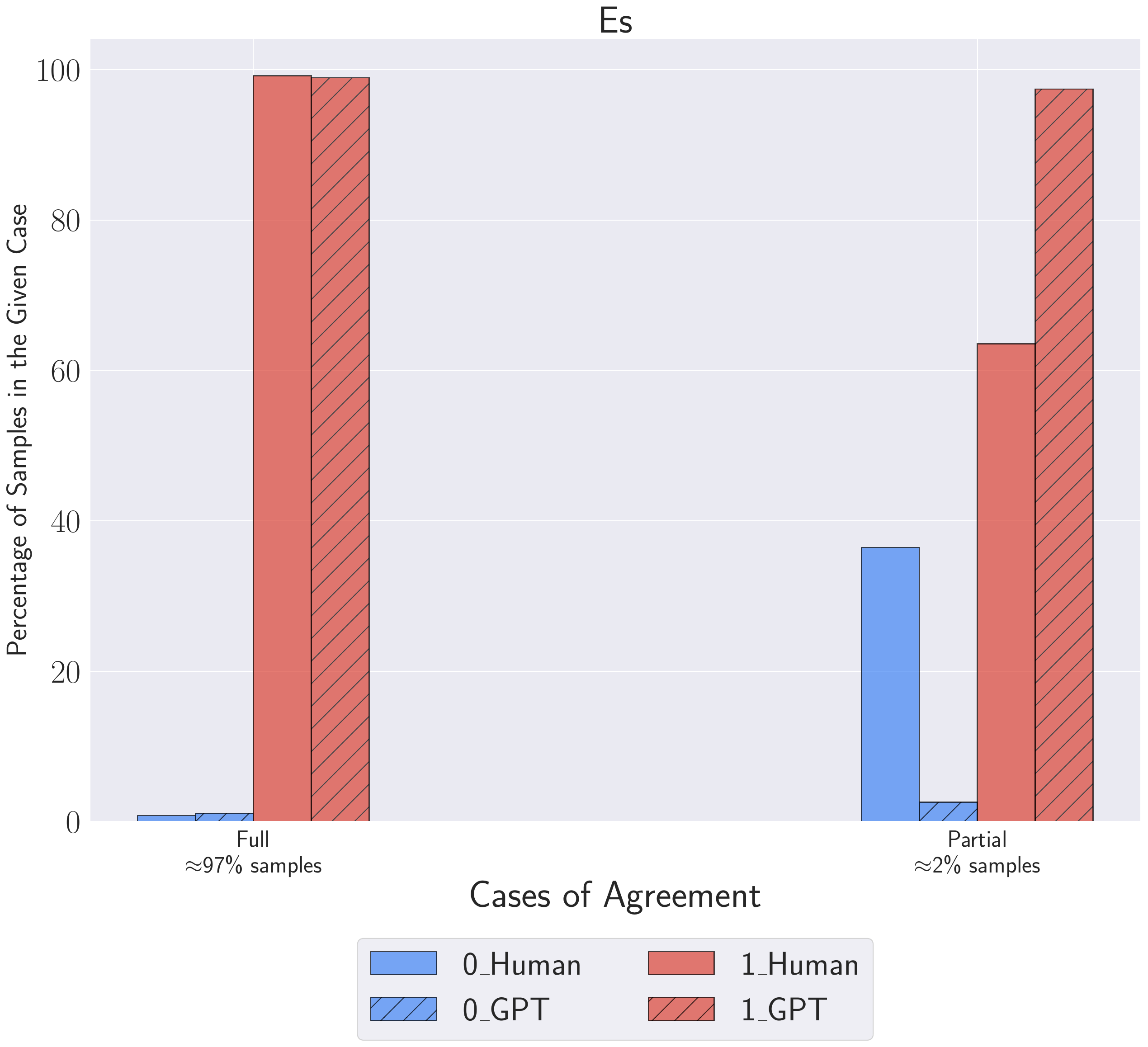}
    \caption{Single Call - Spanish}
    \label{fig:single_spanish}
    \end{subfigure}
    \begin{subfigure}[t]{0.33\textwidth}
    \includegraphics[width=0.90\textwidth]{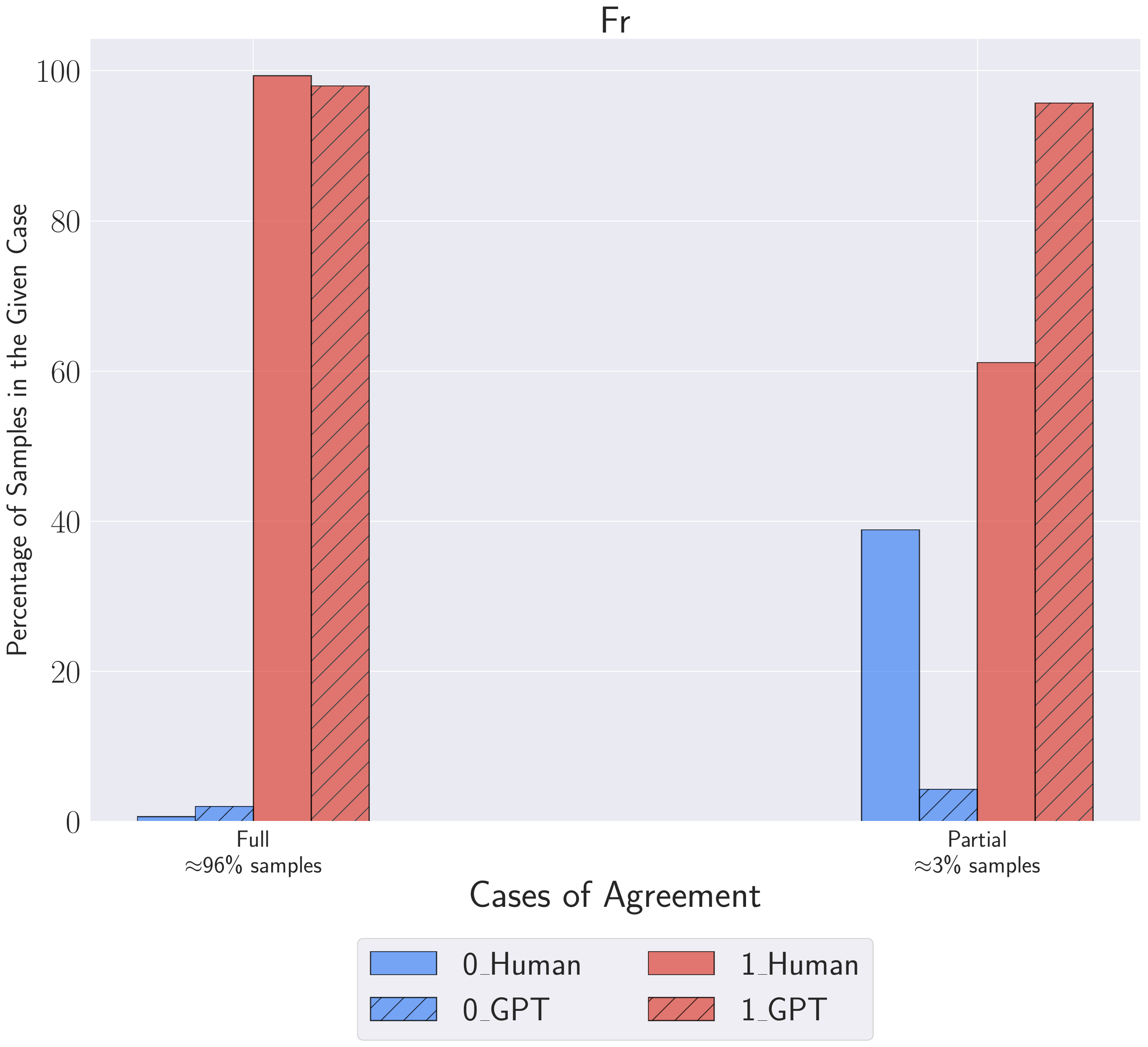}
    \caption{Compound call - French}
    \label{fig:joint_french}
       \end{subfigure}
    \begin{subfigure}[t]{0.33\textwidth}
    \includegraphics[width=0.90\textwidth]{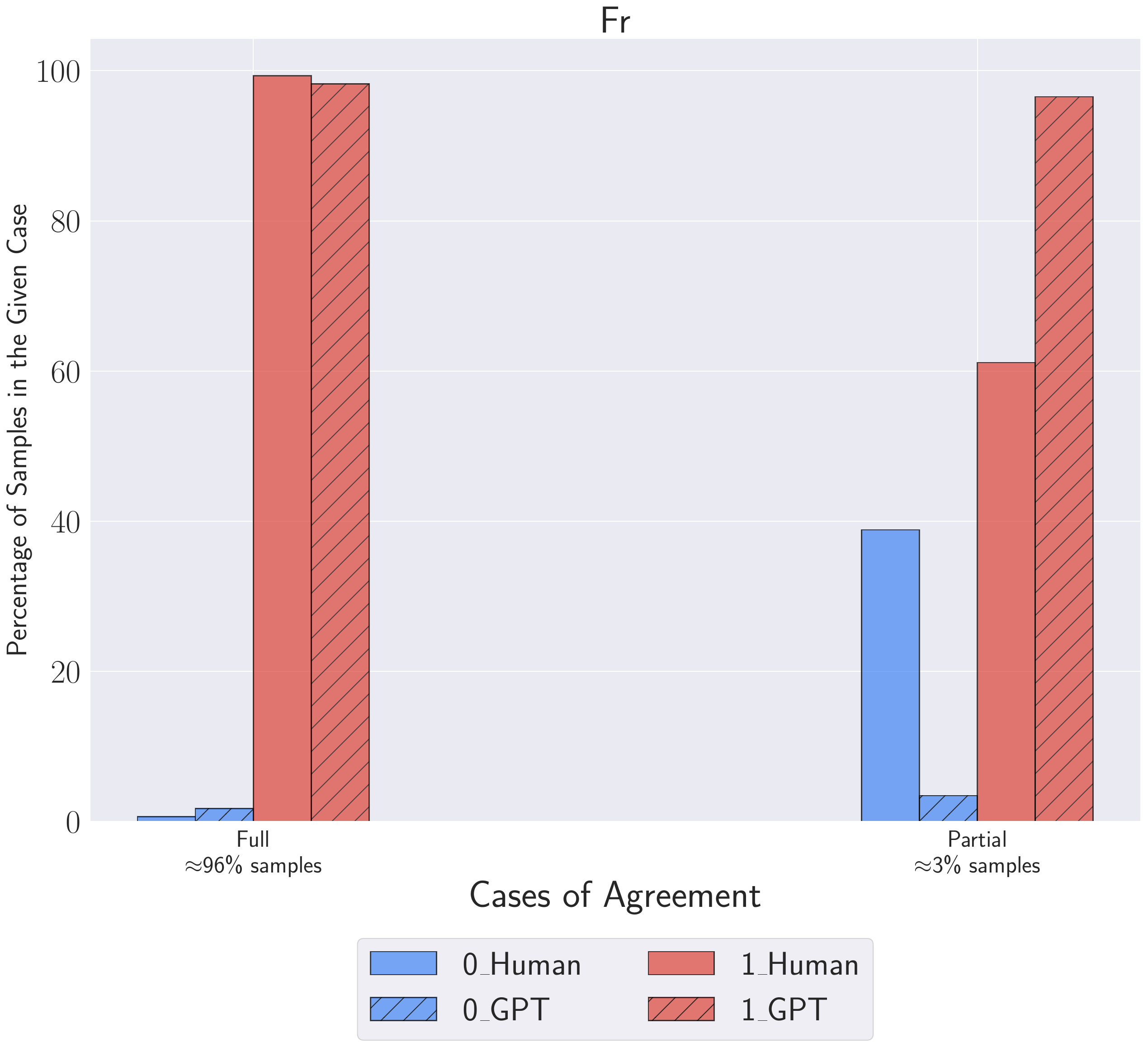}
    \caption{Single Call - French}
    \label{fig:single_french}
    \end{subfigure}
    \begin{subfigure}[t]{0.33\textwidth}
    \includegraphics[width=0.90\textwidth]{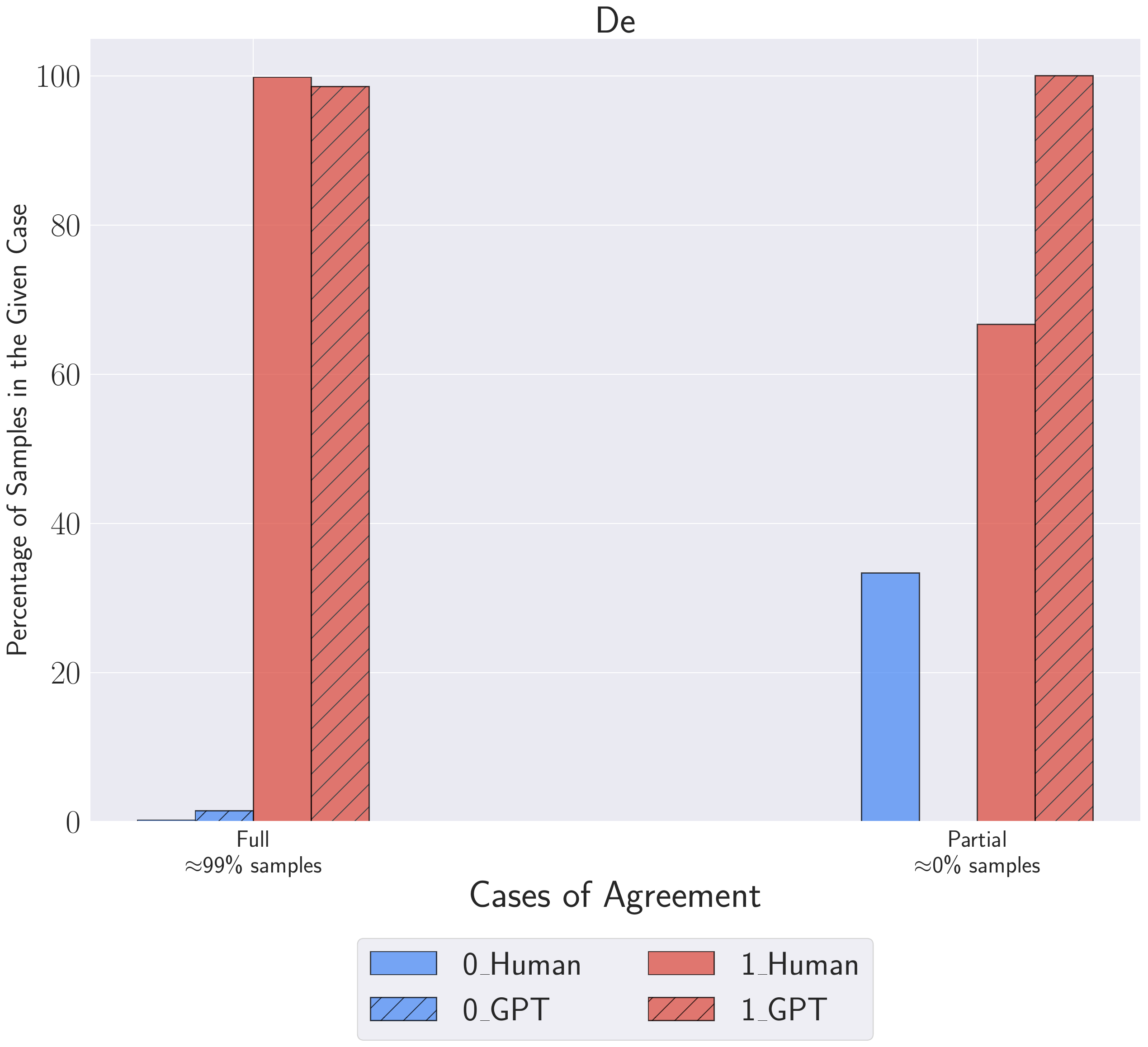}
    \caption{Compound call - German}
    \label{fig:joint_german}
    \end{subfigure}
    \begin{subfigure}[t]{0.33\textwidth}
    \includegraphics[width=0.90\textwidth]{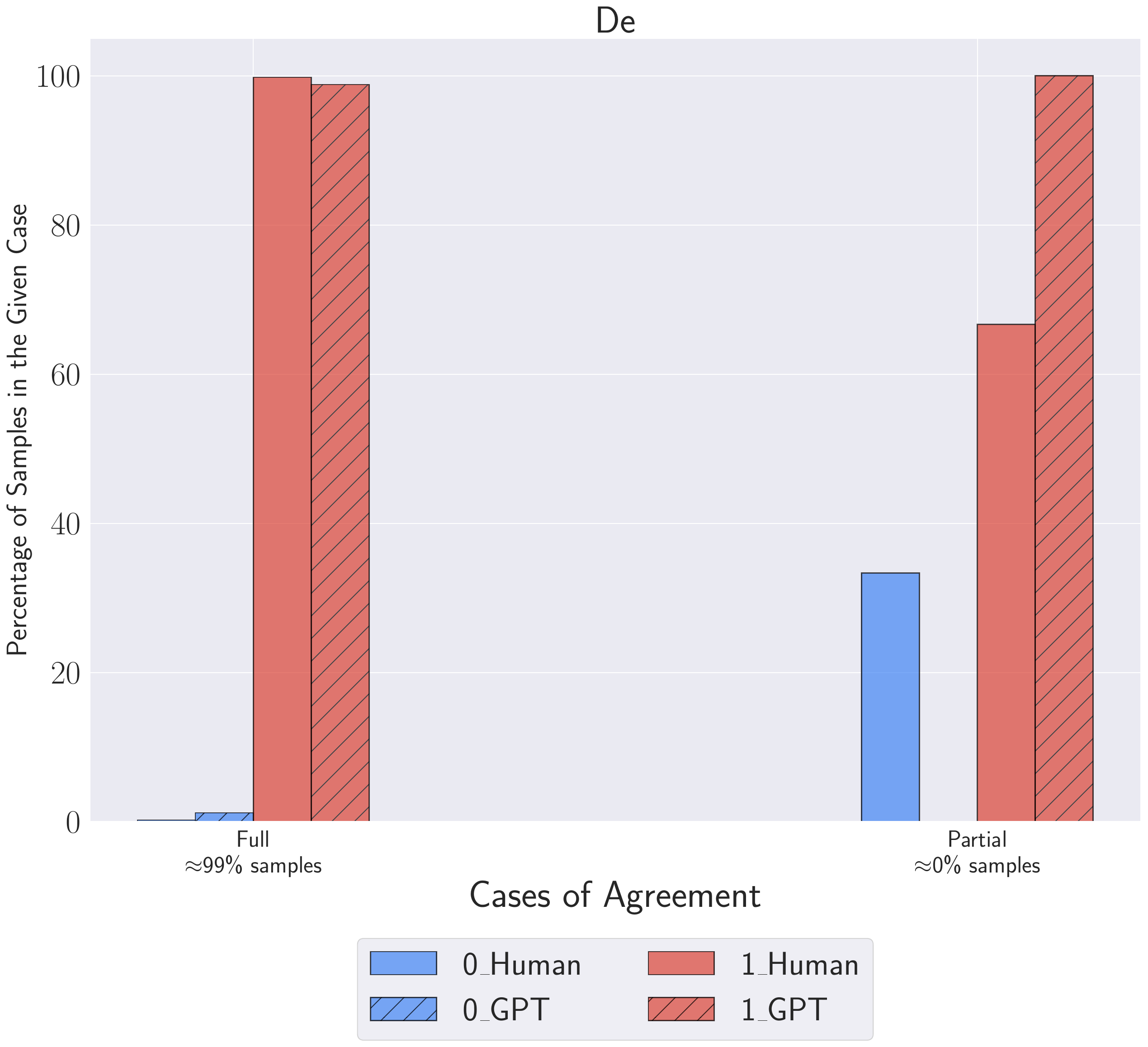}
    \caption{Single Call - German}
    \label{fig:single_german}
    \end{subfigure}
    
    \begin{subfigure}[t]{0.33\textwidth}
    \includegraphics[width=0.90\textwidth]{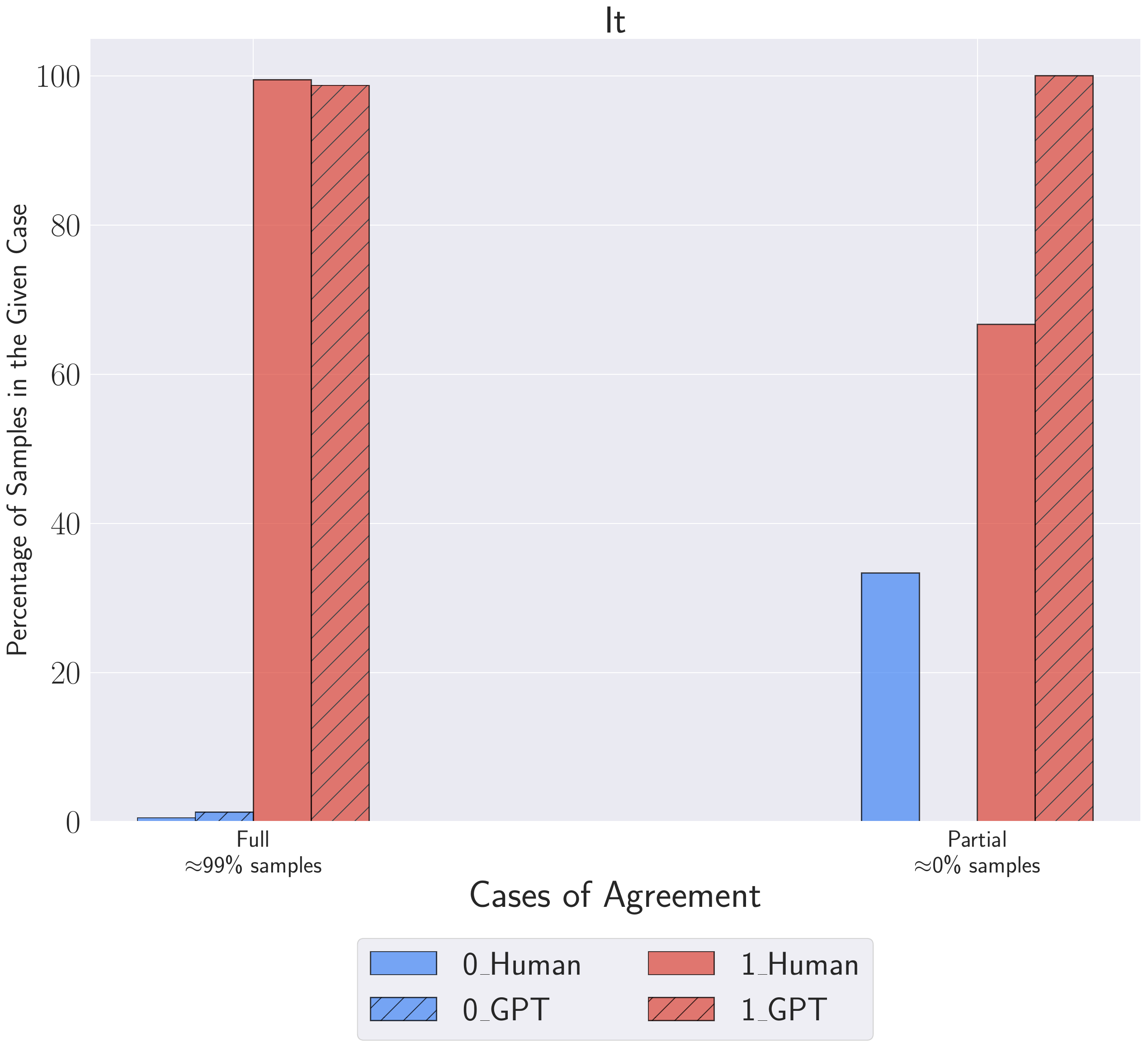}
    \caption{Compound call - Italian}
    \label{fig:joint_italian}
    \end{subfigure}
    \begin{subfigure}[t]{0.33\textwidth}
    \includegraphics[width=0.90\textwidth]{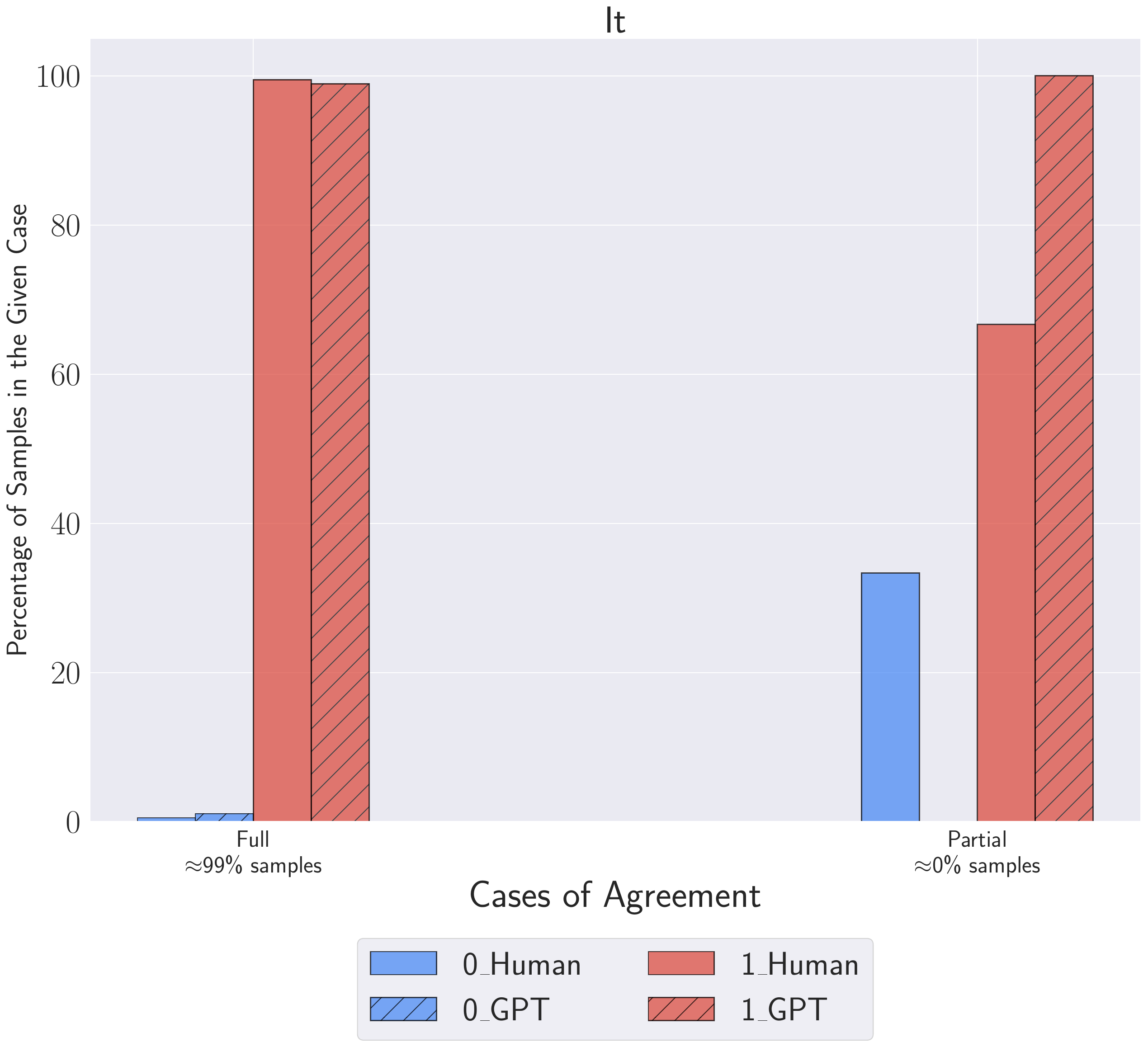}
    \caption{Single Call - Italian}
    \label{fig:single_italian}
    \end{subfigure}\\
    \caption{Class distribution per language (En, Es, Fr, De, It). Results are aggregated over all tasks and metrics with 2 classes (hallucinations and problematic content).}
    \label{fig:classdist3}
\end{figure*}

\begin{figure*}[h]
    \centering
    \begin{subfigure}[t]{0.33\textwidth}
    \includegraphics[width=0.90\textwidth]{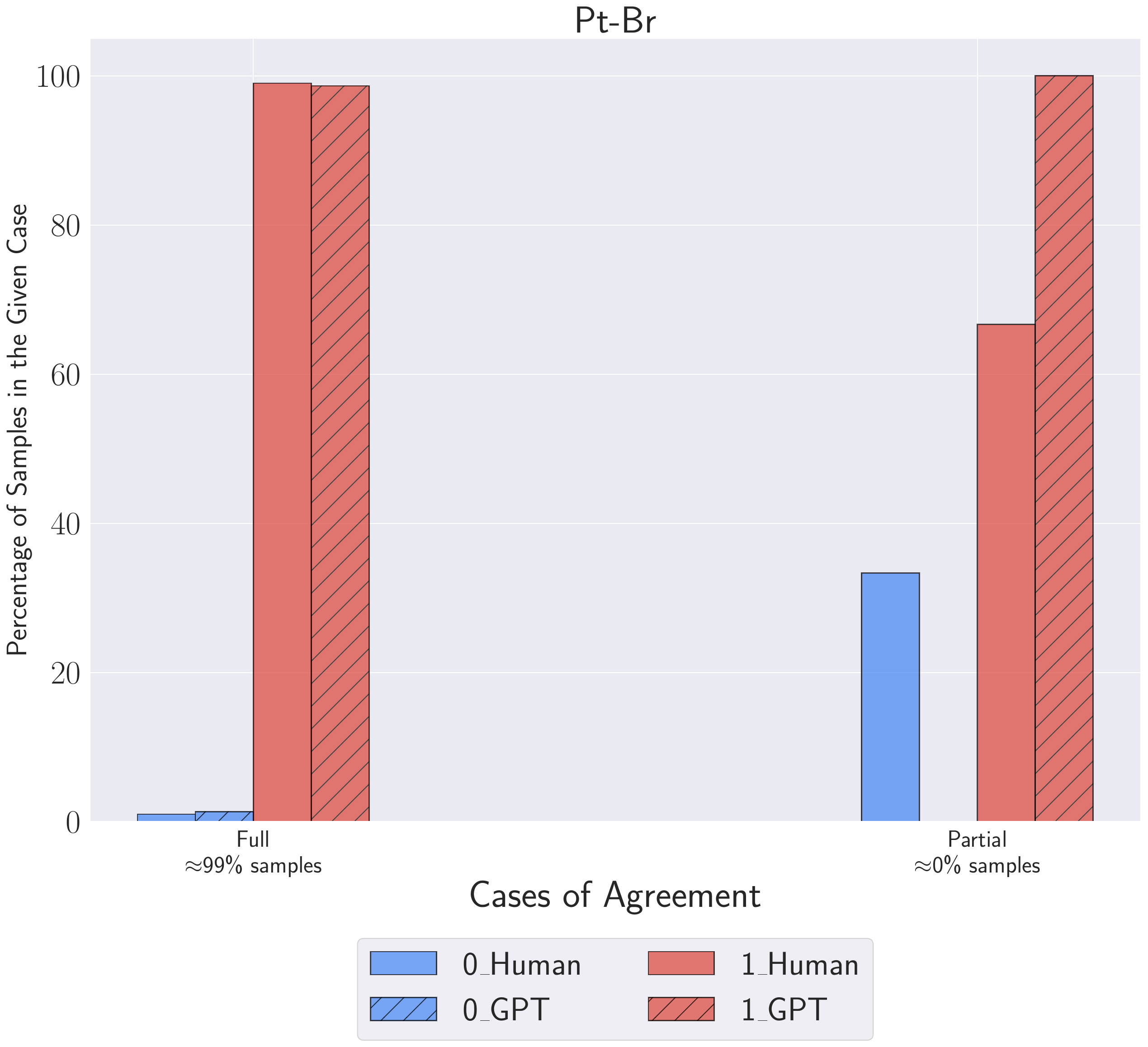}
    \caption{Compound call - Portuguese (Br)}
    \label{fig:joint_brazilian}
    \end{subfigure}
    \begin{subfigure}[t]{0.33\textwidth}
    \includegraphics[width=0.90\textwidth]{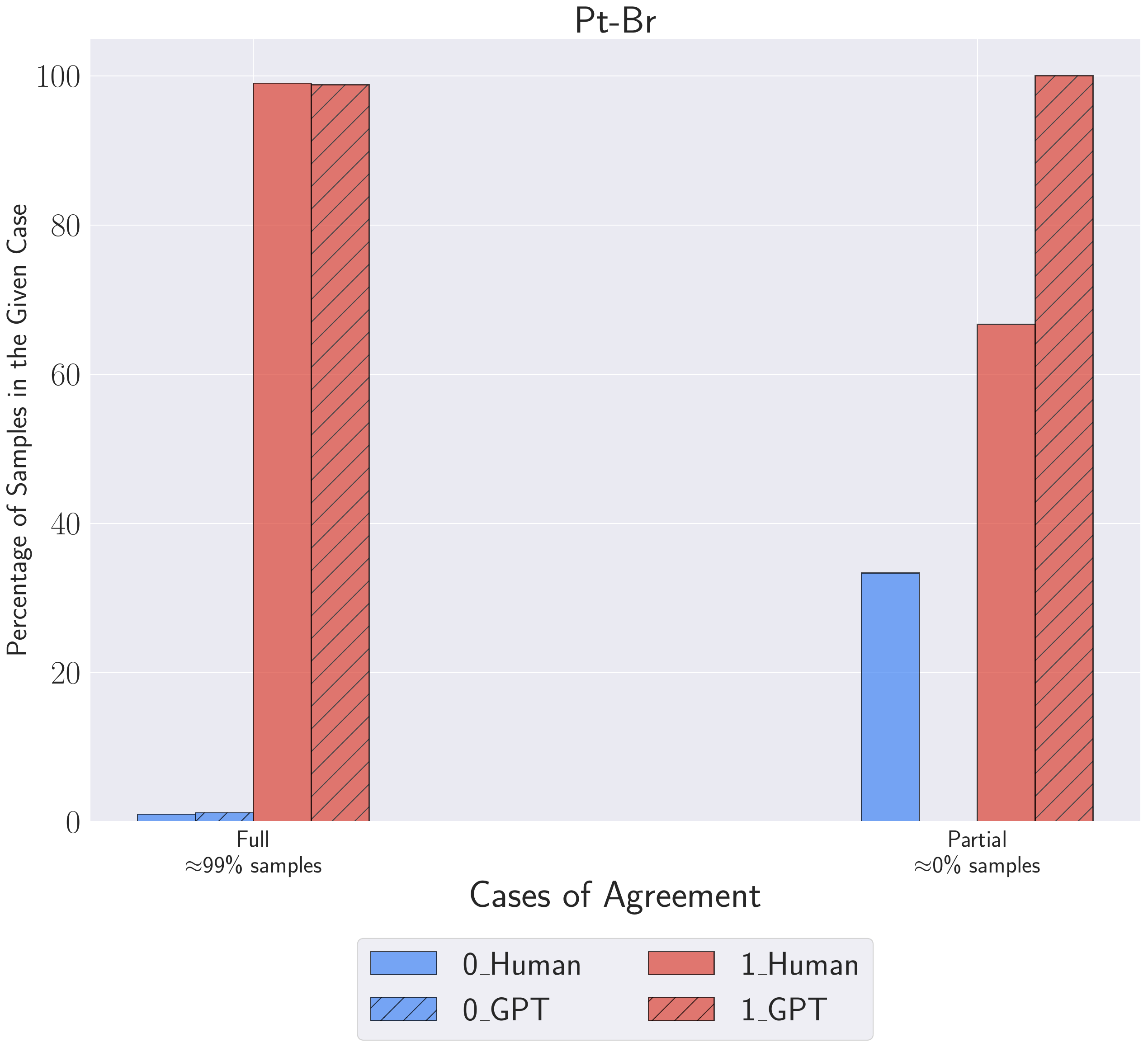}
    \caption{Single Call - Portuguese (Br)}
    \label{fig:single_brazilian}
    \end{subfigure}
    \begin{subfigure}[t]{0.33\textwidth}
    \includegraphics[width=0.90\textwidth]{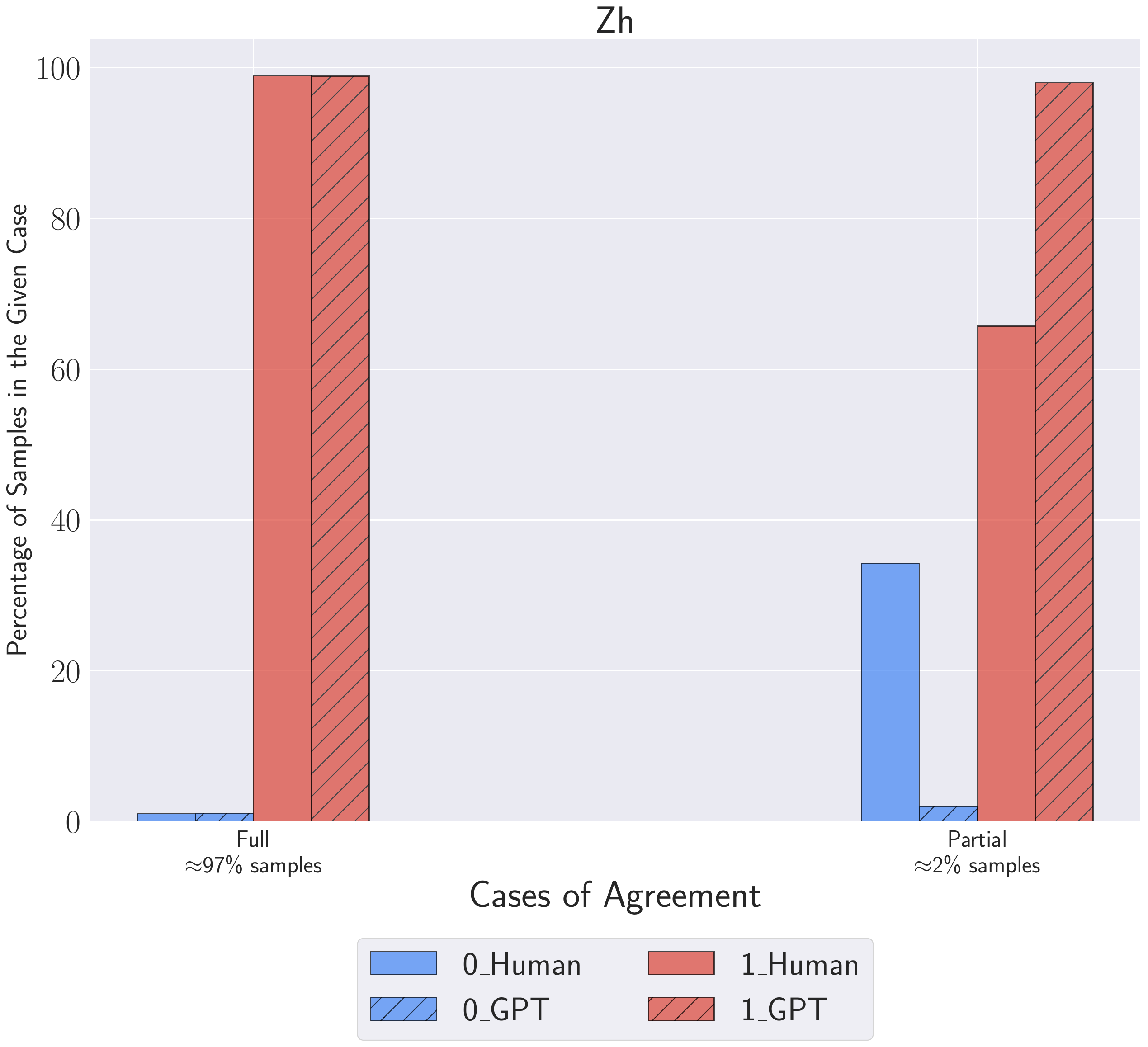}
    \caption{Compound call - Chinese}
    \label{fig:joint_chinese}
    \end{subfigure}
    \begin{subfigure}[t]{0.33\textwidth}
    \includegraphics[width=0.90\textwidth]{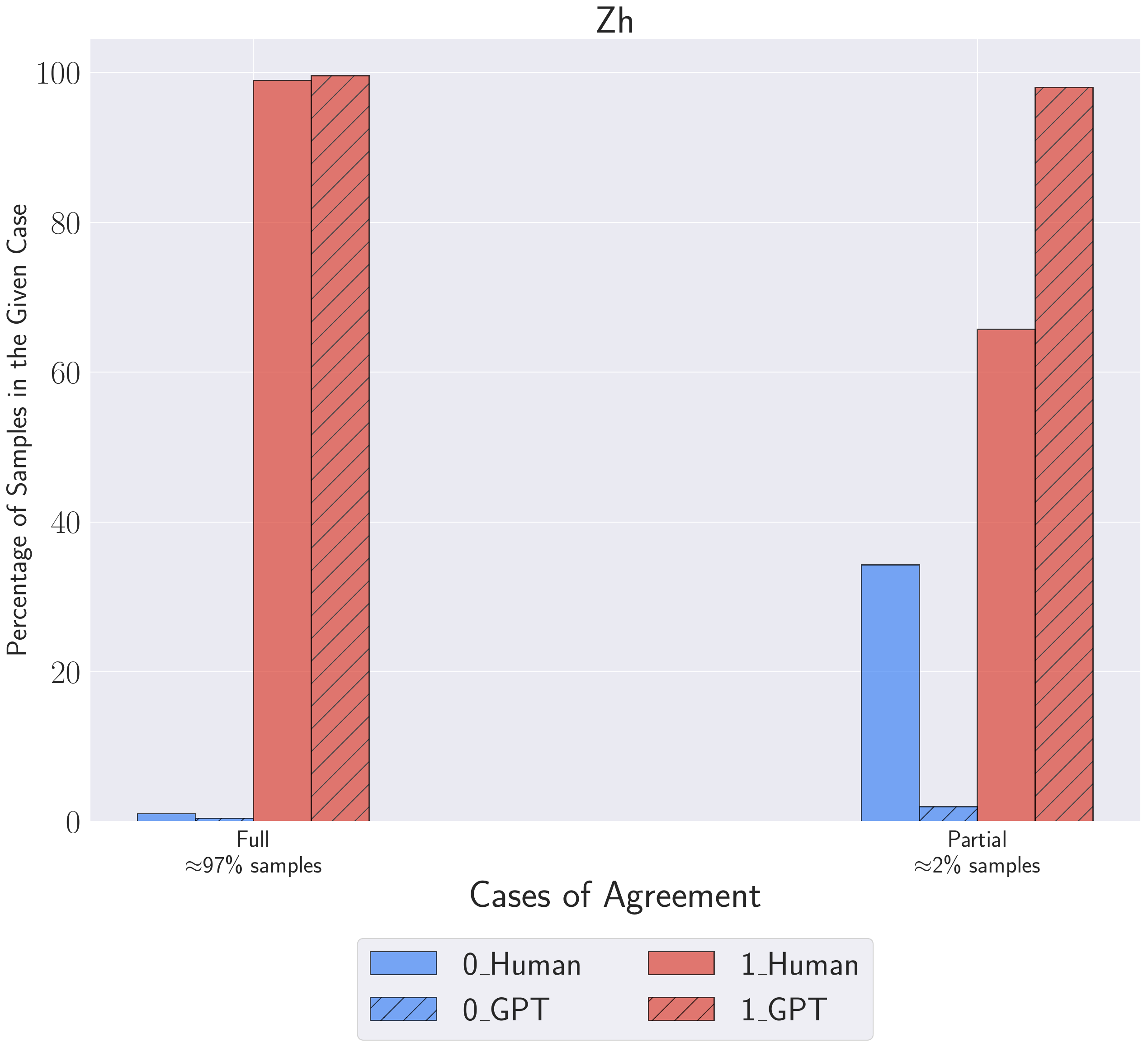}
    \caption{Single Call - Chinese}
    \label{fig:single_chinese}
    \end{subfigure}
    \begin{subfigure}[t]{0.33\textwidth}
    \includegraphics[width=0.90\textwidth]{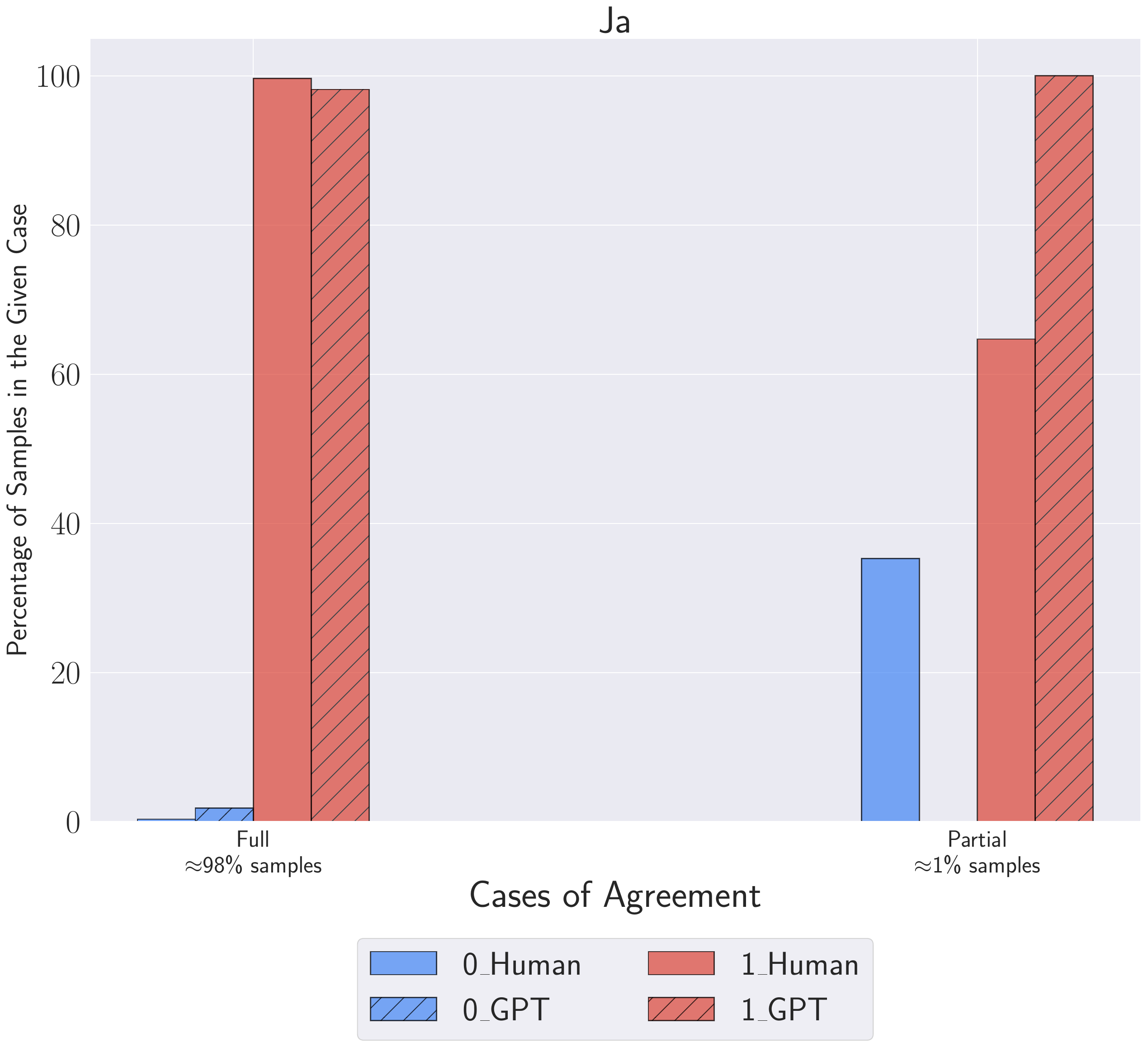}
    \caption{Compound call - Japanese}
    \label{fig:joint_japanese}
    \end{subfigure}
    \begin{subfigure}[t]{0.33\textwidth}
    \includegraphics[width=0.90\textwidth]{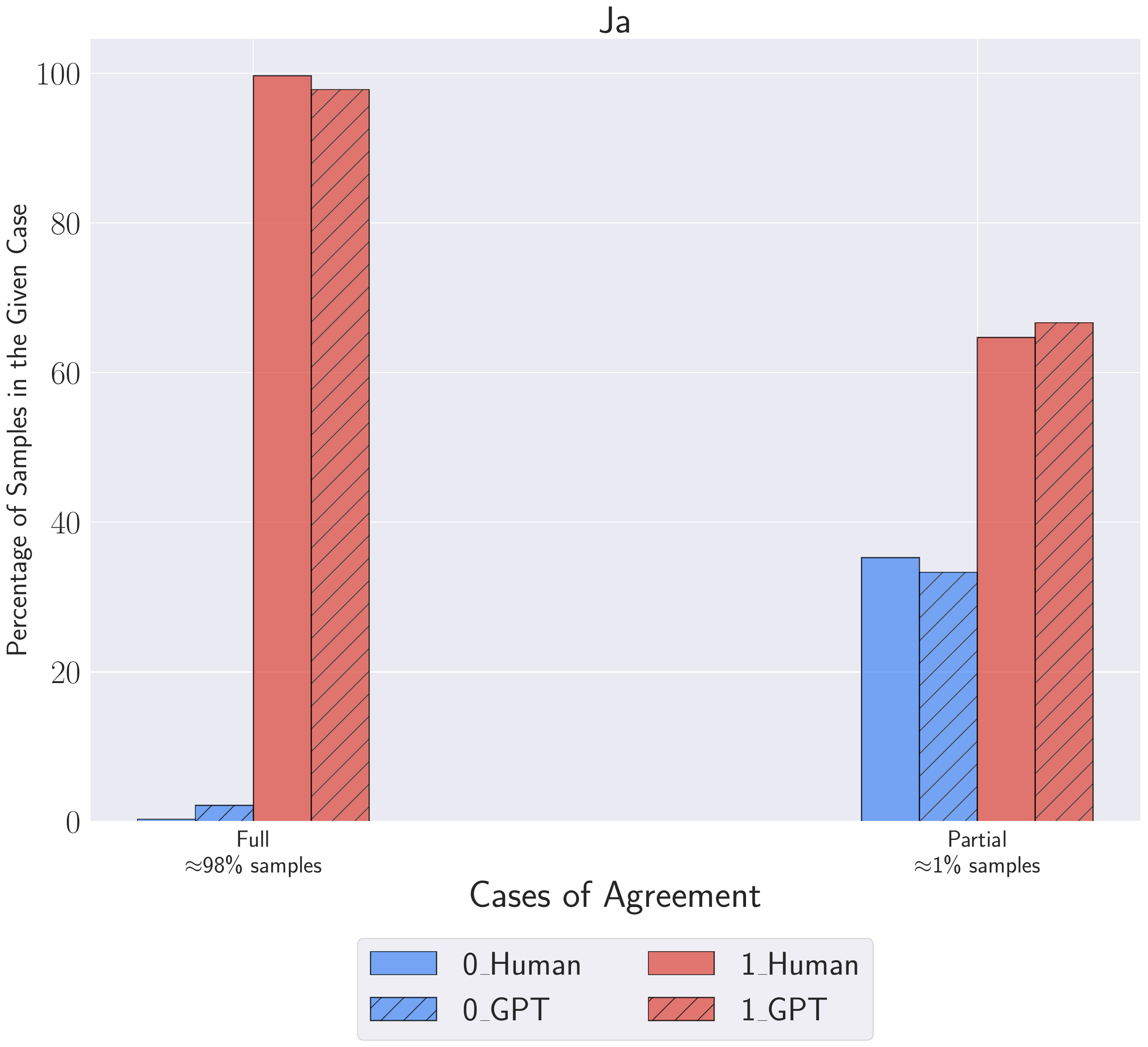}
    \caption{Single Call - Japanese}
    \label{fig:single_japanese}
    \end{subfigure}
    \begin{subfigure}[t]{0.33\textwidth}
    \includegraphics[width=0.90\textwidth]{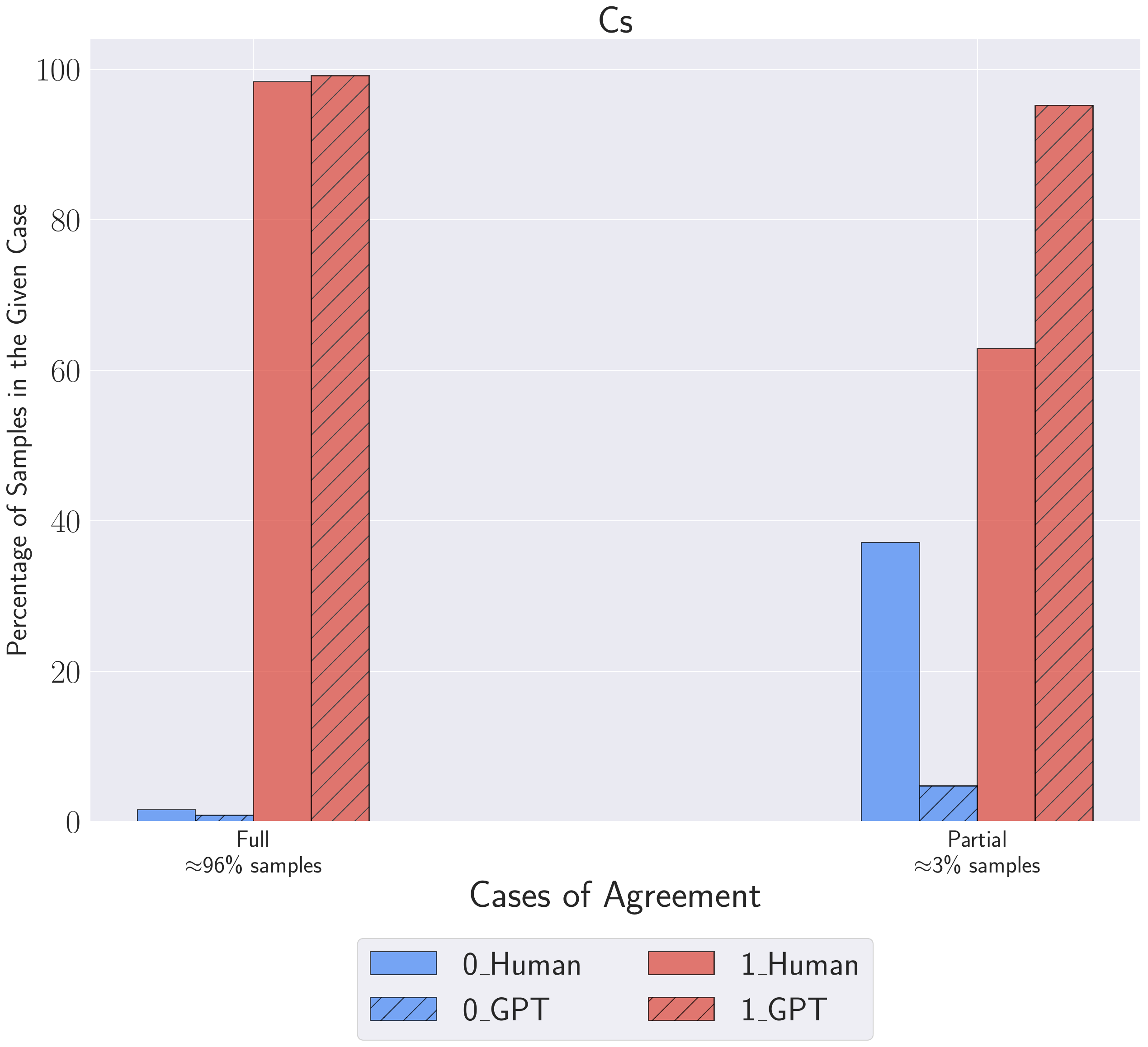}
    \caption{Compound call - Czech}
    \label{fig:joint_czech}
    \end{subfigure}
    \begin{subfigure}[t]{0.33\textwidth}
    \includegraphics[width=0.90\textwidth]{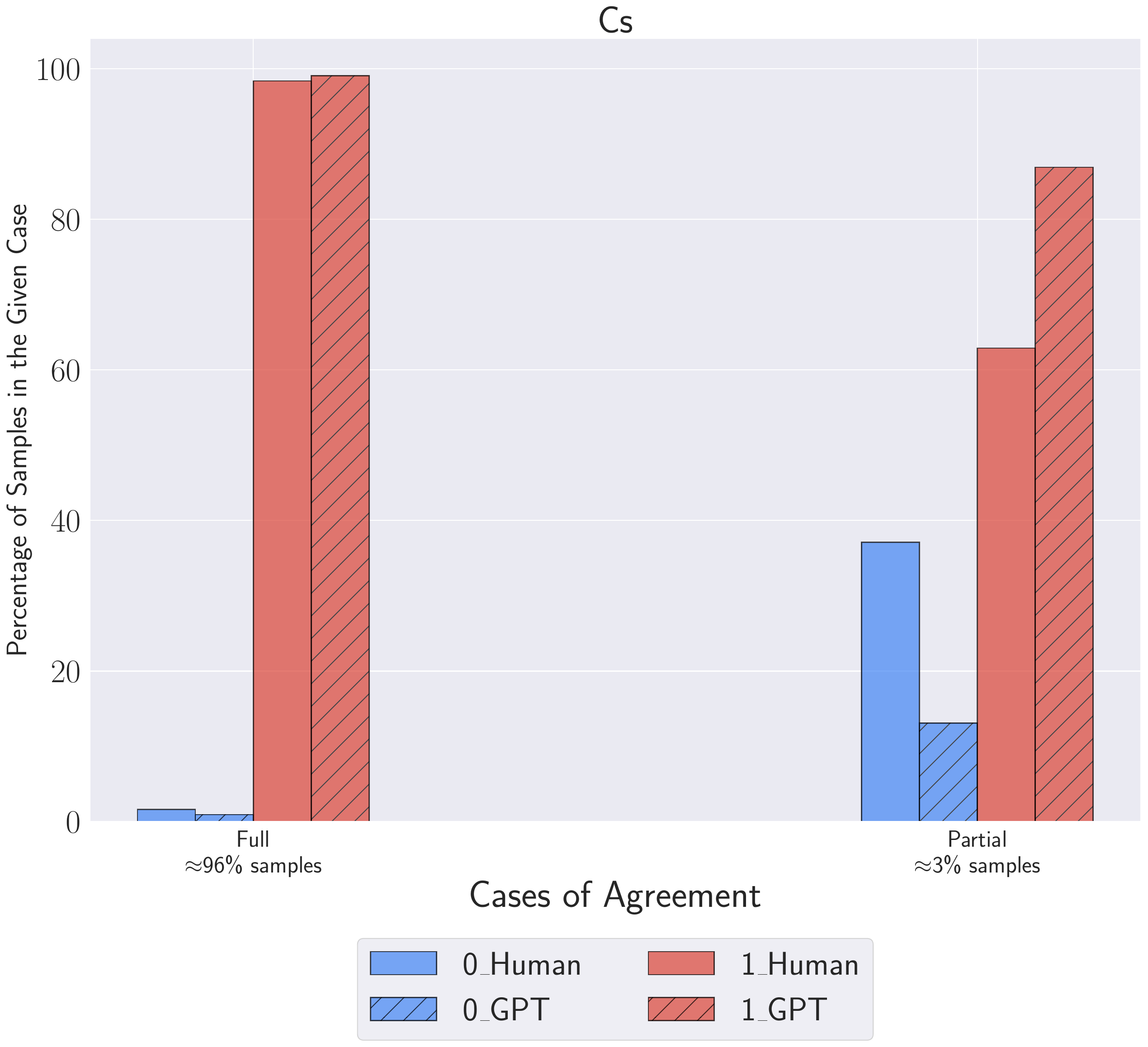}
    \caption{Single Call - Czech}
    \label{fig:single_czech}
    \end{subfigure}\\
    \caption{Class distribution per language (Pt-Br, Zh, Ja, Cz).  Results are aggregated over all tasks and metrics with 2 classes (hallucinations and problematic content).}
    \label{fig:classdist4}
\end{figure*}

Figures \ref{fig:classdist3} and \ref{fig:classdist4} show class distribution for various languages, aggregated over metrics with 2 classes - H, PC. 

\subsection{Temperature Variations}
\label{sec:temp_variations}

\begin{figure}[]
    \centering
    \begin{subfigure}[t]{0.45\textwidth}
        \includegraphics[width=\textwidth]{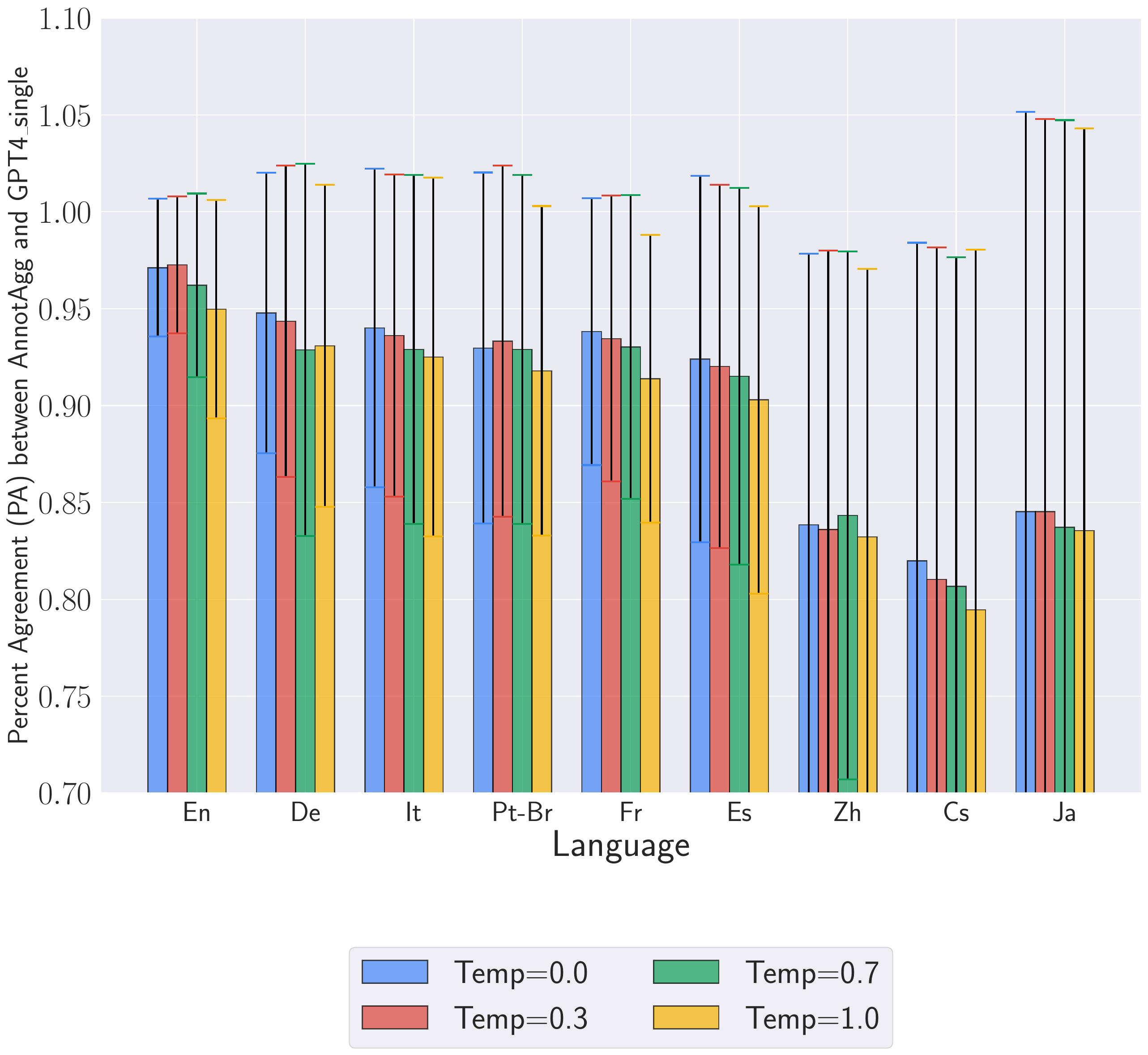}
        \caption{PA by language with temperature variation}
        \label{fig:smallset_PA_by_language_temp}
    \end{subfigure}
    \begin{subfigure}[t]{0.45\textwidth}
        \includegraphics[width=\textwidth]{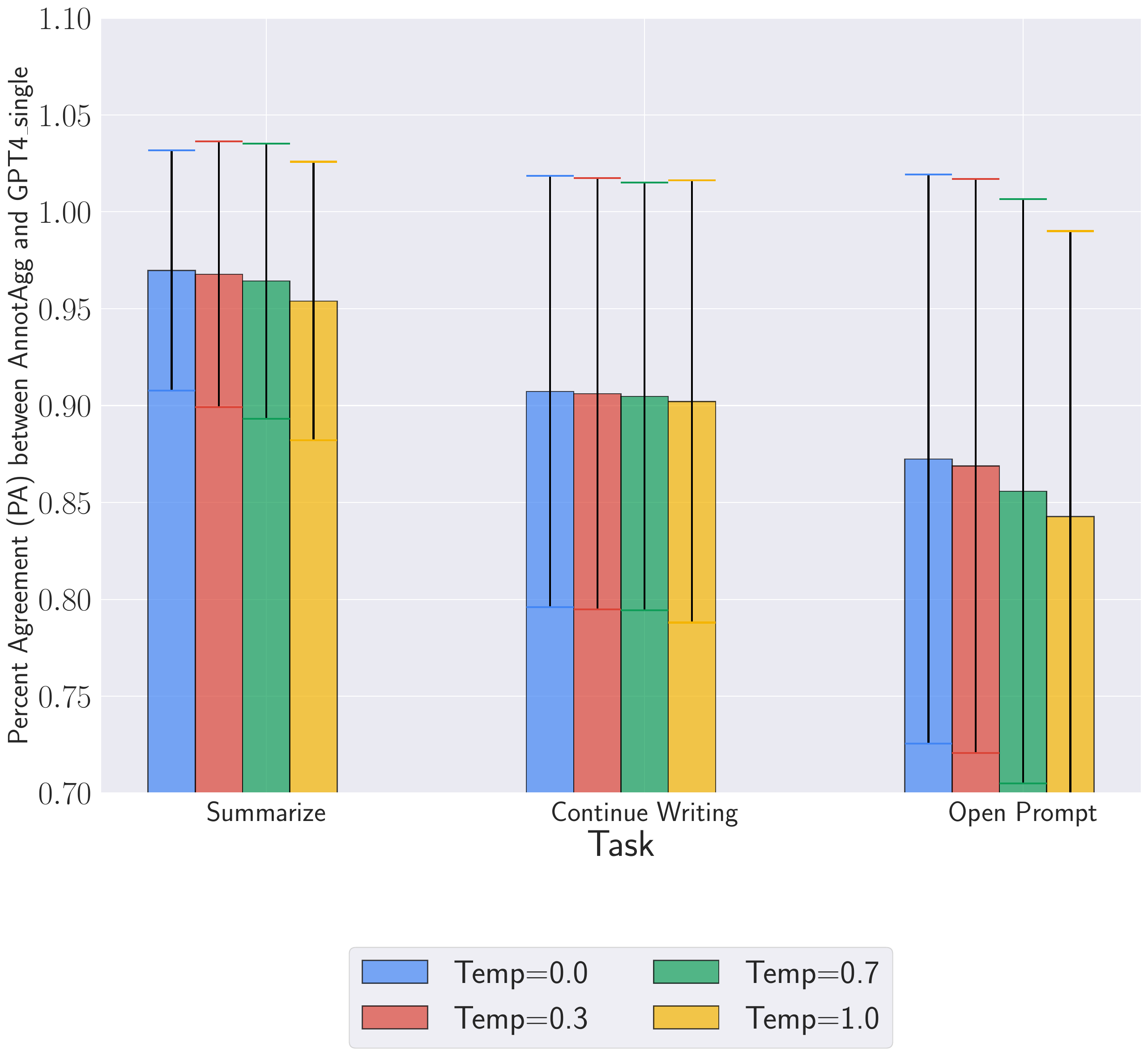}
        \caption{PA by task with temperature variation}
        \label{fig:smallset_PA_by_task_temp}
    \end{subfigure}
    \begin{subfigure}[t]{0.45\textwidth}
        \includegraphics[width=\textwidth]{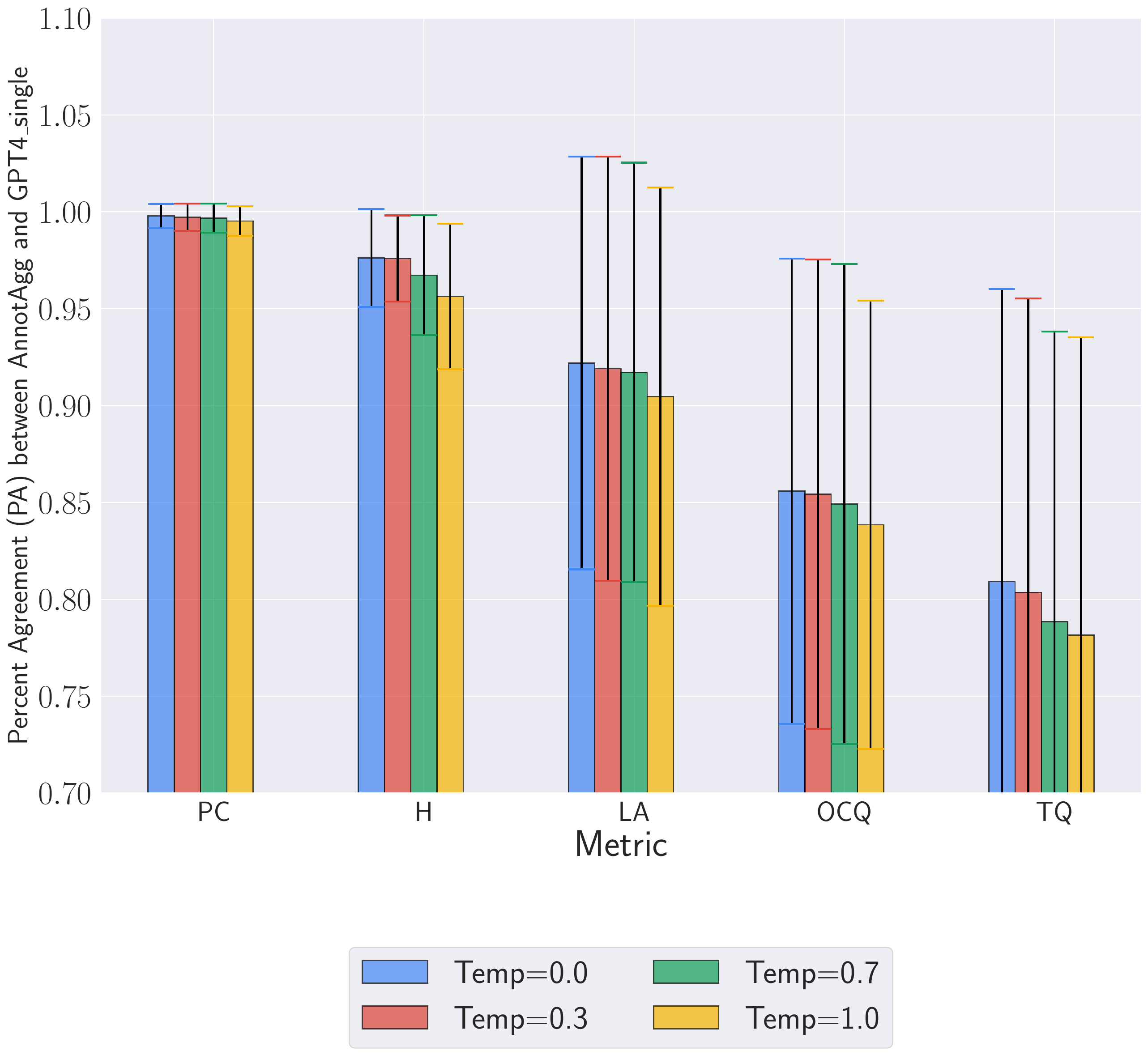}
        \caption{PA by metric with temperature variation}
        \label{fig:smallset_PA_by_metric_temp}
    \end{subfigure}
    \caption{\small{Percentage Agreement (PA) for different cases and temperature variations. Values reported are on the small dataset.}}
    \label{fig:temp_var}
\end{figure}

Figure \ref{fig:temp_var} shows PA values for different temperature values, results are aggregated over language, task, and metrics.

\subsection{few-shot Results}
\begin{figure}[]
    \centering
    \begin{subfigure}[t]{0.45\textwidth}
        \includegraphics[width=\textwidth]{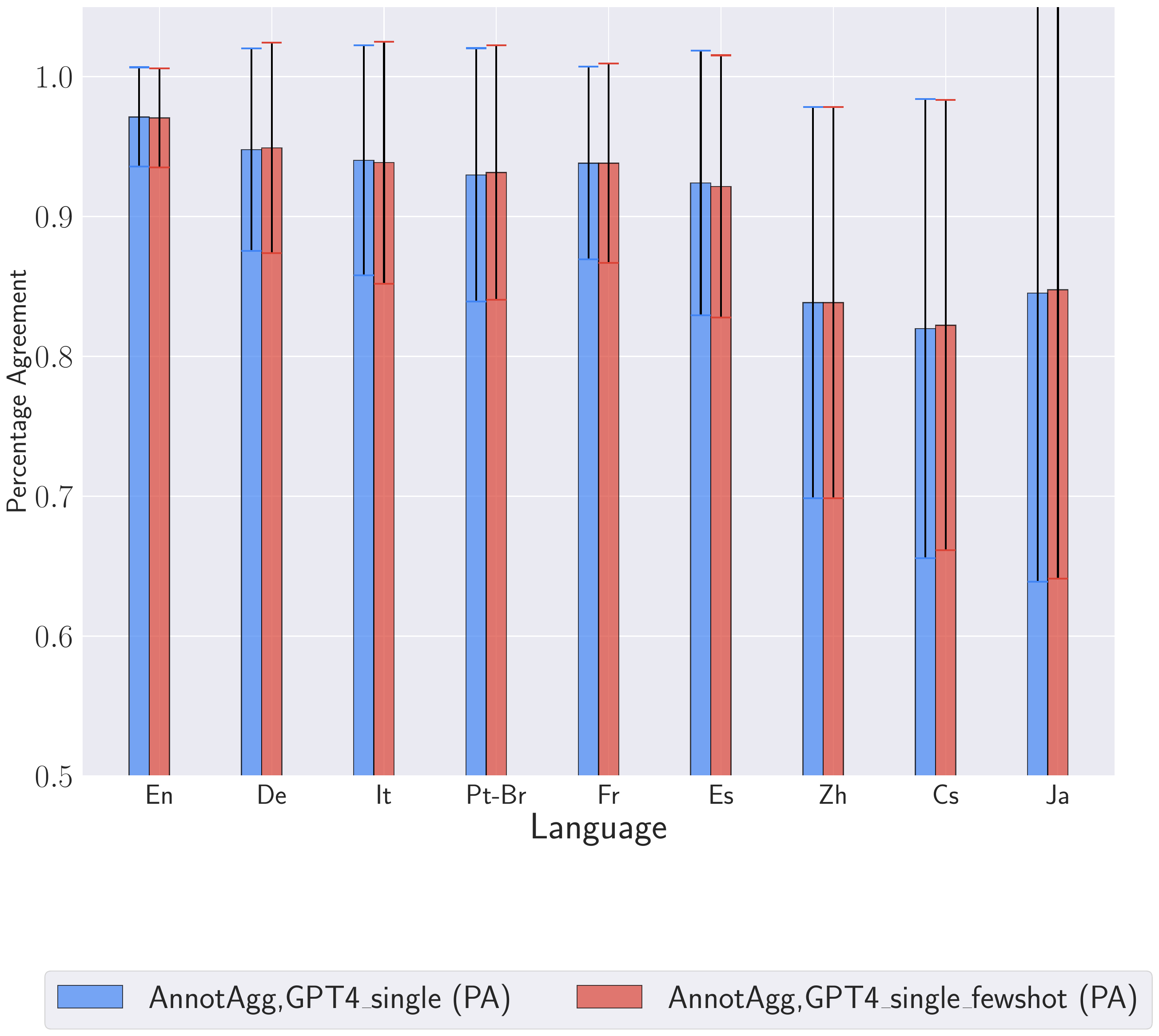}
        \caption{PA by language with few-shot examples}
        \label{fig:smallset_PA_by_language_few}
    \end{subfigure}
    \begin{subfigure}[t]{0.45\textwidth}
        \includegraphics[width=\textwidth]{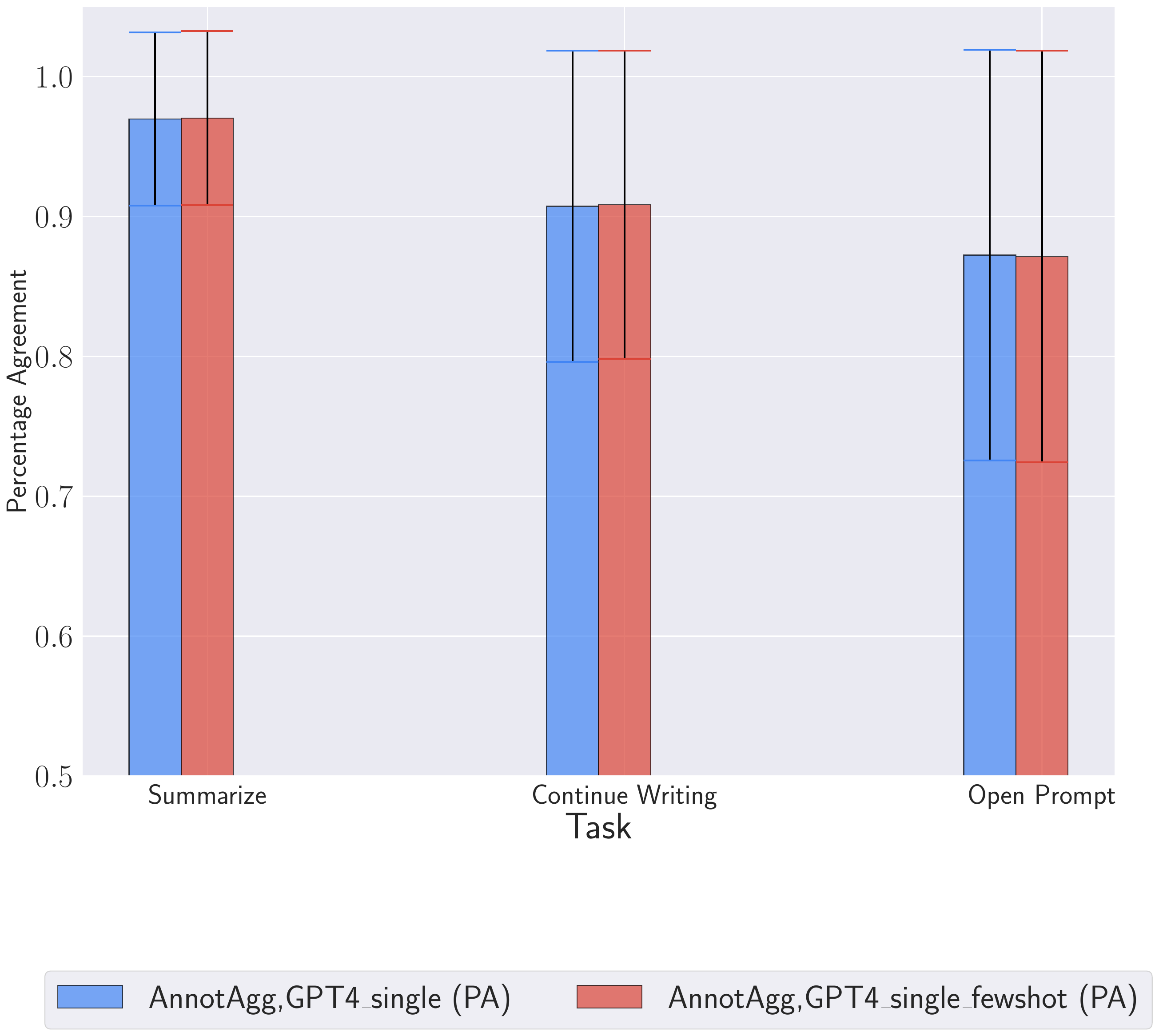}
        \caption{PA by task with few-shot examples}
        \label{fig:smallset_PA_by_task_few}
    \end{subfigure}
    \begin{subfigure}[t]{0.45\textwidth}
        \includegraphics[width=\textwidth]{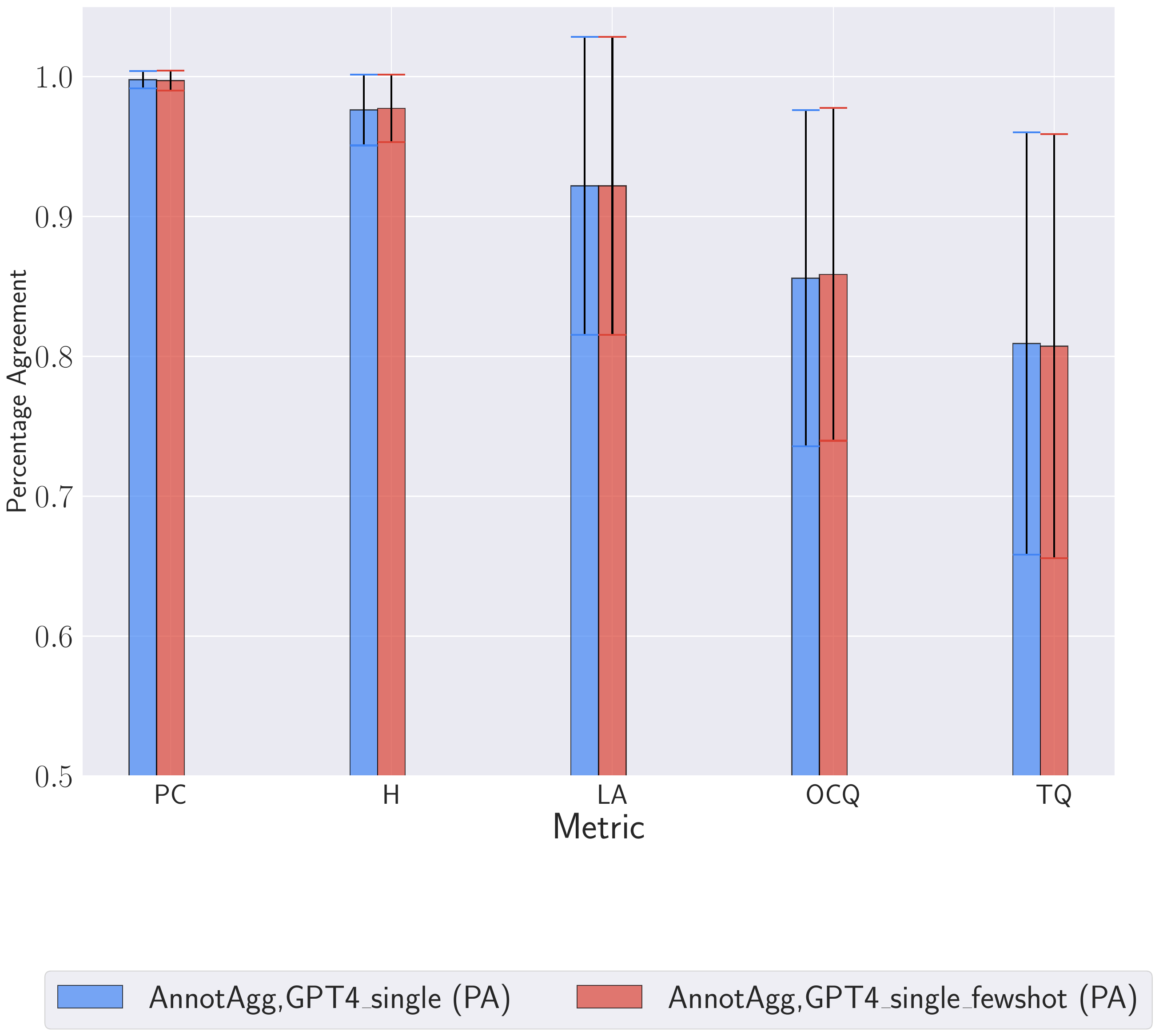}
        \caption{PA by metric with few-shot examples}
        \label{fig:smallset_PA_by_metric_few}
    \end{subfigure}
    \caption{\small{Percentage Agreement (PA) for different cases with few-shot examples. Values reported are on the small dataset.}}
    \label{fig:few-shot} 
\end{figure}

Figure \ref{fig:few-shot} shows PA values for few-shot prompting, results are aggregated over language, task, and metrics.

\label{sec:few-shot}

\end{document}